\documentclass{cambrian}
\usepackage[utf8]{inputenc} 
\usepackage[T1]{fontenc}    
\usepackage{hyperref}       
\usepackage{url}            
\usepackage{booktabs}       
\usepackage{amsfonts}       
\usepackage{nicefrac}       
\usepackage{microtype}      

\usepackage{mwe}
\usepackage{graphicx}
\usepackage{amssymb}
\usepackage{adjustbox}
\usepackage{colortbl}
\usepackage{array}
\usepackage{multirow}
\usepackage{makecell}
\usepackage{bbm}
\usepackage{collcell,xfp}
\usepackage{pgf}
\usepackage{tikz}
\usepackage[most]{tcolorbox}
\usepackage{csquotes}
\usepackage[noorphans,vskip=1em,leftmargin=1em]{quoting}
\usepackage{pifont} 
\usepackage{enumitem} 
\usepackage{tcolorbox} 
\usepackage{forest}
\usepackage{wrapfig}
\usepackage{caption}
\usepackage{subcaption}
\usepackage{longtable}
\usepackage{algpseudocode}
\usepackage{amsmath}
\usepackage[capitalize]{cleveref}
\usepackage[ruled,vlined,linesnumbered]{algorithm2e} 
\usepackage[percent]{overpic}
\usepackage{xspace}
\usepackage[normalem]{ulem}  
\usepackage{tocloft}  
\usepackage{natbib}  

\usepackage{fontawesome}
\newcommand{\huggingface}{\raisebox{-1.5pt}{\includegraphics[height=1.05em]{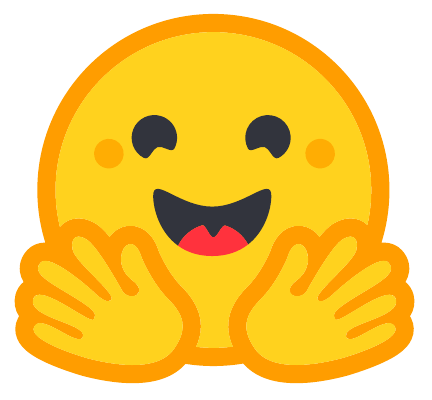}}\xspace}
\newcommand{\github}{\raisebox{-1.5pt}{\includegraphics[height=1.05em]{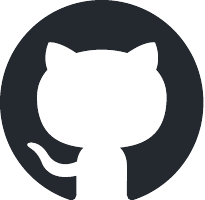}}\xspace}

\newcommand{\worldwideweb}{\raisebox{-1.5pt}{\includegraphics[height=1.05em]{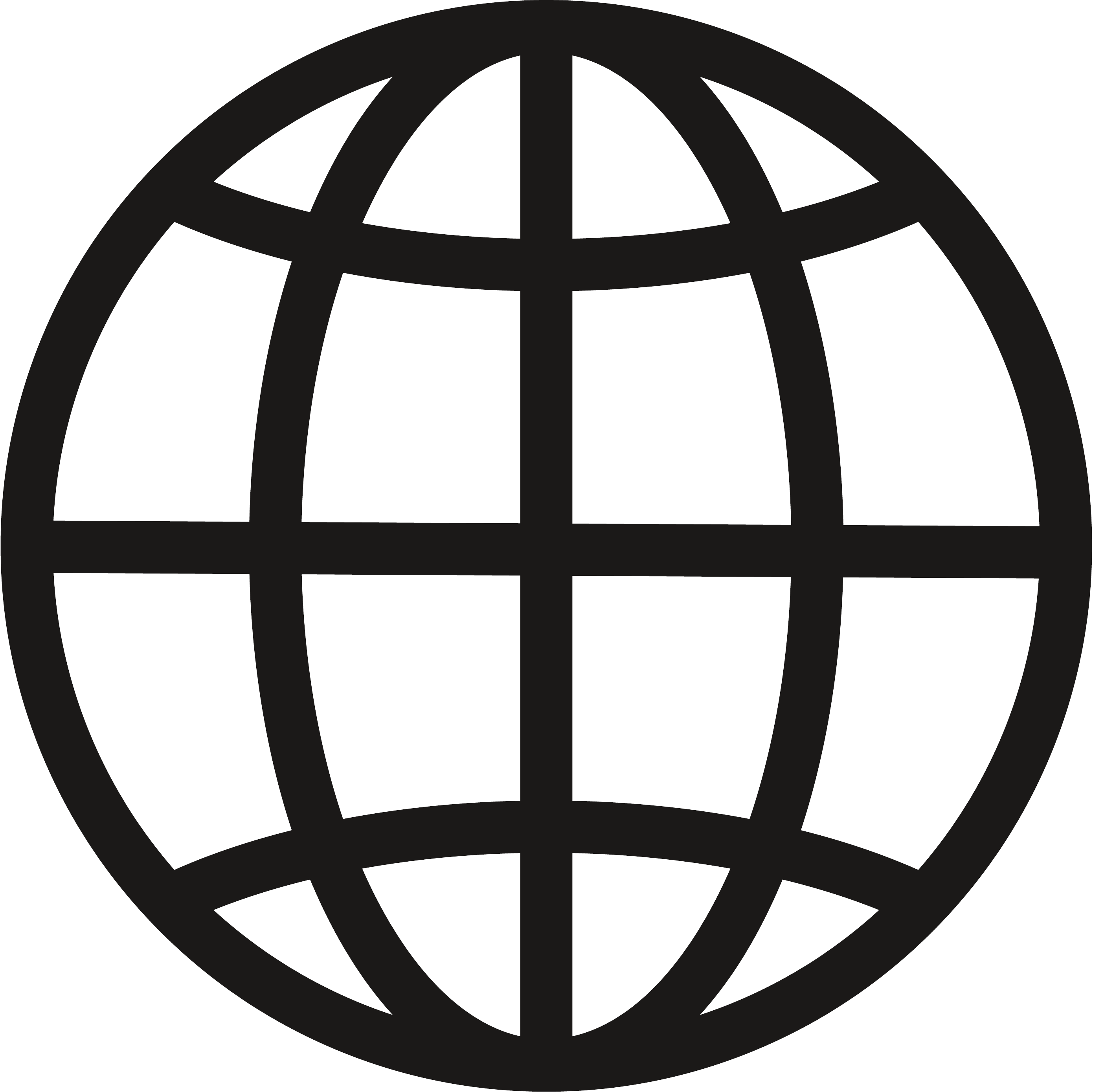}}\xspace}

\definecolor{scholarblue}{rgb}{0.21,0.49,0.74}
\definecolor{bluelink}{RGB}{0,113,188}
\definecolor{greenlink}{RGB}{0,188,113}
\hypersetup{
    colorlinks=true,%
    citecolor=scholarblue,%
    filecolor=red,%
    linkcolor=red!93!black,%
    urlcolor=bluelink
}

\usepackage{titlesec}
\titlespacing*{\paragraph}
    {0pt}     
    {0.25em}  
    {1em}  

\newcommand{\finding}[2]{
    \begin{tcolorbox}[
        colback=white!90!gray,     
        colframe=teal!60!black,     
        arc=5pt,                    
        boxsep=5pt,                 
        left=10pt,                  
        right=10pt,                 
        top=2pt,                    
        bottom=2pt,                 
        boxrule=0.8pt,              
        drop shadow=gray!50!white,  
        enhanced jigsaw             
    ]
    #2
    \end{tcolorbox}
}

\definecolor{navyblue}{HTML}{0071BC}
\iftrue   
\newcommand{\displaytodo}[1]{#1}
\else
\newcommand{\displaytodo}[1]{}
\fi

\newcommand{\cambrianS}{Cambrian-\emph{S}\xspace}
\newcommand{\vsidata}{VSI-590K\xspace}
\newcommand{\osmix}{CambrianS-3M\xspace}
\newcommand{\vso}{VSR\xspace}
\newcommand{\vsc}{VSC\xspace}
\newcommand{\vsisuper}{\textsc{VSI-Super}\xspace}

\def\eg{\emph{e.g.}} 
\def\ie{\emph{i.e.}}
 
\def\etc{\emph{etc.}}

\definecolor{blindcolor}{HTML}{AB2AC6}    
\definecolor{chancecolor}{HTML}{F59E0B}   
\definecolor{singlecolor}{HTML}{06B6D4}   
\definecolor{multiplecolor}{HTML}{2563EB} 
\definecolor{captioncolor}{HTML}{22C55E}  

 \newcommand{\culine}[2]{%
    \def\temp@uline{\bgroup\markoverwith
        {\textcolor{#1}{\rule[-0.5ex]{2pt}{1pt}}}\ULon}%
    \temp@uline{#2}%
}
 \newcommand{\cthickuline}[3][0.8pt]{%
    \def\temp@uline{\bgroup\markoverwith
        {\textcolor{#2}{\rule[-0.5ex]{2pt}{#1}}}\ULon}%
    \temp@uline{#3}%
}

\newcommand{\blindul}[1]    {\cthickuline{blindcolor}{#1}}
\newcommand{\chanceul}[1]   {\cthickuline{chancecolor}{#1}}
\newcommand{\singleul}[1]   {\cthickuline[0.5pt]{singlecolor}{#1}}
\newcommand{\multipleul}[1] {\cthickuline[1pt]{multiplecolor}{#1}}
\newcommand{\captionul}[1]  {\cthickuline{captioncolor}{#1}}

\title{\center{Cambrian-\emph{S}: Towards Spatial Supersensing in Video}}

\renewcommand\Affilfont{\normalfont\fontsize{11}{15}\selectfont\centering}
\author{
    Shusheng~Yang\textsuperscript{1}$^{*}$ \quad\ 
    Jihan~Yang\textsuperscript{1}$^{*}$ \quad\ 
    Pinzhi~Huang\textsuperscript{1}$^{\dagger}$ \quad\ 
    Ellis~Brown\textsuperscript{1}$^{\dagger}$ \quad\ 
    Zihao~Yang\textsuperscript{1} \quad
    Yue~Yu\textsuperscript{1} \quad
    Shengbang~Tong\textsuperscript{1} \quad
    Zihan~Zheng\textsuperscript{1} \quad
    Yifan~Xu\textsuperscript{1} \quad
    Muhan~Wang\textsuperscript{1} \quad
    Daohan~Lu\textsuperscript{1} \quad
    Rob~Fergus\textsuperscript{1} \quad
    Yann~LeCun\textsuperscript{1} \quad
    Li~Fei-Fei\textsuperscript{2} \quad
    Saining~Xie\textsuperscript{1} \\
    \textsuperscript{1} New York University \quad \textsuperscript{2} Stanford University
}

\begin{abstract}
We argue that progress in true multimodal intelligence calls for a shift from reactive, task-driven systems and brute-force long context towards a broader paradigm of \emph{supersensing}.
We frame spatial supersensing as four stages beyond linguistic-only understanding:
semantic perception (naming what is seen), streaming event cognition (maintaining memory across continuous experiences), implicit 3D spatial cognition (inferring the world behind pixels), and predictive world modeling (creating internal models that filter and organize information).
Current benchmarks largely test only the early stages, offering narrow coverage of spatial cognition and rarely challenging models in ways that require true world modeling.
To drive progress in spatial supersensing, we present \vsisuper, a two-part benchmark: {\texttt{\vso}} (long-horizon \underline{v}isual \underline{s}patial \underline{r}ecall) and {\texttt{\vsc}} (continual \underline{v}isual \underline{s}patial \underline{c}ounting). These tasks require arbitrarily long video inputs yet are resistant to brute-force context expansion.
We then test data scaling limits by curating \vsidata{} and training \cambrianS, achieving $+30\%$ absolute improvement on VSI-Bench without sacrificing general capabilities.
Yet performance on \vsisuper\ remains limited, indicating that scale alone is insufficient for spatial supersensing.
We propose \emph{predictive sensing} as a path forward, presenting a proof-of-concept in which a self-supervised next-latent-frame predictor leverages \emph{surprise} (prediction error) to drive memory and event segmentation.
On \vsisuper, this approach substantially outperforms leading proprietary baselines, showing that spatial supersensing requires models that not only see but also anticipate, select, and organize experience.
\end{abstract}

\renewcommand{\thefootnote}{\arabic{footnote}}  

\begin{document}

\maketitle

\begingroup
    \renewcommand{\thefootnote}{\fnsymbol{footnote}}
    \footnotetext[1]{SY led the project, JY and SY contributed equally.}
    \footnotetext[2]{Core contributor.}
\endgroup


\begin{center}
    \renewcommand{\arraystretch}{1.5}
    \begin{tabular}{rll}
        \worldwideweb{} & \textbf{Website} & \url{https://cambrian-mllm.github.io}\\
        \github{} & \textbf{Code} & \url{https://github.com/cambrian-mllm/cambrian-s}\\
        \huggingface{} & \textbf{\cambrianS Models} & \href{https://hf.co/collections/nyu-visionx/cambrian-s-models-68ed68c1caded228ae5a6f85}{\nolinkurl{https://hf.co/collections/nyu-visionx/cambrian-s}} \\
        \huggingface{} & \textbf{\vsidata} & \url{https://hf.co/datasets/nyu-visionx/vsi-590k} \\
        \huggingface{} & \textbf{\vsisuper} & \url{https://hf.co/collections/nyu-visionx/vsi-super} \\
    \end{tabular}
\end{center}


\newpage
{
    \hypersetup{linkcolor=black}
    \tableofcontents
}
\newpage


\section{Introduction}

\begin{figure}[t]
    \centering
    \includegraphics[width=0.955\linewidth]{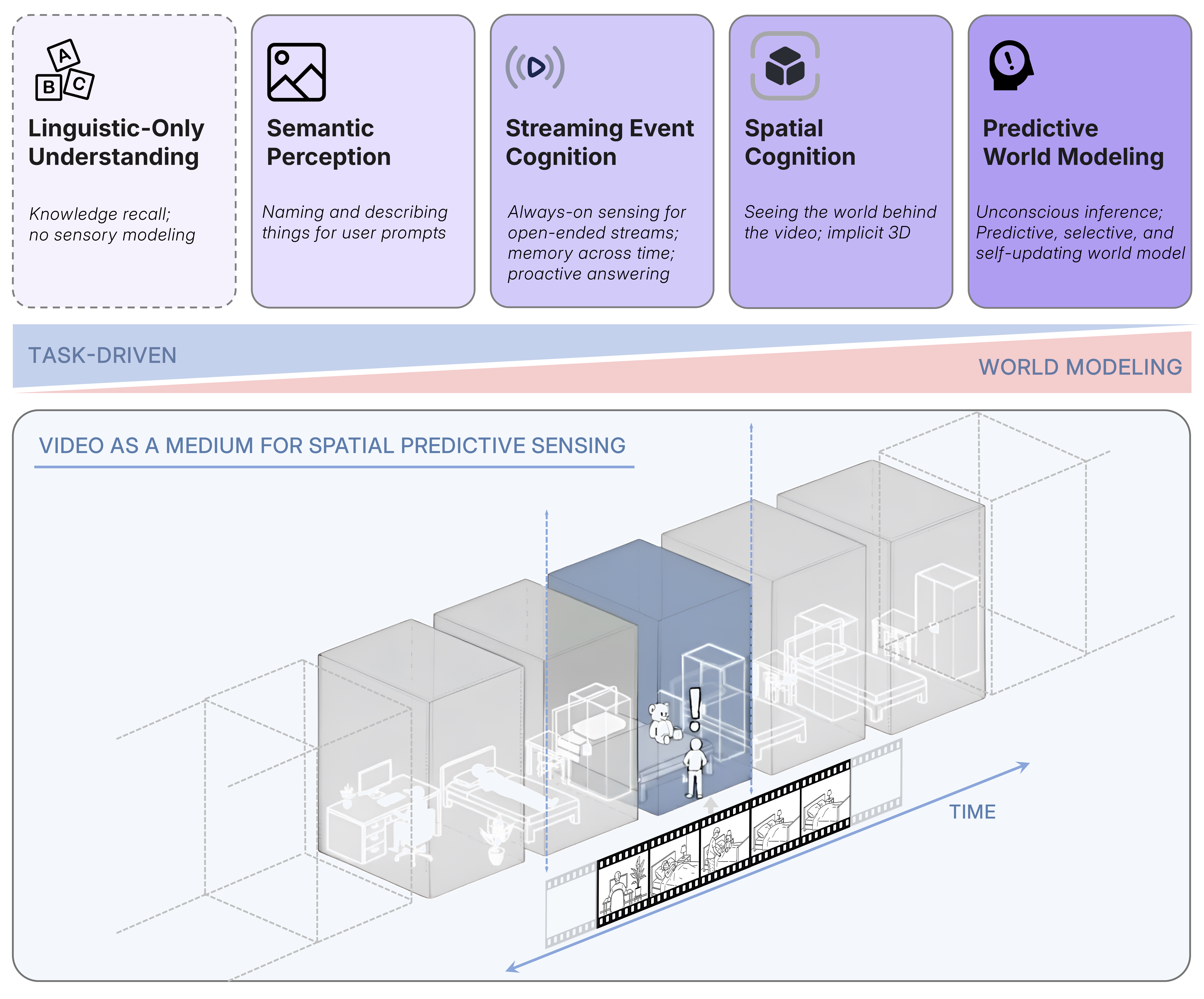}
    \caption{
        \small
        \textbf{From pixels to predictive mind.}
        We look beyond \emph{linguistic-only} understanding to envision multimodal intelligence that sees, remembers, and reasons as part of a continuous, lived world.
        It begins with \emph{semantic perception}: naming and describing what is seen. \emph{Streaming event cognition} goes further, enabling always-on sensing across continuous input streams, integrating memory, and supporting proactive responses. \emph{Spatial cognition} captures the implicit 3D structure of video, enabling reasoning about objects, configurations, and metrics. Finally, a \emph{predictive world model} emerges, one that learns passively from experience, updates through prediction and surprise, and retains information for future use.
        \textbf{Lower illustration:} Video serves as the ideal experimental domain. Models must advance from frame-level Q\&A to constructing implicit world models that enable deeper spatial reasoning, scale to unbounded horizons, and achieve supersensing that rivals, and ultimately surpasses, human visual intelligence.
    }\label{fig:teaser}
\end{figure}

A video is not just a sequence of frames in isolation. It is a continual, high-bandwidth projection of a hidden, evolving 3D world onto pixels~\cite{gibson2014ecological, marr2010vision}. Although multimodal large language models (MLLMs) have advanced rapidly by pairing strong image encoders with language models~\cite{achiam2023gpt, team2024gemini, anthropic_claude35_sonnet_2024, liu2023visual, tong2024cambrian}, most video extensions~\cite{wang2024internvideo2, li2024llava, bai2025qwen2} remain fundamentally constrained. They still treat video as sparse frames, underrepresent spatial structure and dynamics~\cite{yang2024think}, and lean heavily on textual recall~\cite{zohar2024apollo}, thus overlooking what makes the video modality uniquely powerful. 

\noindent In this paper, we argue that advancing toward true multimodal intelligence requires a shift from language-centric perception toward spatial \emph{supersensing:} the capacity not only to see, but also to construct, update and predict with an implicit model of the 3D world from continual sensory experience. We do not claim to realize supersensing here; rather, we take an initial step toward it by articulating the developmental path that could lead in this direction and by demonstrating early prototypes along that path:

\newpage
\begin{enumerate}[start=0,leftmargin=1em,nosep]
    \item \textbf{\color{gray} (Linguistic-only understanding):} no sensory capabilities; reasoning confined to text and symbols. Current MLLMs have progressed beyond this stage, yet still retain traces of its bias.
    \item \textbf{Semantic perception:} parsing pixels into objects, attributes, and relations. This corresponds to the strong multimodal \emph{``show and tell''} capabilities present in MLLMs.
    \item \textbf{Streaming event cognition:} processing live, unbounded streams while proactively interpreting and responding to ongoing events. This aligns with efforts to make MLLMs real-time assistants.
    \item \textbf{Implicit 3D spatial cognition:} understanding video as projections of a 3D world. Agents must know what is present, where, how things relate, and how configurations change over time. Today's video models remain limited here.
    \item \textbf{Predictive world modeling:} the brain makes \emph{unconscious inferences}~\cite{von1867handbuch} by predicting latent world states based on prior expectations. When these predictions are violated, surprise guides attention, memory, and learning~\cite{friston2010free,stahl2015observing,kennedy2024prediction}. However, current multimodal systems lack an internal model that anticipates future states and uses surprise to organize perception for memory and decision making.
\end{enumerate}

Our paper unfolds in three parts.
\textbf{First} (\S~\ref{sec:benchmark}), we re-examine existing benchmarks through the lens of our supersensing hierarchy. We find that most benchmarks map to the first few stages, while some, such as VSI-Bench~\cite{yang2024think}, begin to probe spatial reasoning. However, none sufficiently address the final crucial stage of predictive world modeling.
To make this gap concrete and motivate a shift in approach, we introduce \vsisuper{} (VSI stands for \textit{visual-spatial intelligence}), a two-part benchmark for spatial supersensing: \vsisuper{} Recall (\texttt{\vso}) targets long-horizon spatial observation and recall, while \vsisuper{} Count (\texttt{\vsc}) tests continual counting across changing viewpoints and scenes.
Built from arbitrarily long spatiotemporal videos, these tasks are deliberately resistant to the predominant multimodal recipe; they require perception to be \emph{selective} and \emph{structured} rather than indiscriminately accumulated.
We show that even the best long-context commercial models struggle on \vsisuper.

\textbf{Second} (\S~\ref{sec:limits}), we investigate whether spatial supersensing is simply a data problem. We curate \emph{\vsidata{}}, a spatially focused instruction-tuning corpus over images and videos, which we use to train \emph{\cambrianS}, a family of spatially-grounded video MLLMs.
Under the current paradigm, careful data design and training push \cambrianS{} to state-of-the-art spatial cognition on \textsc{VSI-Bench} (>$30\%$ absolute gain) without sacrificing general capabilities.
Nevertheless, \cambrianS{} still falls short on \vsisuper, indicating that while scale lays crucial groundwork, it alone is not sufficient for spatial supersensing.

This motivates the \textbf{third} and final part (\S~\ref{sec:predictive-sensing}), where we propose \emph{predictive sensing} as a first step toward a new paradigm.
We present a proof-of-concept solution built upon self-supervised next-latent-frame prediction.
Here, we leverage the model's prediction error, or ``surprise,'' for two key functions:
(1) managing memory by allocating resources to unexpected events, and
(2) event segmentation, breaking unbounded streams into meaningful chunks.
We demonstrate that this approach, though simple, significantly outperforms strong long-context baselines such as Gemini-2.5 on our two new tasks. Although not a final solution, this result provides compelling evidence that the path to true supersensing requires models that not only \emph{see} but actively predict and learn from the world.

Our work makes the following contributions.
(\textbf{1}) We define a hierarchy for spatial supersensing and introduce \vsisuper, a supersensing benchmark that reveals the limitations of the current paradigm.
(\textbf{2}) We develop \cambrianS, a state-of-the-art model that pushes the limits of spatial cognition. \cambrianS serves as a powerful new baseline, and, by delimiting the boundaries of current methods on our new benchmark, paves the path for a new paradigm.
(\textbf{3}) We propose predictive sensing as a promising new direction for MLLMs, showing that leveraging model surprise is more effective for long-horizon spatial reasoning than passive context expansion.


\section{Benchmarking Spatial Supersensing}\label{sec:benchmark}
\begin{figure}[t]
    \centering
    \includegraphics[width=1.0\linewidth]{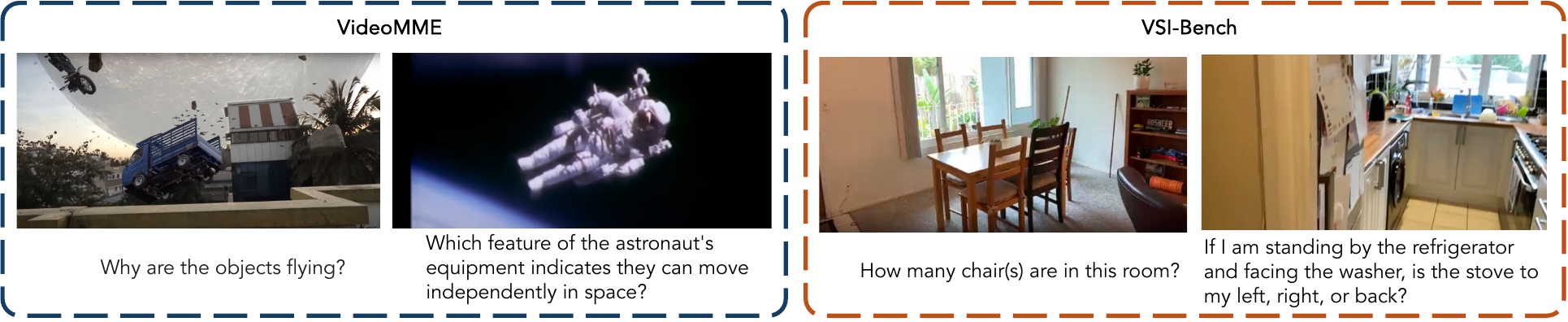}
    \caption{\textbf{Illustrations of how spatial sensing is conceptualized in current video benchmarks}. The left panel features examples from the ``spatial reasoning'' subcategory of VideoMME~\cite{fu2025video}, including a question regarding gravity from Shutter Authority's ``\texttt{What if the Moon Crashed into the Earth?}'' and a question regarding astronaut gear from NASA's ``\texttt{Astronaut Bruce McCandless II Floats Free in Space.}'' In contrast, the right panel shows samples from VSI-Bench~\cite{yang2024think}, which highlight visual-spatial reasoning tasks such as object counting, identifying relative directions, route planning, and more.
    }
    \label{fig:benchmark}
\end{figure}

To ground our pursuit of spatial supersensing, we first establish how to measure it. This section undertakes a two-part investigation into benchmarking this capability. We begin by auditing a suite of popular video MLLM benchmarks, where our analysis (\cref{fig:benchmark}) reveals that they overwhelmingly focus on linguistic understanding and semantic perception while neglecting the more advanced spatial and temporal reasoning
required for supersensing (\cref{sec:benchmark:deconstructing}). 
To address this critical gap, we then introduce \vsisuper, a new benchmark specifically designed to probe these harder, continual aspects of spatial intelligence in arbitrarily long streaming scenarios (\cref{sec:benchmark:vsi-super}).
We use this benchmark to test the limits of the current MLLM paradigm throughout the rest of the paper.

\begin{figure}[t]
    \centering
    \begin{overpic}[width=\linewidth]{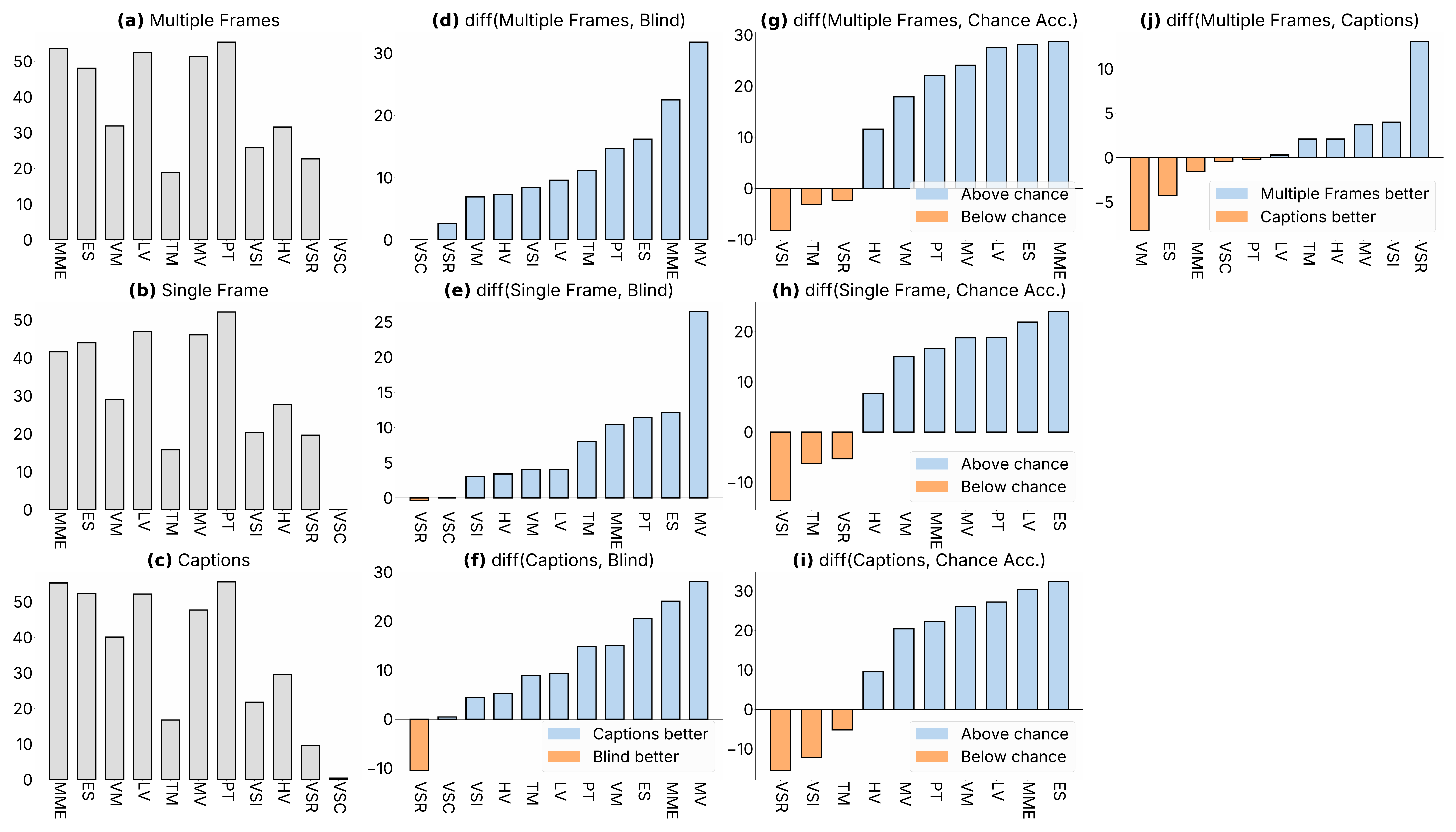}
      \put(100,20){%
        \makebox[0pt][r]{%
          \fcolorbox{white}{white}{%
            \parbox{0.23\linewidth}{\footnotesize
      \begin{tabular}{@{}l@{\hspace{2pt}}l@{}}
        \textbf{MME}: & VideoMME~\cite{fu2025video} \\
        \textbf{ES}:  & EgoSchema~\cite{mangalam2023egoschema} \\
        \textbf{VM}:  & VideoMMMU~\cite{hu2025video} \\
        \textbf{LV}:  & LongVideoBench~\cite{wu2024longvideobench} \\
        \textbf{TM}:  & Tomato~\cite{shangguan2024tomato} \\
        \textbf{MV}:  & MVBench~\cite{li2024mvbench} \\
        \textbf{PT}:  & Perception Test~\cite{patraucean2023perception} \\
        \textbf{HV}:  & HourVideo~\cite{chandrasegaran2024hourvideo}  \\
        \textbf{VSI}: & VSIBench~\cite{yang2024think} \\
        \textbf{\vso}: & \vsisuper Rec. \\
        \textbf{\vsc}: & \vsisuper Cnt. \\ 
      \end{tabular}
    }
            }}}
    \end{overpic}
    \caption{
        \textbf{Benchmark diagnostic results reveal varying dependence on visual input.}
        We evaluate model under distinct input conditions:
        (\textbf{a}) multiple (32) uniformly sampled frames,
        (\textbf{b}) a single (middle) frame, and
        (\textbf{c}) frame captions, benchmarked against chance-level and blind test results (visual input ignored).
        Panels (\textbf{a}--\textbf{c}) show absolute accuracies; panels (\textbf{d}--\textbf{j}) show performance differences between conditions.
        Visual inputs are substantially more critical for VSI-Bench~\cite{yang2024think}, Tomato~\cite{shangguan2024tomato}, and HourVideo~\cite{chandrasegaran2024hourvideo}, while their impact is less pronounced for VideoMME~\cite{fu2025video}, MVBench~\cite{li2024mvbench}, and VideoMMMU~\cite{hu2025video}.
        \vso{} and \vsc{} are new supersensing benchmarks introduced in \cref{sec:benchmark:vsi-super}.
    }\label{fig:benchmark_analysis}
    \vspace{0.25em}
\end{figure}

\subsection{Deconstructing Existing Video Benchmarks}\label{sec:benchmark:deconstructing}

Recent advances in MLLMs have led to a surge of Video-QA benchmarks. However, a critical question remains: \emph{to what extent do existing video benchmarks truly examine visual sensing capabilities rather than simply testing language priors?} Our diagnostic tests disentangle the model's reliance on visual sensing versus linguistic priors by varying the richness of visual input and the informativeness of textual cues. Benchmarks solvable with text-only inputs (\eg, captions or a blind MLLM) are skewed towards examining linguistic understanding. In contrast, benchmark questions that can only be answered with multi-frame inputs require genuine visual sensing.
We use an image-based multimodal large language model Cambrian-1~\cite{tong2024cambrian} for evaluation, which allows us to probe the underlying task demands without conflating them with the capabilities of video-specific architectures and post-training recipes.

\noindent We establish several experimental conditions for feeding video input to a Cambrian-1~\cite{tong2024cambrian} model:
\begin{itemize}
    \item \textbf{\multipleul{Multiple} Frames}: The model processes 32 frames uniformly sampled from the video clip. This is the standard method for representing video input in the literature~\cite{li2024llava}.
    \item \textbf{\singleul{Single} Frame}: The model processes only the middle frame of a given video clip. This condition tests the reliance on minimal, contextually-central visual information.
    \item \textbf{Frame \captionul{Captions}}: Instead of video frames, the model receives captions corresponding to the same 32 uniformly-sampled frames.
    This condition is designed to reveal how solvable a task is \emph{without} low-level perceptual grounding.
    We use the Gemini-2.0-Flash API to re-caption video frames.
\end{itemize}
To contextualize the performance under these conditions, we introduce two other baselines:
\begin{itemize}
    \item \textbf{\blindul{Blind} Test}:
        The model attempts the task using solely the task's question. \emph{All visual input is ignored}, no visual captions are used.
        This baseline measures the model's performance based on its pre-existing knowledge, language priors, and any potential biases in the benchmark questions. 
    \item \textbf{\chanceul{Chance} Acc}: This represents the accuracy achievable by randomly guessing for the specific task format (\eg, multiple-choice questions), serving as a floor for performance.
\end{itemize}

\noindent We conduct a fine-grained analysis of each benchmark's characteristics by comparing performance across these conditions and baselines. We focus on the following key comparisons (\texttt{diff(A,B)} $=$ \texttt{A-B}): 
\begin{itemize}
    \item \texttt{diff($\bf x$, \blindul{Blind})},\ $\bf x \in \big\{$\multipleul{Multiple}, \singleul{Single}, \captionul{Captions}$\big\}$ 
        to quantify the uplift provided by different input modalities over the blind baseline; 
    \item \texttt{diff($\bf x$, \chanceul{Chance})},\ $\bf x \in \big\{$\multipleul{Multiple}, \singleul{Single}, \captionul{Captions}$\big\}$ 
        to measure performance gains over chance; 
        \item \texttt{diff(\multipleul{Multiple}, \captionul{Captions})}
            to understand the performance gap between the current mainstream practice and a strong language-only baseline
\end{itemize}

\noindent
Results presented in \cref{fig:benchmark_analysis} (a-c) demonstrate that Cambrian-1~\cite{tong2024cambrian}, an \singleul{image-based MLLM} without any video post-training, can attain reasonable performance across many benchmarks, in some instances surpassing \chanceul{chance-level} accuracy by 10-30\% (see \cref{fig:benchmark_analysis}-g,h). This suggests that much of the knowledge these benchmarks target is accessible via standard single-image instruction-tuning pipelines. 
Nevertheless, on two existing datasets, VSI-Bench~\cite{yang2024think} and Tomato~\cite{shangguan2024tomato}, the model's performance falls below chance-level. For VSI-Bench, this is largely because its spatial understanding questions require true video sensing and targeted data curation and training. For Tomato, this underperformance is expected: the benchmark demands understanding of fine-grained details from higher frame-rate video, rendering the largely temporally-subsampled single-frame and 32-frame inputs inadequate.

Employing \captionul{textual captions} in place of visual inputs also yields notable performance improvements, surpassing \chanceul{chance accuracy} by more than 20\% on benchmarks such as EgoSchema~\cite{mangalam2023egoschema}, VideoMME~\cite{fu2025video}, LongVideoBench~\cite{wu2024longvideobench}, VideoMMMU~\cite{hu2025video}, Perception Test~\cite{patraucean2023perception}, and MVBench~\cite{li2024mvbench} (\cref{fig:benchmark_analysis}-i). Similar conclusions can be drawn when comparing benchmark performance against \blindul{blind} test results (\cref{fig:benchmark_analysis}-d,f). Such performance implies that these benchmarks primarily probe abilities inferable from textual summaries of video content.
Interpreting the performance difference between using ``\multipleul{multiple frames}'' and ``\captionul{frame captions}'' (\cref{fig:benchmark_analysis}-j), a significantly positive margin (in favor of \multipleul{multi-frame inputs}) signifies a benchmark's demand for nuanced visual sensing. Conversely, a small or negative margin (more in favor ``\captionul{frame captions}'') suggests a more language-centric nature. Our analysis places VideoMMMU, EgoSchema, VideoMME, Perception Test, and LongVideoBench in this latter category, indicating their potential reliance on \emph{linguistic understanding} rather than visual cues. A notable exception is VSC, which is so challenging for current MLLMs that all three input conditions yield near-zero performance, precluding any meaningful comparison between them.

\finding{}{\faBookmark~~Existing benchmarks overwhelmingly focus on linguistic understanding and semantic perception while neglecting the more advanced spatial and temporal reasoning required for supersensing.}

We hope to emphasize the inherent challenges in benchmarking and the impracticality of creating a single, all-encompassing benchmark to evaluate every capability. For example, reliance on language priors should not be viewed merely as a drawback, as access to rich world knowledge and its effective retrieval is undoubtedly beneficial in many scenarios. We argue that \textbf{video benchmarks should not be treated as measuring a single, uniform notion of ``video understanding.'' Instead, their design and evaluation should be grounded in the specific capabilities they aim to assess.} The preceding analyses are therefore intended to guide the development of tasks that more effectively drive progress towards \emph{spatial supersensing}, which will be the central focus of the rest of the paper.

\subsection{\vsisuper: Towards Benchmarking Spatial Supersensing in Multimodal LLMs}\label{sec:benchmark:vsi-super}
Referring to \cref{fig:teaser}, spatial supersensing requires MLLMs to have four key capabilities: \emph{semantic perception}, \emph{streaming event cognition}, \emph{implicit 3D spatial cognition}, and \emph{predictive world modeling}. However, as outlined by our analysis in \cref{fig:benchmark_analysis}, most existing video QA benchmarks mainly evaluate the linguistic understanding and semantic perception aspects, which are more reactive and driven by specific tasks~\cite{fu2025video,mangalam2023egoschema,hu2025video}. While recent research has begun to address streaming event cognition through continual sensing, memory architectures, and proactive answering ~\cite{chen2024videollm,qian2025dispider,niu2025ovo,wu2024streambench,song2024moviechat,zhang2024flash}, this capability is often engineered at test time rather than being a native model skill. 
Furthermore, although spatial reasoning occasionally appears as a category in existing benchmarks, these tasks seldom reach the level of true spatial cognition, and are far from probing the world-modeling capacity that defines supersensing (\cref{fig:benchmark}).
Although VSI-Bench~\cite{yang2024think} takes an initial step toward examining spatial cognition, its videos remain short-form and single-scene, and it neither formalizes the problem nor evaluates the essential capability of predictive modeling of the world.

To illuminate the gap between current MLLMs and spatial supersensing, we introduce \vsisuper, a two-part benchmark for continual spatial sensing. The tasks are intuitive and generally easy for humans, where one simply watches and keeps track of what happens, but they remain surprisingly challenging for machines. They demand selective filtering and structured accumulation of visual information across unbounded spatial videos to maintain coherent understanding and answer questions. Importantly, they are resistant to brute-force context expansion, exposing the need for true spatial reasoning. We detail the two components below.

\begin{figure}[h]
    \centering
    \includegraphics[width=0.95\linewidth]{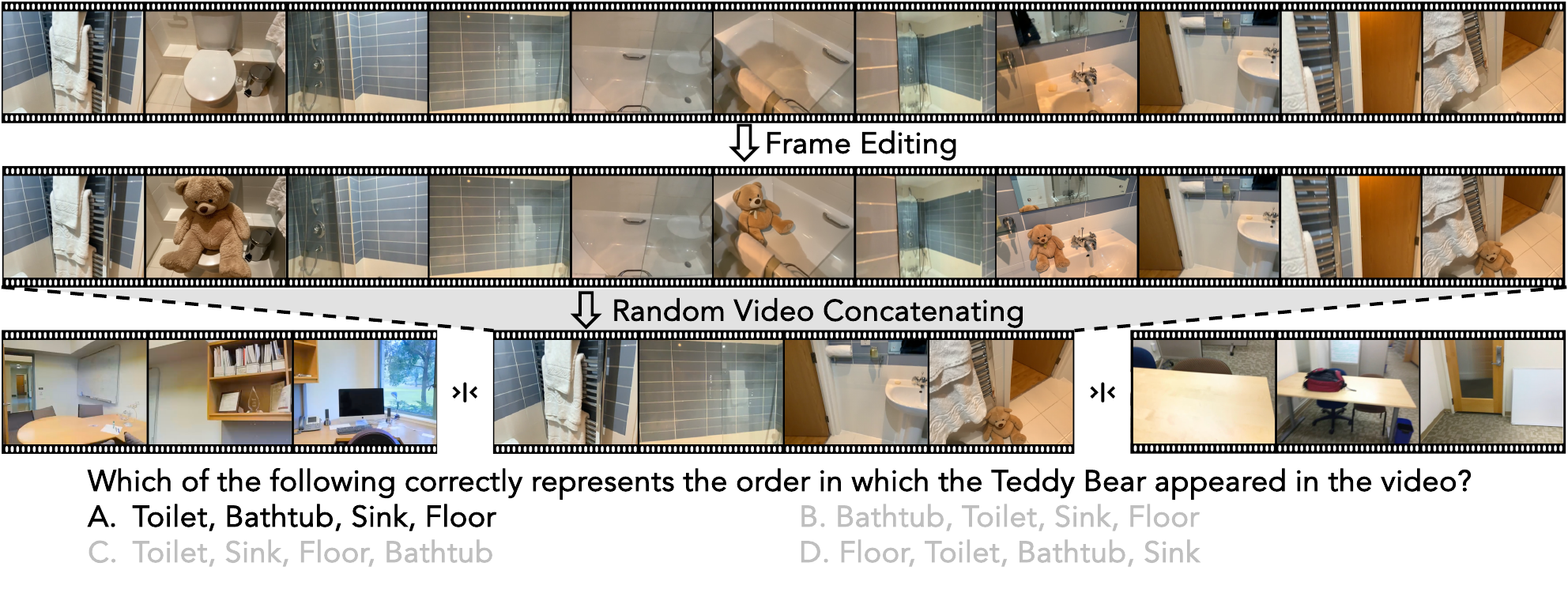}
    \caption{\small \textbf{Illustration of the \vso benchmark's construction process and format}. We use generative models to edit videos by inserting \emph{surprising} or out-of-place objects into the space. The core task then challenges models to recall the spatial placements of these objects in the correct order of their appearance across arbitrarily long videos.}
    \label{fig:sor_benchmark_illustration}
\end{figure}

\paragraph{\vsisuper Recall: Long-horizon spatial observation and recall.}
The \vso benchmark requires MLLMs to observe long-horizon spatiotemporal videos, and sequentially recall the locations of an unusual object.
As shown in \cref{fig:sor_benchmark_illustration}, to construct this benchmark, human annotators use an image editing model (\ie, Gemini~\cite{comanici2025gemini}) to insert surprising or out-of-place objects (\eg, a Teddy Bear) into four distinct frames (and spatial location) of a video capturing a walkthrough of an indoor environment~\cite{dai2017scannet,yeshwanth2023scannet++,dehghan2021arkitscenes}. This edited video is then concatenated with other similar room-tour videos to create an arbitrarily long and continuous visual stream. This task parallels the needle-in-a-haystack (NIAH) test commonly used in the language domain to stress test the long-context capabilities of LLMs~\cite{liu2023lost}. Similar NIAH setups have also been proposed for long-video evaluation~\cite{zhao2024needle,wei2025videorope,hu2025nemo}. However, unlike benchmarks that insert unrelated text segments or frames, \vso preserves the realism of the ``needle'' through in-frame editing. It further extends the challenge by requiring sequential recall, effectively a multi-hop reasoning task, and remains arbitrarily scalable in video length. To thoroughly evaluate model performance across different time scales, the benchmark is provided in five durations: 10, 30, 60, 120, and 240 minutes. Further details on the \vso benchmark construction are provided in \cref{appx:vsisuper}.

\begin{figure}[h]
    \centering
    \includegraphics[width=\linewidth]{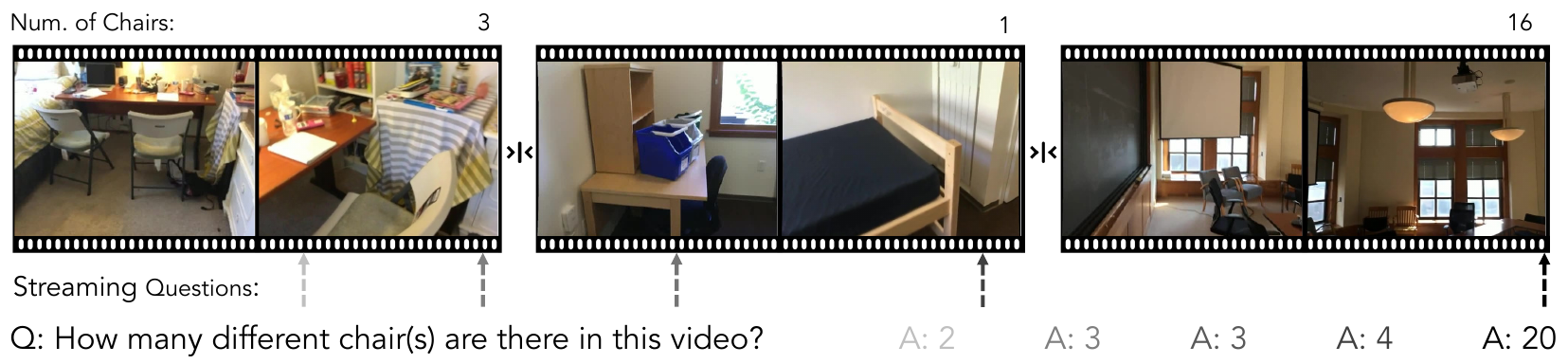}
    \caption{
        \small
        \textbf{Overview of the \vsc{} benchmark.}
        The benchmark evaluates counting capabilities on long-horizon, multi-room videos composed of concatenated scenes.
        Queries are posed at various time points to simulate a streaming question-answering setting.
    }\label{fig:mem-count-illustration}
\end{figure}

\paragraph{\vsisuper Count: Continual counting under changing viewpoints and scenes.}
Here we test the capacity of MLLMs to continuously accumulate information in long-form spatial videos.
To build \vsc, we concatenate multiple room-tour video clips from VSI-Bench~\cite{yang2024think} and task models with counting the \emph{total} number of target objects across all rooms (see \cref{fig:mem-count-illustration}).
This setting is challenging because the model must handle viewpoint shifts, repeat sightings, and scene transitions, all while maintaining a consistent cumulative count. For humans, counting is an intuitive and generalizable process. Once the concept of ``one'' is understood, extending it to larger quantities is natural. In contrast, as we later demonstrate, current MLLMs lack true spatial cognition and depend excessively on learned statistical patterns.

In addition to standard evaluations (\ie, ask question at the end of video), we query the model at multiple timestamps to assess its performance in streaming settings, where the correct answer in \vsc evolves dynamically over time. To examine long-term consistency, \vsc includes four video durations: 10, 30, 60, and 120 minutes. For this quantitative task, we report results using the mean relative accuracy ($\mathcal{MRA}$) metric, consistent with the VSI-Bench evaluation protocol~\cite{yang2024think}.

\begin{figure}[!h]
    \centering
    \includegraphics[width=0.875\linewidth]{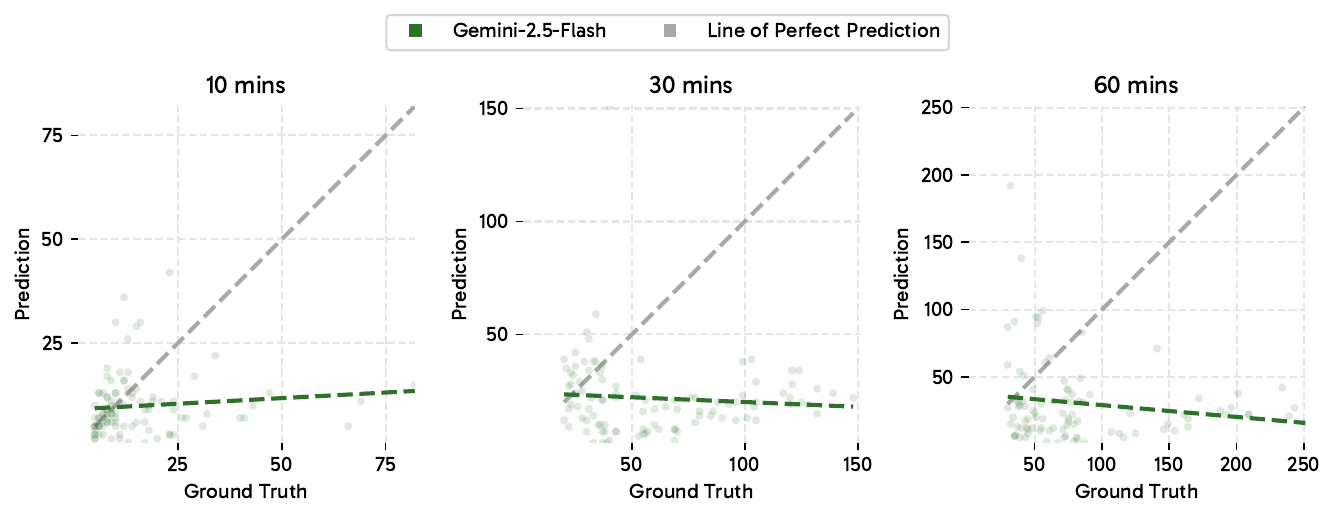}
    \caption{
        \textbf{Visualization of Gemini-2.5-Flash's predictions \emph{v.s.} ground truth on \vsc.} The model's predicted object counts saturate at small constant values and fail to scale with video length or true object counts, indicating limited generalization in counting and reliance on training distribution priors.
    }\label{fig:gemini_vsc_prediction_distribution}
\end{figure}

\paragraph{State-of-the-art models struggle on \vsisuper.}
To test whether \vsisuper{} poses a real challenge for frontier MLLMs, we evaluate the latest Gemini-2.5-Flash~\cite{team2024gemini}. As shown in \cref{tab:gemini_2.5_flash_results}, the model reaches its context limit when handling two-hour videos, despite a context length of 1,048,576 tokens. This highlights the \emph{open-ended} nature of video understanding, where continuous streams effectively require an ``infinite-in, infinite-out'' context and can grow arbitrarily long, suggesting that simply scaling up tokens, context length, or model size may not suffice. Though synthetic, our benchmark reflects a real challenge in spatial supersensing: humans effortlessly integrate and retain information from ongoing sensory experiences that unfold over hours or years, yet current models lack comparable mechanisms for sustained perception and memory. 
Gemini-2.5-Flash demonstrates strong performance on semantic-perception and linguistic-understanding-focused video benchmarks such as VideoMME~\cite{fu2025video} and VideoMMMU~\cite{hu2025video}, achieving around 80\% accuracy. However, even for 60-minute videos in \vsisuper that fall well within its context window, performance on \vso and \vsc remains limited---only 41.5 and 10.9, respectively. As shown in \cref{fig:gemini_vsc_prediction_distribution}, the model's predicted object counts fail to scale with video length or the true number of objects, instead saturating at a small constant value, suggesting a lack of generalization in counting ability and a reliance on training distribution priors.

\begin{table}[h]
    \centering
    \resizebox{0.95\textwidth}{!}{
    \begin{tabular}{c|ccc cc cc}
    \multirow{2}{*}{Model} & \multirow{2}{*}{VideoMME\cite{fu2025video}} & \multirow{2}{*}{VideoMMMU\cite{hu2025video}} & \multirow{2}{*}{VSI-Bench\cite{yang2024think}} & \multicolumn{2}{c}{\vso} & \multicolumn{2}{c}{\vsc}\\
     & & & & 60 min & 120 min & 60 min & 120 min \\
    \hline
    Gemini-2.5-Flash & \cellcolor{green!15}81.5 & \cellcolor{green!15}79.2 & \cellcolor{red!5}45.7 & \cellcolor{red!15}41.5 & \cellcolor{red!35}Out of Ctx. & \cellcolor{red!15}10.9 & \cellcolor{red!35}Out of Ctx. \rule{0pt}{10pt} \\
    \end{tabular}}
    \caption{\textbf{Gemini-2.5-Flash results.} As a state-of-the-art video understanding model with long-context capabilities, Gemini demonstrates strong performance on general video benchmarks but shows clear limitations towards spatial supersensing.}
    \label{tab:gemini_2.5_flash_results}
\end{table}

\paragraph{How \vsisuper challenges the current paradigm.}
Although the task setup is simple, the challenge posed by \vsisuper goes beyond just spatial reasoning and reveals fundamental limitations of the current MLLM paradigm.

\finding{}{\faBookmark~~\vsisuper tasks challenge the belief that scaling alone guarantees progress.}

By allowing arbitrarily long video inputs that emulate the dynamics of streaming cognition, \vsisuper is intentionally constructed to exceed any fixed context window. This design suggests that frame-by-frame tokenization and processing are unlikely to be computationally viable as a long-term solution.
Humans address such problems efficiently and adaptively by selectively attending to and retaining only a small fraction of sensory input\footnote{Each eye's 6 million cone photoreceptors can send about 1.6 Gbits/s, yet the brain uses only 10 bits/s to guide behavior~\cite{koch2006much,zheng2025unbearable}.}, often unconsciously~\cite{fei2007we,von1867handbuch}. This predictive and selective mechanism, core to human cognition, remains absent in current MLLMs but is fundamental to a predictive world model.

\finding{}{\faBookmark~~\vsisuper tasks demand generalization to new temporal and spatial scales at test time.}
For example, \vsc requires counting in arbitrarily long videos, similar to how humans, who understand the concept of counting, can extend it to any number. The key is not maintaining an extremely long context window, humans do not retain every visual detail from extended visual experiences, but rather learning the process of counting itself. Predictive sensing facilitates this by \emph{segmenting} continuous visual streams into coherent events, using moments of ``surprise'' to impose temporal structure. This segmentation acts as a divide-and-conquer mechanism that allows the model to decide when to start, continue, or reset behaviors in dynamically changing scenes. 

Together, these challenges, which span computational efficiency, generalization, and cognitive mechanisms such as unconscious inference and predictive sensing, call for a paradigm shift. Rather than relying solely on scaling data, parameters, or context length, future models should learn internal world models capable of perceiving and predicting within an endlessly unfolding visual world across space and time.

To further motivate this paradigm shift, the next section investigates the extent to which progress remains possible within the current paradigm through improved engineering and targeted data curation. We assess whether the existing MLLM framework can be adapted to address the challenges posed by \vsisuper. These efforts, while operating within the limits of the present framework, are indispensable for building the data and empirical foundations of the next generation of spatial supersensing models.


\section{Spatial Sensing Under the Current Paradigm}\label{sec:limits}
As demonstrated in the previous section, Gemini-2.5-Flash exhibits subpar performance on spatial sensing tasks (see \cref{tab:gemini_2.5_flash_results}). This observation raises a key question: \emph{Is limited spatial sensing simply a data issue?} It is a valid question to ask, as current video MLLMs do not explicitly prioritize spatial-focused videos during training, and it remains s whether existing pre-training and post-training designs are well-suited for our target tasks. We begin by enhancing Cambrian-1~\cite{tong2024cambrian} with a series of architectural and training improvements to establish a stronger image MLLM as our base model (\cref{sec:limits:upgrade-cambrian-1}). We proceed to construct a large-scale, spatial-focused instruction-tuning dataset, \vsidata (\cref{sec:limits:data}). The dataset is curated from diverse sources and carefully annotated. As such data does not currently exist publicly, \vsidata is intended to provide a strong data foundation for spatial sensing. Finally, with a refined training recipe (\cref{sec:limits:training_recipe}), we introduce the spatially-grounded \cambrianS model family (\cref{sec:limits:cambrian-s}).

The \cambrianS{} model family demonstrates strong performance on established spatial reasoning benchmarks such as VSI-Bench~\cite{yang2024think} and offers valuable insights into base model design, data curation, and training strategies for spatial supersensing. However, despite these advances, this approach does not directly address the continual sensing challenges of \vsisuper (\cref{sec:limits:resuls}); instead, it provides a crucial foundation that motivates the new paradigm introduced in (\cref{sec:predictive-sensing}).

\subsection{Base Model Training: Upgraded Cambrian-1}\label{sec:limits:upgrade-cambrian-1}
We begin by developing an image-based MLLM base model, as robust semantic perception forms the foundation for higher-level spatial cognition. We follow the two-stage training pipeline of Cambrian-1~\cite{tong2024cambrian}. We upgrade the visual encoder to SigLIP2-SO400m~\cite{tschannen2025siglip} and the language model to the instruction-tuned Qwen2.5~\cite{yang2024qwen2.5}.
For the vision-language connector, we adopt a simple two-layer MLP primarily for its computational efficiency. Other training components from Cambrian-1, including hyperparameters and the data recipe, remain unchanged. Full implementation details are provided in \cref{appx:impl_details}.

\subsection{Spatial Video Data Curation: \vsidata}\label{sec:limits:data}

\begin{figure}[h]
    \centering
    \includegraphics[width=1\textwidth]{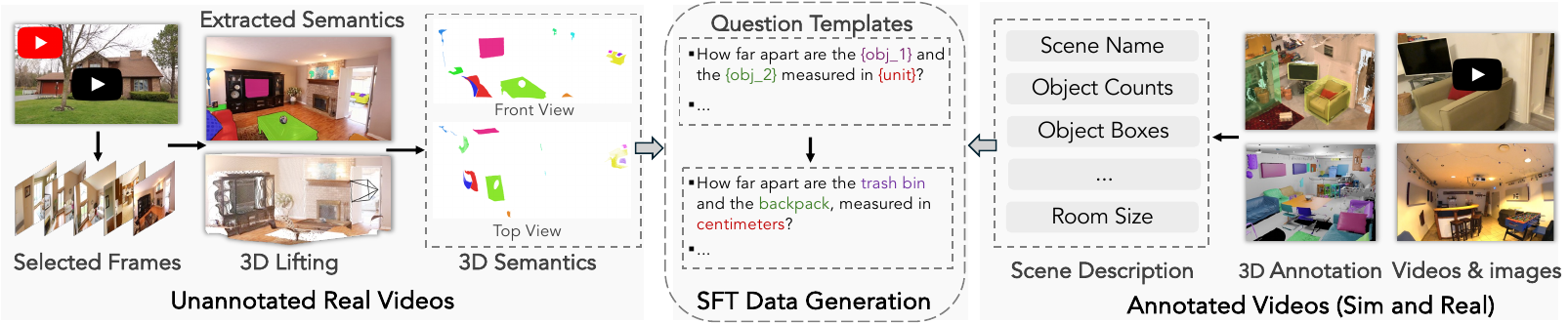}
    \caption{\textbf{\vsidata data curation pipeline.} We collect data from 3D-annotated real and simulated video sources, as well as from pseudo-annotated frames extracted from web videos. We then use diverse templates to automatically generate question–answer pairs for instruction tuning.}
    \label{fig:data_collect_pipeline}
\end{figure}

\begin{table}[h]
    \centering
    \caption{\textbf{Data statistics for \vsidata.}
    We collect data from 10 sources with different video types and annotations to improve diversity.}
    \label{tab:data_statistics}
    \scalebox{0.85}{
    \begin{tabular}{lrrr}
      \toprule
      \textbf{Dataset} & \textbf{\# Videos} & \textbf{\# Images} & \textbf{\# QA Pairs} \\
      \hline
      \rowcolor{navyblue!5} \multicolumn{1}{l}{\textcolor{black}{\textit{Annotated Real Videos}}} & & &  \\
      S3DIS~\cite{armeni20163d} & 199 & - & 5,187 \\
      Aria Digital Twin~\cite{pan2023aria} & 183 & - & 60,207 \\
      ScanNet~\cite{dai2017scannet} & 1,201 & - & 92,145 \\
      ScanNet++ V2~\cite{yeshwanth2023scannet++} & 856 & - & 138,701 \\
      ARKitScenes~\cite{dehghan2021arkitscenes} & 2,899 & - & 57,816 \\
      \hline
      \rowcolor{navyblue!5} \multicolumn{1}{l}{\textcolor{black}{\textit{Simulated Data}}} & & &  \\
      ProcTHOR~\cite{deitke2022ProcTHOR} & 625 & - & 20,092 \\
      Hypersim~\cite{roberts2021hypersim} & - & 5,113 & 176,774 \\
      \hline
      \rowcolor{navyblue!5} \multicolumn{1}{l}{\textcolor{black}{\textit{Unannotated Real Videos}}} & & &  \\
      YouTube Room Tour & - & 20,100 & 20,100 \\
      Open X-Embodiment~\cite{o2024open} & - & 14,801 &  14,801\\
      AgiBot-World~\cite{bu2025agibot} & - & 4,844 & 4,844 \\
      \midrule
      \textbf{Total} & 5,963 & 44,858 & 590,667 \\
      \bottomrule
    \end{tabular}}
\end{table}

It is well recognized that data quality and diversity play a critical role in the training of MLLMs~\cite{tong2024cambrian,mckinzie2024mm1}. We hypothesize that the performance gap on VSI-Bench~\cite{yang2024think} comes mainly from the lack of high-quality, spatially grounded data in current instruction-tuning datasets~\cite{zhang2024video,cui2025comprehensive}. To fill this gap, we build \textbf{\vsidata}, a large-scale instruction-tuning dataset designed to improve visual-spatial understanding.

\paragraph{Data curation and processing.}
We construct \vsidata from a diverse span of data sources and types (\ie, simulated and real). See \cref{tab:data_statistics} for the data sources and for dataset statistics on the number of videos, images, and QA pairs from each dataset. We find that this yields a dataset substantially more robust than one of comparable size derived from a single source. 
Below, we detail the data processing procedure.

\begin{itemize}[leftmargin=16pt,topsep=0pt,itemsep=0.1em]

    \item \emph{Annotated real videos.}
        Multimodal visual–spatial reasoning relies on a solid understanding of 3D geometry and spatial relationships. Following VSI-Bench, we repurpose the training splits of existing indoor scan and first-person video datasets that provide 3D instance-level annotations, including S3DIS~\cite{armeni20163d}, ScanNet~\cite{dai2017scannet}, ScanNet++~V2~\cite{yeshwanth2023scannet++}, ARKitScenes~\cite{dehghan2021arkitscenes}, and ADT~\cite{pan2023aria}. For each dataset, annotations are consolidated into a meta-information file capturing scene-level attributes such as object counts by category, object bounding boxes, room dimensions, and related metadata. Question templates are then automatically instantiated to generate corresponding questions.
    \item \emph{Simulated data.}
        Due to the limited availability of 3D-annotated data, constructing a large-scale and diverse 3D-annotated SFT dataset solely from real annotated videos is challenging. Following SIMS-V~\cite{brown2025simsv}, we utilize embodied simulators to procedurally generate spatially grounded video trajectories and QA pairs, rendering 625 video traversals within ProcTHOR~\cite{deitke2022ProcTHOR} scenes featuring diverse layouts, object configurations, and visual appearances.
        We apply the same methodology to Hypersim~\cite{roberts2021hypersim}, sampling 5,113 images from 461 indoor scenes. Using instance-level bounding boxes, we generate question-answer pairs consistent with our annotated real-video setup.
    
    \item \emph{Unannotated real videos.}
        Although web-sourced videos lack explicit annotations, they offer rich diversity in indoor environment types, geographical regions, and spatial layouts. We collected approximately 19K room tour videos from YouTube and additionally incorporated videos from robotic learning datasets, including Open-X-Embodiment~\cite{o2024open} and AgiBot-World~\cite{bu2025agibot}. Since these videos do not contain the 3D annotations required for constructing spatial instruction-tuning data, we develop a pseudo-annotation pipeline. As illustrated in \cref{fig:data_collect_pipeline}, we subsample and filter video frames, applying object detection~\cite{liu2024grounding}, segmentation model~\cite{ravi2024sam}, and 3D reconstruction model~\cite{wang2025vggt} to generate pseudo-annotated images following the approach of SpatialVLM~\cite{chen2024spatialvlm}. We choose to generate annotations at the image level rather than across full videos, as full-video pseudo-annotations derived from recognition and reconstruction models tend to be too noisy for training.
\end{itemize}

\paragraph{Question type definition and template augmentation.}
We define 12 question types within a spatiotemporal taxonomy to construct a comprehensive and diverse set of questions for instruction tuning. We define five main question types---\texttt{size}, \texttt{direction}, \texttt{count}, \texttt{distance}, and \texttt{appearance order}---broadly categorized as measuring \texttt{configuration}, \texttt{measurement}, or \texttt{spatiotemporal} capabilities following~\cite{yang2024think}.
Except for the \texttt{appearance order} type, each question category includes both relative and absolute variants, reflecting the importance of these complementary forms of reasoning in visual–spatial understanding~\cite{yang2024think}. For example, for \texttt{size}, we ask for both size comparison between two objects (\textit{relative}) and the metric dimensions of an object (\textit{absolute}).
To enhance diversity, we vary the perspective used in formulating \texttt{direction} and \texttt{distance} questions. For instance, a \texttt{distance} question may ask which of two objects is closer to the camera or which object is closer to a third reference object. We also diversify the dataset through variations in question wording and in measurement units (\emph{e.g.}, meters versus feet). Additional details of the dataset are provided in \cref{appx:vsidata}.

\begin{table}[t]
    \caption{\textbf{Contributions of Different Data Sources in the \vsidata Mixture.} This table illustrates the impact of different data sources on VSI-Bench performance. The combined dataset, \vsidata Full Mix, achieves the best overall results. Among individual sources, annotated real video datasets contribute the most significant improvements, followed by simulated videos, and then pseudo-annotated images.
    }\label{tab:vsidata_source_ablation}
    \centering
    \setlength\tabcolsep{2pt} 
    \setlength{\tabcolsep}{0.5em}
    \resizebox{0.85\textwidth}{!}{
    \begin{tabular}{l |ccc |c cccc cccc}
        \toprule
        &
            \multicolumn{3}{c|}{Image} &
            \multicolumn{9}{c}{VSI-Bench (Video)} \\
        VSI Data Mixture
        &
            \rotatebox{90}{MMVP} &
            \rotatebox{90}{3DSR} &
            \rotatebox{90}{CV-B} &
            \rotatebox{90}{Avg} &
            \rotatebox{90}{Obj Ct} &
            \rotatebox{90}{Abs Dst} &
            \rotatebox{90}{Obj Sz} &
            \rotatebox{90}{Rm Sz} &
            \rotatebox{90}{Rel Dst} &
            \rotatebox{90}{Rel Dir} &
            \rotatebox{90}{Rte Pln} &
            \rotatebox{90}{Ap Ord} \\
        \hline
        \rowcolor{gray!10}
        \rule{0pt}{10pt} 
        Baseline & 52.7 & 54.5 & 73.5 & 28.5 & 18.1 & 20.0 & 36.0 & 22.2 & 42.9 & 31.3 & 24.6 & 33.0 \\
        \hline
        \rowcolor{gray!10!blue!5}\textit{Real Videos} & & & & & & & & & & & & \\
        + S3DIS & 54.0 & 54.9 & 75.3 & 41.6 & 63.8 & 21.0 & 44.9 & 37.0 & 43.8 & 47.4 & 34.0 & 41.1 \\
        + ADT & 50.6 & 56.5 & 77.5 & 41.0 & 51.0 & 29.8 & 52.5 & 40.2 & 42.3 & 38.8 & 34.0 & 39.8 \\
        + ARKitScenes & 50.0 & 56.7 & 77.3 & 51.0 & 70.2 & 32.7 & 64.5 & 60.0 & 55.1 & 45.2 & \textbf{37.1} & 43.5 \\
        + ScanNet & 54.7 & \textbf{57.7} & 77.5 & 56.3 & 70.9 & 37.9 & 67.5 & 59.3 & 57.0 & 46.7 & 35.1 & 76.1 \\
        + ScanNet++ V2 & 52.7 & 57.3 & 77.5 & 56.3 & 72.5 & 40.7 & 65.7 & 56.9 & 59.7 & 47.1 & 31.4 & 76.2 \\
        \hline
        \rowcolor{gray!10!blue!5}\textit{Simulated Videos} & & & & & & & & & & & & \\
        + ProcThor & 53.3 & 55.7 & 74.9 & 36.4 & 21.0 & 29.7 & 49.3 & 3.8  & 52.3 & 45.7 & 30.4 & 58.7 \\
        + HyperSim & 52.0 & 56.0 & \textbf{79.7} & 45.6 & 67.8 & 32.0 & 59.3 & 36.4 & 53.2 & 47.0 & 32.5 & 36.6 \\
        \hline
        \rowcolor{gray!10!blue!5}\textit{Pseudo-Annotated Images} & & & & & & & & & & & & \\
        + YTB RoomTour & 55.3 & 52.6 & 75.0 & 32.5 & 43.4 & 25.8 & 24.2 & 27.3 & 38.7 & 31.4 & 28.4 & 40.9 \\
        + OXE \& AGIBot & \textbf{56.0} & 54.4 & 72.5 & 30.6 & 40.3 & 23.1 & 27.9 & 26.6 & 38.0 & 22.8 & 32.0 & 33.8 \\
        \hline
        \rowcolor{green!5}
        \textbf{Full Mix} & 54.7 & 54.0 & 77.9 & \textbf{63.2} & \textbf{73.5} & \textbf{49.4} & \textbf{71.4} & \textbf{70.1} & \textbf{66.9} & \textbf{61.5} & 36.6 & \textbf{76.6} \rule{0pt}{10pt} \\ 
        \bottomrule
    \end{tabular}}
\end{table}
\paragraph{\vsidata data source ablation.}
To evaluate the effectiveness of our proposed \vsidata dataset, we perform an ablation study by finetuning the improved Cambrian-1 MLLM described in ~\cref{sec:limits:upgrade-cambrian-1} with part of the video instruction tuning samples from LLaVA-Video-178K~\cite{zhang2024video}. This model serves as the \textit{baseline} in \cref{tab:vsidata_source_ablation}. 
The contribution of each data source is evaluated by fine-tuning the model on individual datasets as well as their combination. The \vsidata Full Mix achieves the highest overall performance on video spatial reasoning tasks, outperforming both the baseline and all single-source counterparts. \textbf{All data sources contribute positively} after fine-tuning, though their effectiveness varies. 
\finding{}{\faBookmark~~Data effectiveness ranks as: annotated real videos $>$ simulated data $>$ pseudo-annotated images.}
This indicates that videos are inherently more informative than static images for spatial reasoning, as training exclusively on video data yields superior performance on \emph{both} video- and image-based spatial reasoning benchmarks. These findings support the intuition that the temporal continuity and multi-view diversity of videos are key to developing robust spatial representations.

\subsection{Post-Training Recipe for Spatial Sensing}\label{sec:limits:training_recipe}

We further analyze and ablate our video instruction-tuning pipeline, focusing on the roles of the pretrained base video model and the instruction-tuning dataset mixture. As shown in \cref{tab:training_pipeline}, we begin with four \emph{base models} that represent a progressive increase in video understanding capability:

\begin{itemize}[nosep, leftmargin=*]
    \item \textbf{A1} is trained only with image-text alignment on Cambrian-1 alignment data. The language model is identical to base QwenLM as it is frozen during training.
    \item \textbf{A2} is finetuned with image instruction tuning on top of A1, essentially our improved Cambrian-1.
    \item \textbf{A3} is initialized from A2 and finetuned on 429K video instruction tuning data.
    \item \textbf{A4} is initialized from A2 and finetuned on 3M video instruction tuning data.
\end{itemize}

We then finetune these models using two different data recipes:
(1) \vsidata only, and (2) \vsidata mixed with a similar amount of general video instruction tuning data.

\finding{}{\faBookmark~~A stronger base model with greater exposure to general video data leads to improved spatial sensing after SFT.}
As shown in \cref{tab:training_pipeline}, SFT with a stronger base model, one that performs well on general video benchmarks such as VideoMME~\cite{fu2025video} and EgoSchema~\cite{mangalam2023egoschema}, leads to enhanced spatial understanding. This highlights the importance of broad exposure to general video data during base model training. 

\finding{}{\faBookmark~~Mixing general video data prevents the generalization loss caused by in-domain SFT.}
Furthermore, while in-domain SFT solely on \vsidata achieves the highest performance on VSI-Bench, it results in a noticeable decline on general video benchmarks. However, this performance drop can be effectively mitigated by training on a data mix that includes general videos.

\begin{table}[tbp]
    \centering
    \caption{
    \textbf{Post-training exploration for spatial sensing.} We examine four base models with progressively increasing exposure to visual data, from image-only training to extensive video training, and analyze their distinct trends during spatial sensing tuning under two different data recipes. \textbf{A1}: only the connector is trained for image–language alignment; \textbf{A2}: A1 \emph{w/.} Cambrian-7M image instruction-tuning data; \textbf{A3}: A2 further finetuned on 429K video instruction-tuning samples; \textbf{A4}: A2 further finetuned on 3M video instruction-tuning samples. From A1 to A4, the models show a monotonic improvement in video understanding ability. I-IT and V-IT denote instruction finetuning on image and video data, respectively. Finally, we show that stronger base models yield better SFT performance on spatial sensing tasks.}
    \label{tab:training_pipeline}
    \resizebox{0.8\textwidth}{!}{%
    \begin{tabular}{c cccc} %
    \toprule
    \textbf{Model} & \textbf{VSI-Bench} & \textbf{VideoMME} & \textbf{EgoSchema} & \textbf{Perception Test} \\
    \hline
    \rowcolor{gray!10!blue!5} \multicolumn{5}{l}{Different Base Models} \\
        \multicolumn{1}{l}{A1 \scriptsize{(\emph{w/o.} I-IT, \ie~QwenLM)}} & 21.4 & 44.2 & 42.9 & 44.5 \\
        \multicolumn{1}{l}{A2 \scriptsize{(A1 + I-IT, \ie~Cambrian-1)}} & 25.8 & 53.7 & 48.1 & 55.4 \\
        \multicolumn{1}{l}{A3 \scriptsize{(A2 + V-IT, 429K data)}} & 28.9 & 61.2 & 50.3 & 66.3 \\
        \multicolumn{1}{l}{A4 \scriptsize{(A2 + V-IT, 3M data)}} & \textbf{35.7} & \textbf{62.6} & \textbf{77.0} & \textbf{70.9} \\
    \hline
    \rowcolor{gray!10!blue!5} \multicolumn{5}{l}{SFT \emph{w/.} \vsidata} \\
        from A1 & 57.2 & 40.3 & 38.7 & 52.3 \\
        from A2 & 66.8 & 46.7 & 47.2 & 52.3 \\
        from A3 & 68.8 & 52.3 & 48.4 & 55.8 \\
        from A4 & \textbf{69.2} & \textbf{54.1} & \textbf{55.2} & \textbf{59.2} \\
    \hline
    \rowcolor{gray!10!blue!5} \multicolumn{5}{l}{SFT \emph{w/.} \vsidata \& general V-IT data mixture} \\
        from A1 & 61.3 & 60.5 & 52.8 & 65.0\\
        from A2 & 63.2 &  \textbf{62.6} & 52.9 & 65.6 \\
        from A3 & 64.0 & 61.0 & 54.9 & 66.8   \\
        from A4 & \textbf{65.1} & 61.9 & \textbf{77.3} & \textbf{71.2}  \\
    \bottomrule
    \end{tabular}}
\end{table}

\subsection{\cambrianS: Spatially-Grounded MLLMs}\label{sec:limits:cambrian-s}

\begin{figure}[t]
    \centering
    \includegraphics[width=\linewidth]{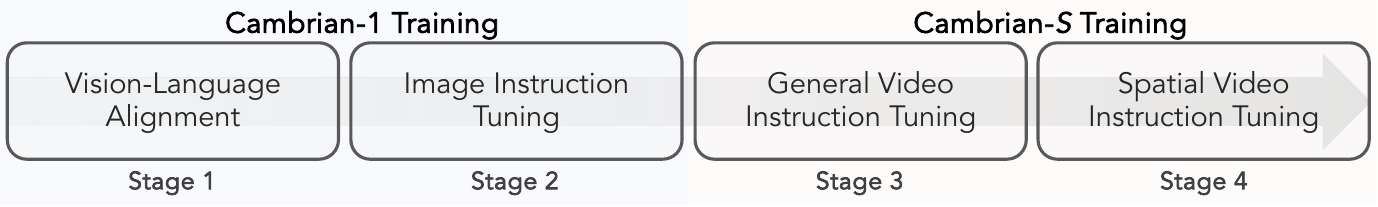}
    \caption{
        \small
        \textbf{Overall \cambrianS training pipeline.} Stages 1 and 2 enhance image understanding, stage 3 improves general video understanding, and stage 4 strengthens spatial sensing capability.
    }\label{fig:training_stage_simple_illustration}
\end{figure}

Building on all the previous insights, we develop \textbf{\cambrianS}, a family of spatially-grounded models with varying LLM scales: 0.5B, 1.5B, 3B, and 7B parameters. 
These models are built through a four-stage training pipeline specifically designed to first establish general semantic perception and then develop specialized spatial sensing skills, as illustrated in \cref{fig:training_stage_simple_illustration}. 

The first two stages adhere to the Cambrian-1 framework to develop strong image understanding capabilities. In stage 3, we extend the models to video by conducting general video instruction tuning on \osmix, a curated dataset composed of 3 million samples (see detailed composition in \cref{fig:osmix_sources}). This stage establishes a solid foundation for general video understanding prior to introducing specialized skills. In the final and crucial stage 4, the models are trained for spatial sensing. Here, we finetune the models on a blended corpus combining our specialized \vsidata with a proportional subset of the general video data used in stage 3, following the setup described in \cref{tab:training_pipeline}. Complete training details are provided in \cref{appx:cambrians_training_recipe}.

\begin{table}[t]
    \centering
    \caption{
        \small
        \textbf{Comparison of \cambrianS with other leading MLLMs.} \cambrianS outperforms both proprietary and open-source models across a range of image and video visual–spatial benchmarks and model sizes. For video evaluation, we uniformly sample 128 frames as input. Detailed evaluation settings are provided in \cref{appx:cambrians_additional_results}.
    }\label{tab:final_results}
    \resizebox{1\textwidth}{!}{
    \begin{tabular}{l|l|cccccccccc|ccc}
      \toprule
      & & \multicolumn{10}{c|}{Video} & \multicolumn{3}{c}{Image} \\[-0.5em]
          Model & Base LM &
          \rotatebox{90}{VSI-Bench} &
          \rotatebox{90}{VSI-Bench\textsuperscript{\textbf{Debiased}}} &
          \rotatebox{90}{Tomato} &
          \rotatebox{90}{HourVideo} &
          \rotatebox{90}{{Video\textsuperscript{MME}}~\textcolor{White}{\textbf{\textdagger}}} &
          \rotatebox{90}{{EgoSchema}~\textcolor{White}{\textbf{\textdagger}}} &
          \rotatebox{90}{{Video\textsuperscript{MMMU}}~\textcolor{White}{\textbf{\textdagger}}} &
          \rotatebox{90}{{LongVBench}~\textcolor{White}{\textbf{\textdagger}}} &
          \rotatebox{90}{{MVBench}~\textcolor{White}{\textbf{\textdagger}}} &
          \rotatebox{90}{{Percept.\ Test}~\textcolor{White}{\textbf{\textdagger}}} &
          \rotatebox{90}{MMVP} &
          \rotatebox{90}{3DSR} &
          \rotatebox{90}{CV-Bench} \\
 
      \hline\\[-1.15em]
      \hline

      \rowcolor{gray!10!blue!5} \textit{Proprietary Models} & & & & & & & & & & & & & & \\
      Claude-3.5-sonnet & UNK. & -    & -    & 27.8 & -    & 62.9 & -    & 65.8 & -    & - & - & - & \textbf{48.2} & - \\
      GPT-4o              & UNK. & 34.0 & -    & 37.7 & 37.2 & 71.9 & -    & 61.2 & 66.7 & - & - & \textbf{66.0} & 44.2 & - \\
      Gemini-1.5-Pro    & UNK. & 45.4 & 40.1 & 36.1 & 37.3 & 75.0 & 72.2 & 53.9 & 64.0 & - & - & - & -    & - \\
      Gemini-2.5 Pro    & UNK. & 51.5 & 49.1 & -    & -    & -    & -    & 83.6 & 67.4 & - & - & 51.3 & -    & - \\
     
      \hline
      \rowcolor{gray!10!blue!5}\textit{Open-Source Models} & & & & & & & & & & & & & & \\
      LLaVA-Video-7B      & Qwen2-7B       & 35.6 & 30.7 & 22.5 & 28.6 & 63.3         & 57.3 & 36.1         & 58.2         & 58.6         & 67.9         & -    & -    & 75.7 \\
      LLaVA-One-Vision-7B & Qwen2-7B       & 32.4 & 28.5 & 25.5 & 28.3 & 58.2         & 60.1 & 33.9         & 56.4         & 56.7         & 57.1         & 54.7 & -    & 74.3 \\
      Qwen-VL-2.5-7B      & Qwen2.5-7B     & 33.5 & 29.6 & -    & -    & 65.1         & 65.0 & 47.4         & 56.0         & 69.6         & -            & 56.7 & 48.4 & -    \\
      InternVL2.5-8B      & InternLM2.5-7B & 34.6 & 24.9 & -    & -    & 64.2         & 50.6 & -            & 60.0         & 72.0         & -            & 55.3 & 50.9 & -    \\
      InternVL3.5-8B      & Qwen3-8B       & 56.3 & 49.7 & -    & -    & \textbf{66.0} & 61.2 & \textbf{49.0} & \textbf{62.1} & \textbf{72.1} & -            & 56.0 & -    & -    \\
 
      \rowcolor{green!5} \textbf{\cambrianS-7B} & Qwen2.5-7B & \textbf{67.5} & \textbf{59.9} & \textbf{27.0} & \textbf{36.5} & 63.4 & \textbf{76.8} & 38.6 & 59.4 & 64.5 & \textbf{69.9} & \textbf{60.0} & \textbf{54.8} & \textbf{76.9}  \rule{0pt}{10pt} \\
      \hline
      VILA1.5-3B & Sheared-LLaMA-2.7B & - & - & - & - & 42.2 & - & - & 42.9 & - & 49.1 & - & - & - \\
      Qwen2.5-VL-3B & Qwen2.5-3B & 26.8 & 22.7 & - & - & \textbf{61.5} & - & - & \textbf{54.2} & - & \textbf{66.9} & 39.3 & - & - \\
      \rowcolor{green!5} \textbf{\cambrianS-3B} & Qwen2.5-3B & \textbf{57.3} & \textbf{49.7} & \textbf{25.4} & \textbf{36.8} & 60.2 & \textbf{73.5} & \textbf{25.2} & 52.3 & \textbf{60.2} & 65.9 & \textbf{50.0} & \textbf{50.9} & \textbf{75.2} \\
      \hline
      SmolVLM2-2.2B & SmolLM2-1.7B & 27.0 & 22.3 & - & - & - & 34.1 & - & - & 48.7 & 51.1 & - & - & - \\ 
      InternVL2.5-2B & InternLM2.5-1.8B & 25.8 & 20.7 & - & - & 51.9 & 47.4 & - & 52.0 & \textbf{68.8} & - & \textbf{45.3} & - & - \\
      InternVL3.5-2B & Qwen3-1.7B & 51.5 & 46.1 & - & - & \textbf{58.4} & 50.8 & - & \textbf{57.4} & 65.9 & - & 44.0 & - & - \\
      \rowcolor{green!5} \textbf{\cambrianS-1.5B} & Qwen2.5-1.5B & \textbf{54.8} & \textbf{47.5} & \textbf{22.5} & \textbf{31.4} & 55.6 & \textbf{68.8} & 24.9 & 50.0 & 58.1 & \textbf{63.2} & 42.7 & \textbf{51.9} & \textbf{69.6} \\
      \hline
      SmolVLM2-0.5B & SmolLM2-360M & 26.1 & 23.1 & - & - & - & 20.3 & - & - & 43.7 & 44.8 & - & - & - \\
      LLaVA-One-Vision-0.5B & Qwen2-0.5B & 28.5 & 20.6 & - & - & 44.0 & 26.8 & - & 45.8 & 45.5 & 49.2 & 28.7 & - & 55.5 \\
      InternVL2.5-1B & Qwen2.5-0.5B & 22.5 & 17.5 & - & - & 50.3 & 39.8 & - & 47.9 & \textbf{64.3} & - & \textbf{33.3} & - & - \\
      InternVL3.5-1B & Qwen3-0.6B & 49.9 & 41.8 & - & - & \textbf{51.0} & 41.5 & \textbf{33.0} & \textbf{53.0} & 61.0 & - & 32.0 & - & - \\
      \rowcolor{green!5} \textbf{\cambrianS-0.5B} & Qwen2.5-0.5B & \textbf{50.6} & \textbf{42.2} & \textbf{23.4} & \textbf{27.9} & 44.0 & \textbf{62.4} & 15.7 & 44.0 & 51.8 & \textbf{56.0} & 26.0 & \textbf{48.5} & \textbf{59.8} \\
      \bottomrule
    \end{tabular}}
    \vspace{-1.0em}
\end{table}

\subsection{Empirical Results: Improved Spatial Cognition}\label{sec:limits:resuls}

We next evaluate the \cambrianS multimodal models to assess both the strengths and limitations of our data-driven approach.

\begin{table}[h]
    \centering
    \caption{
    \textbf{VSI-Bench sub-task breakdown.}
    Best results are \textbf{bolded}. Notably, even without any route planning data in training, \cambrianS-7B outperforms Gemini-1.5-Pro on this task.
    }\label{tab:vsibench_subtask_results}
    \vspace{-1em}
    \resizebox{0.75\textwidth}{!}{
        \begin{tabular}{l|c|cccccccc} & & \rotatebox{75}{Obj. Count} & \rotatebox{75}{Abs. Dist.} & \rotatebox{75}{Obj. Size} & \rotatebox{75}{Room Size} & \rotatebox{75}{Rel. Dist.} & \rotatebox{75}{Rel. Dir.} & \rotatebox{75}{Route Plan} & \rotatebox{75}{Appr. Order} \\
        Methods & Avg. & \multicolumn{4}{c}{\cellcolor{orange!10}Numerical Answer} & \multicolumn{4}{c}{\cellcolor{yellow!10}Multiple-Choice Answer} \\
        \hline \rowcolor{navyblue!5} \multicolumn{1}{l|}{\textcolor{black}{\textit{Statistics}}} & & & & & & & & & \\
        Chance Level (Random) & - & - & - & - & - & 25.0 & 36.1 & 28.3 & 25.0 \\
        Chance Level (Frequency) & 34.0 & 62.1 & 32.0 & 29.9 & 33.1 & 25.1 & 47.9 & 28.4 & 25.2 \\
        \hline \rowcolor{navyblue!5} \multicolumn{1}{l|}{\textcolor{black}{\textit{Proprietary Models (API)}}} & & & & & & & & & \\ GPT-4o & 34.0 & 46.2 & 5.3 & 43.8 & 38.2 & 37.0 & 41.3 & 31.5 & 28.5 \\
        Gemini-1.5 Flash & 42.1 & 49.8 & 30.8 & 53.5 & 54.4 & 37.7 & 41.0 & 31.5 & 37.8 \\
        Gemini-1.5 Pro & 45.4 & 56.2 & 30.9 & 64.1 & 43.6 & 51.3 & 46.3 & 36.0 & 34.6 \\
        Gemini-2.5 Pro & 51.5 & 43.8 & 34.9 & 64.3 & 42.8 & 61.1 & 47.8 & \textbf{45.9} & 71.3 \\
        \hline \rowcolor{navyblue!5} \multicolumn{1}{l|}{\textcolor{black}{\textit{Open-source Models}}} & & & & & & & & & \\
        \cambrianS-7B & \textbf{67.5} & \textbf{73.2} & \textbf{50.5} & \textbf{74.9} & \textbf{72.2} & \textbf{71.1} & \textbf{76.2} & 41.8 & \textbf{80.1}\\
        \cambrianS-3B & 57.3 & 70.7 & 40.6 & 68.0 & 46.3 & 64.8 & 61.9 & 27.3 & 78.8 \\
        \cambrianS-1.5B & 54.8& 68.4 & 40.0 & 61.5 & 50.1 & 62.4 & 48.9 & 29.9 & 77.5 \\
        \cambrianS-0.5B & 50.6 & 67.9 & 35.4 & 52.2 & 52.5 & 52.3 & 46.5 & 25.8 & 72.2\\
    \end{tabular}}
    \vspace{-1em}
\end{table}

\paragraph{Improved spatial cognition.}
As shown in \cref{tab:final_results}, our models achieve state-of-the-art performance in visual-spatial understanding in video. \cambrianS-7B achieves 67.5\% on VSI-Bench, significantly outperforming all open-source models and surpassing the proprietary Gemini-2.5-Pro by over 16 absolute points. 
Since our work in this section can be viewed as a data scaling effort, a natural question is: \emph{are the performance improvements simply due to broader data coverage (including more diverse visual configurations and question–answer pairs), or has the model actually developed stronger spatial cognition?} First, we emphasize that there is no data overlap between \vsidata and the benchmark datasets. Although some datasets originate from the same sources (\eg from ScanNet), we only use the training split, while the benchmarks use validation and test splits. Moreover, we observe clear signs of generalization in spatial reasoning. For example, in the challenging ``Route Planning'' subtask, whose question types are absent from \vsidata{} because of the high annotation cost, \cambrianS-7B still performs strongly, showing pronounced scaling behavior with increasing model size too (see \cref{tab:vsibench_subtask_results}). 

Furthermore, our training approach proves highly effective even with smaller model sizes: our smallest \emph{0.5B model} achieves performance comparable to Gemini-1.5 Pro on VSI-Bench. Importantly, this emphasis on spatial reasoning does not come at the expense of general capabilities: \cambrianS{} continues to deliver competitive results on standard video benchmarks such as Perception Test~\cite{patraucean2023perception} and EgoSchema~\cite{mangalam2023egoschema} (see \cref{tab:video_benchmarks_results} for complete results).

\finding{}{\faBookmark~~\cambrianS{} achieves state-of-the-art spatial sensing performance with robust generalization to unseen spatial question types, while staying competitive in general video understanding.}

\paragraph{Robust spatial reasoning on VSI-Bench-Debiased.}
A recent study~\cite{brown2025shortcuts} reveals that models can rely on strong language priors for spatial reasoning tasks. For instance, when asked to estimate a table's length, a model might leverage natural world knowledge about typical table sizes (\eg, 120--180 cm) rather than analyzing the visual evidence.
To investigate whether \cambrianS{} learns to reason visually, we evaluate it on VSI-Bench-Debiased~\cite{brown2025shortcuts}, a benchmark specifically designed to eliminate language shortcuts through debiasing. As shown in \cref{tab:final_results}, although performance decreases by about 8\% compared to standard VSI-Bench, our models still outperform proprietary counterparts, demonstrating robust visual-spatial reasoning capabilities and confirming that our training extends beyond language-based learning.

\paragraph{Results on VSI-Super: limitations in continual spatial sensing.}

Despite its strong performance on spatial reasoning tasks in short, pre-segmented videos from VSI-Bench, \cambrianS isn't well-equipped for continual spatial sensing. This limitation is evident in two ways. First, its performance deteriorates significantly on long videos.
As shown in \cref{tab:cams7b_on_vsi_super}, when evaluated on \vsisuper{} with 1 FPS sampling in a streaming-style setup, scores drop steadily from 38.3\% to 6.0\% as video length increases from 10 to 60 minutes, and the model fails completely on videos longer than 60 minutes.
Second, the model has difficulty generalizing to new test scenarios. Although trained on multi-room house tour videos, it fails to handle unseen examples with just a few additional rooms. This issue isn't simply about context length: performance drops even on short 10-minute videos that fit comfortably within model's context window. These results highlight that a purely data-driven approach within the current MLLM framework, no matter how much data or engineering effort is invested, faces fundamental limits. Addressing these limitations calls for a paradigm shift toward AI systems that can actively model and anticipate the world while organizing their experiences more efficiently, which we explore next.

\finding{}{\faBookmark~~~Scaling data and models is essential, but alone it cannot unlock true spatial supersensing.}

\begin{table}[t]
    \centering
    \caption{
        \textbf{
            \cambrianS-7B results on \vsisuper{}.}
            Despite strong performance on VSI-Bench, accuracy on \vso{} drops sharply from 38.3\% (10 min) to 0.0\% (>60 min), and \vsc{} completely fails. Note that \vsisuper{} focuses on continual, streaming evaluation, where uniform sampling 128 frames across the entire video does not align with the online setting; results shown in \textcolor{gray!50}{gray} are provided for reference only.
    }\label{tab:cams7b_on_vsi_super}
    \resizebox{1.0\textwidth}{!}{
        \begin{tabular}{c|ccccc|cccc}
            & \multicolumn{5}{c|}{\vso} & \multicolumn{4}{c}{VSC} \\
            Eval Setup & 10 min & 30 min & 60 min & 120 min & 240 min & 10 mins & 30 min & 60 min & 120 min \\
            \hline
            \textcolor{gray!50}{Uni. Sampling, 128F} & \textcolor{gray!50}{26.7} & \textcolor{gray!50}{21.7} & \textcolor{gray!50}{23.3} & \textcolor{gray!50}{30.0} & \textcolor{gray!50}{28.2} & \textcolor{gray!50}{16.0} & \textcolor{gray!50}{0.0} & \textcolor{gray!50}{0.0} & \textcolor{gray!50}{0.0} \rule{0pt}{10pt} \\
            FPS Sampling, 1FPS & 38.3 & 35.0 & 6.0 & 0.0 & 0.0 & 0.6 & 0.0 & 0.0 & 0.0 \rule{0pt}{10pt} \\
        \end{tabular}
    }
\end{table}


\section{Predictive Sensing as a New Paradigm}\label{sec:predictive-sensing}

Performance of both Gemini-2.5-Flash (\cref{tab:gemini_2.5_flash_results}) and \cambrianS{} (\cref{tab:cams7b_on_vsi_super}) drops sharply on \vsisuper{}, revealing a fundamental paradigm gap: scaling data and context alone is insufficient for supersensing.
We propose \emph{predictive sensing} as a path forward, where models learn to anticipate their sensory input and construct internal world models to handle unbounded visual streams.
This design is inspired by theories of human cognition. Unlike current video multimodal models that tokenize and process entire data streams, human perception (and memory) is highly selective, retaining only a fraction of sensory input~\cite{von1867handbuch,millidge2022predictive,hohwy2013predictive,rao1999predictive}.
The brain continuously updates internal models to predict incoming stimuli, compressing or discarding predictable inputs that contribute no novel information~\cite{clark2013whatever,friston2010free}.
In contrast, unexpected sensory information that violates predictions generates ``surprise'' and drives increased attention and memory encoding~\cite{schultz2000neuronal,gershman2017computational,kennedy2024prediction}.
We prototype this concept via a self-supervised next-latent-frame prediction approach (\cref{sec:predictive-sensing:lfp}).
The resulting prediction error serves as a control signal for two key capabilities: memory management to selectively retain important information (\cref{sec:case_study_1_vso}), and event segmentation to partition unbounded streams into meaningful chunks (\cref{sec:case_study_2_vsc}).
We demonstrate through two case studies on \vsisuper{} that this approach substantially outperforms strong long-context and streaming video model baselines.

\subsection{Predictive Sensing via Latent Frame Prediction}
\label{sec:predictive-sensing:lfp}
\begin{figure}[h]
    \centering
    \includegraphics[width=1.0\textwidth]{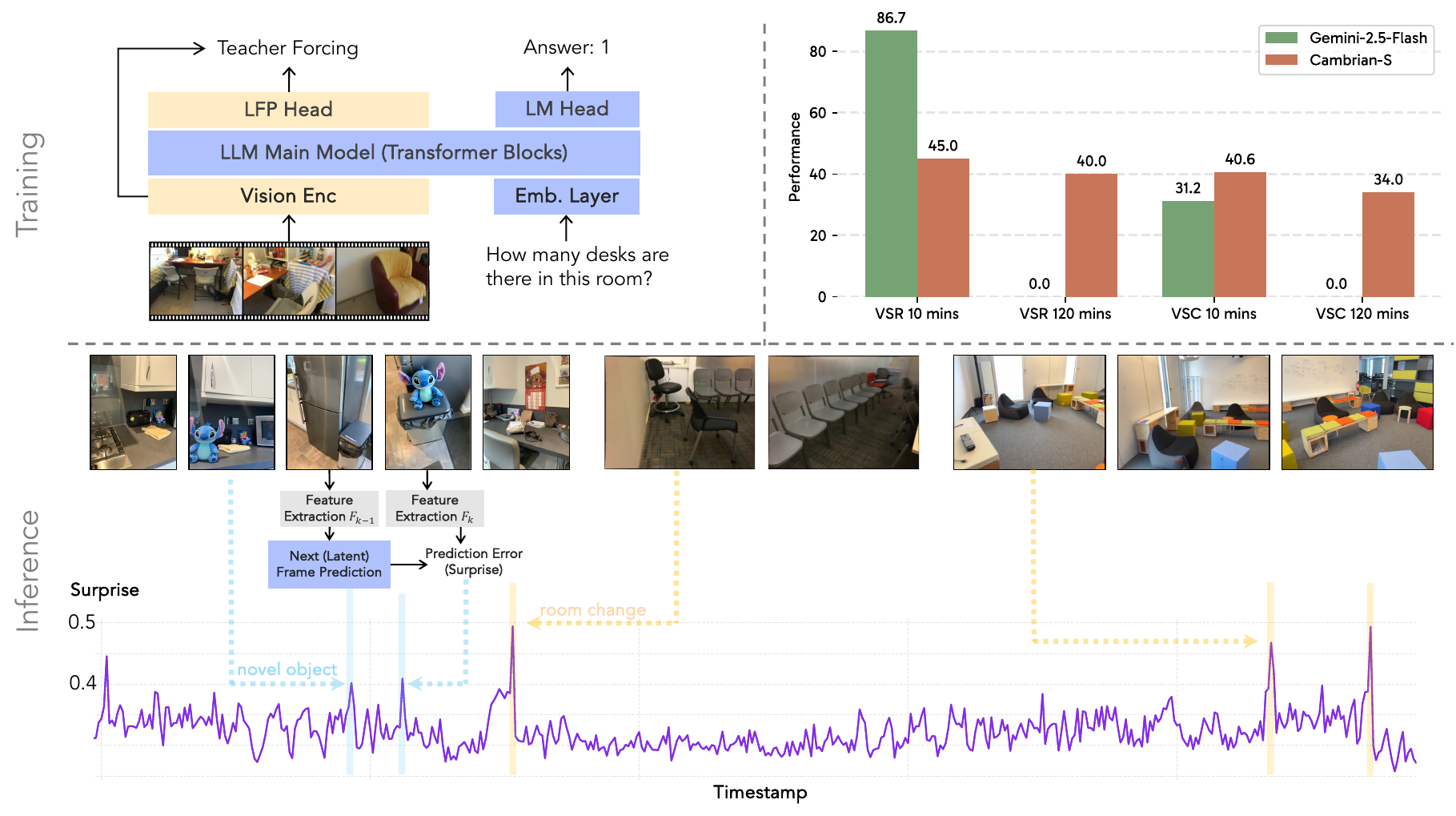}
    \caption{
        \textbf{Training and inference pipeline for the latent frame prediction (LFP) approach.}
        Our model employs a Latent Frame Prediction (LFP) head to predict the next frame in latent space.
        During training, the LFP head predicts the latent representation of the subsequent video frame.
        During inference, the model measures surprise by computing the cosine distance between the LFP head's prediction and the actual latent features of the subsequent frame.
        The surprise signal exhibits distinct spikes for events such as the sudden appearance of unusual objects and abrupt scene changes.
        Our predictive-sensing prototype allows \cambrianS{} to generalize to longer videos on \vsisuper{}, outperforming frontier models (\eg, Gemini-2.5-Flash) that rely solely on context length expansion.
    }\label{fig:lfp_framework}
\end{figure}

We implement our predictive sensing paradigm through a lightweight, self-supervised module called the Latent Frame Prediction (LFP) head, which is trained jointly with the primary instruction-tuning objective. This is achieved by modifying the stage 4 training recipe as follows:

\begin{itemize}[nosep, leftmargin=*]
    \item \textbf{Latent frame prediction head.}
    We introduce an LFP Head, a two-layer MLP that operates in parallel with the language head, to predict the latent representation of the subsequent video frame. This architecture is illustrated in the top left of \cref{fig:lfp_framework}.
    \item \textbf{Learning objectives.}
    To optimize the LFP head, we introduce two auxiliary losses, mean squared error (MSE) and cosine distance, which measure the discrepancy between the predicted latent feature and the ground truth feature of the next frame. A weighting coefficient balances the LFP loss against the primary instruction-tuning next token prediction objective.
    \item \textbf{Data for LFP training.}
    We augment stage 4 data with a 290K video subset from \vsidata used exclusively for the LFP objective. Unlike instruction tuning, these videos are sampled at a constant rate of 1 FPS to ensure uniform temporal spacing for latent frame prediction.
\end{itemize}

During this modified stage 4 finetuning, we train the connectors, language model, and both the language and LFP heads jointly in an end-to-end manner, while keeping the SigLIP vision encoder frozen. All other training settings remain consistent with the original stage 4 configuration.
For brevity, we still denote the model jointly optimized with the LFP objective as \cambrianS{} in subsequent experiments.

\paragraph{Inference: Estimating surprise via prediction error.}
During inference, we leverage the trained LFP head to evaluate the ``surprise'' for every incoming visual sensory input. In psychology, this framework is often described as the Violation-of-Expectation (VoE) paradigm~\cite{burgoon1988nonverbal}. Specifically, during inference, video frames are fed into \cambrianS at a constant sampling rate. Unless otherwise noted, the videos in the following experiments are sampled at 1 FPS before being input into the model.
As the model receives incoming video frames, it continuously predicts the latent features of the next frame. We then measure the \textit{cosine distance} between the model's prediction and the actual ground truth feature of that incoming frame. This distance serves as a quantitative measure of surprise: a larger value indicating a greater deviation from the model's learned expectations. This surprise score acts as a powerful, self-supervised guidance signal for the downstream tasks explored next.

\begin{figure}[h]
    \centering
    \includegraphics[width=0.5\linewidth]{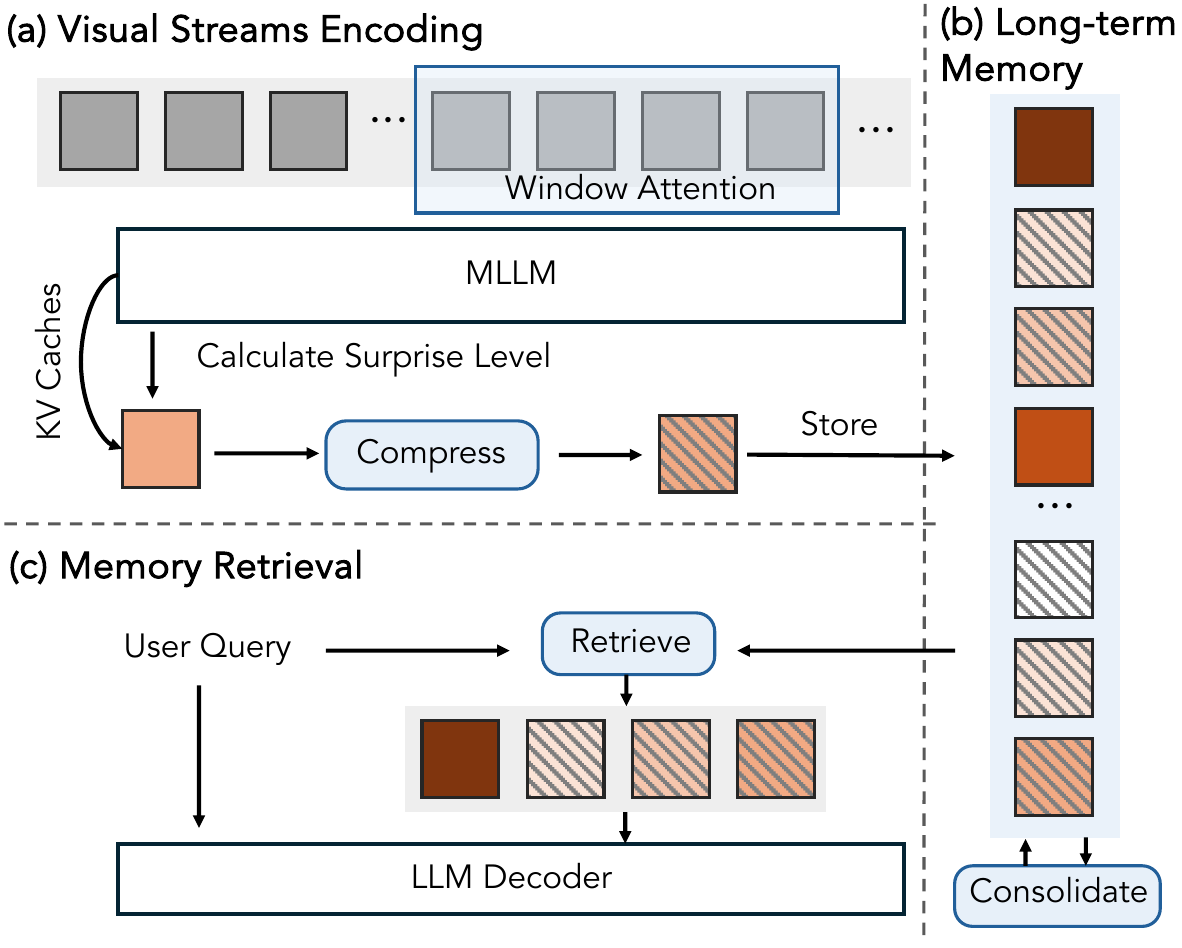}
    \caption{
        \textbf{Surprise-driven memory management framework design.}
        The proposed memory system (a) encodes incoming visual streams, compressing frames with low surprise; (b) performs consolidation when memory is full by dropping or merging the least surprising frames; and (c) retrieves relevant frames during query answering. Color shading (dark$\rightarrow$light) reflects the degree of surprise, with hatched boxes denoting compressed frames and solid boxes representing uncompressed ones.
    }\label{fig:sor_memory_design}
\end{figure}

\subsection{Case Study I: Surprise-driven Memory Management System for \vsisuper{} Recall.}\label{sec:case_study_1_vso}
Most current MLLMs treat all video frames equally, storing every frame without selective compression or forgetting, which limits efficiency and scalability. In this case study, we explore augmenting MLLMs with a \textbf{surprise–driven memory management} framework to support continual spatial-sensing question answering over long-duration videos.
We show that through the surprise-guided compression, \cambrianS{} maintains consistent accuracy and stable GPU memory footprints, independent of video length.

\paragraph{Surprise-driven memory management system.}
Our memory management system dynamically compresses and consolidates visual streams based on the estimate of ``surprise''.
As shown in~\cref{fig:sor_memory_design}-a, we encode incoming frames using sliding window attention with fixed window size. The latent frame prediction module then measures a ``surprise level'' and assigns it to each frame's KV caches. Frames with a surprise level below a predefined threshold undergo $2\times$ compression before being pushed into long-term memory.
To maintain a stable GPU memory footprint, this long-term memory is constrained to a fixed size by a consolidation function that, once again, operates based on \emph{surprise}: dropping or merging frames according to their surprise scores (see \cref{fig:sor_memory_design}-b).
Finally, upon receiving a user query, the system retrieves the top-$K$ most relevant frames from the long-term memory by calculating the cosine similarity between the query and the stored frame features (see \cref{fig:sor_memory_design}-c).
See \cref{appx:pred_sensing:vso_mem_design} for more design details.
While prior works have explored memory system designs for long videos~\cite{song2024moviechat,zhang2024flash}, our focus is on exploring prediction errors (\ie, surprise) as guiding signals.

\begin{figure}[h]
    \centering
    \begin{subfigure}[t]{0.329\textwidth}
        \centering
        \includegraphics[width=\linewidth]{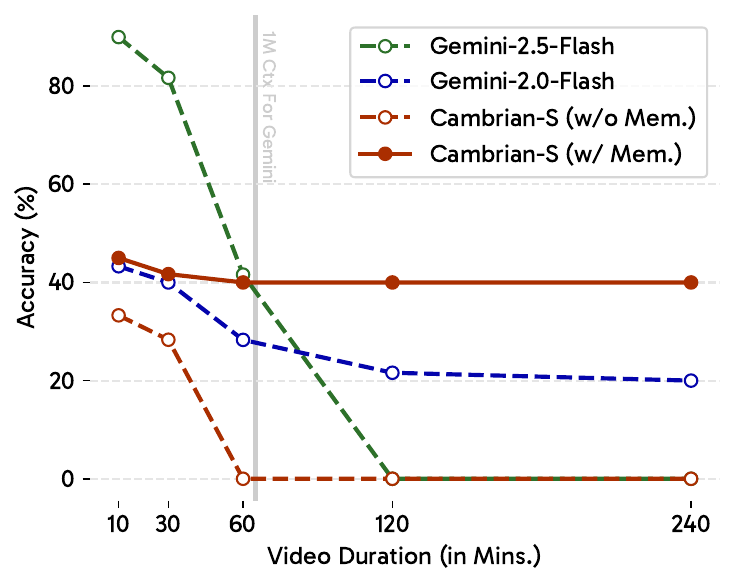}
        \vspace{-1.5em}
        \caption{\centering\small \vso{} results}\label{fig:accuracy_on_vlo}
    \end{subfigure}\hfill
    \begin{subfigure}[t]{0.329\textwidth}
        \centering
        \includegraphics[width=\linewidth]{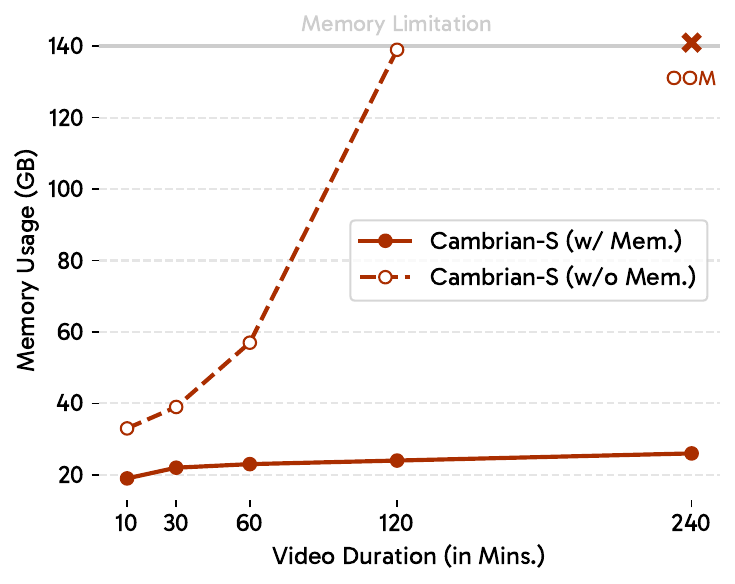}
        \vspace{-1.5em}
        \caption{\centering\small GPU memory usage}\label{fig:gpu_mem_on_vlo}
    \end{subfigure}\hfill
    \begin{subfigure}[t]{0.329\textwidth}
        \centering
        \includegraphics[width=\linewidth]{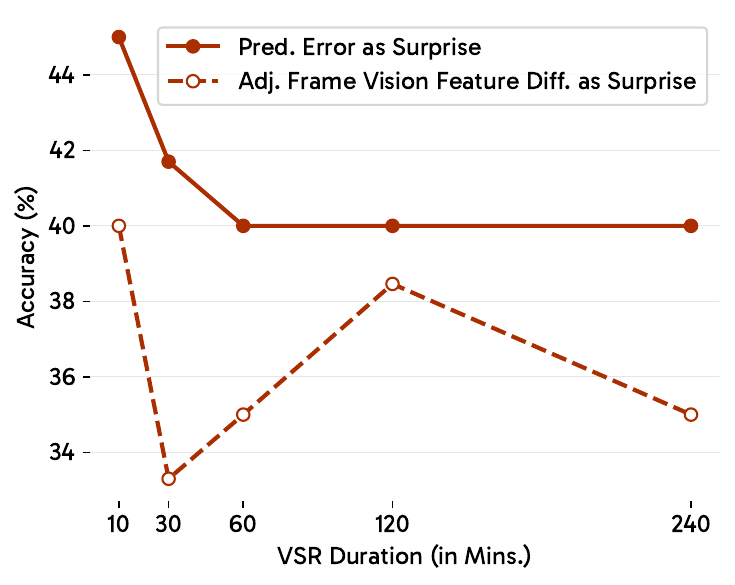}
        \vspace{-1.5em}
        \caption{\centering\small Surprise comparison}\label{fig:ablation_on_vso}
    \end{subfigure}
    \vspace{-0.5em}
    \caption{
        \textbf{Performance analysis of surprise-driven memory on \vso{}.}
        (a) Surprise-driven memory allows \cambrianS{} to maintain strong performance as video length increases.
        (b) Surprise-driven memory maintains a stable GPU memory footprint as video length increases.
        (c) Ablation: Using LFP prediction error as the surprise signal is more robust and consistently outperforms using adjacent-frame similarity.
    }\label{fig:vso_analysis}
    \vspace{-1em}
\end{figure}

\paragraph{Results.}
We compare \cambrianS{} with and without the surprise-based memory system against two advanced proprietary models, Gemini-1.5-Flash~\cite{team2024gemini} and Gemini-2.5-Flash~\cite{comanici2025gemini}, on the \vso{} benchmark.
As shown in \cref{fig:accuracy_on_vlo}, \cambrianS{} (\emph{w/} Mem.) outperforms both Gemini-1.5-Flash and \cambrianS{} (\emph{w/o.} Mem.) at all video lengths, demonstrating consistent spatial sensing performance across video durations. Although Gemini-2.5-Flash yields strong results for videos within an hour, it fails to process longer inputs.
In addition to maintaining high accuracy, \cambrianS{} (\emph{w/} Mem.) also maintains stable GPU memory usage across different video lengths (\cref{fig:gpu_mem_on_vlo}).
This demonstrates that surprise-based memory effectively compresses redundant data without losing critical information. We include two long-video baselines, MovieChat~\cite{song2024moviechat} and Flash-VStream\cite{zhang2024flash}, for comparison in \cref{tab:comparisons_with_existing_long_video_methods}.

\paragraph{Ablation on surprise measurement.}
Central to our surprise-based memory system is the mechanism for measuring surprise, which dictates how frames are compressed or consolidated in a passive sensing manner---without assuming any prior knowledge of future queries. Here, we compare our design, prediction error as surprise, to another straightforward baseline: adjacent-frame visual-feature similarity. Specifically, we use SigLIP2 as the vision encoder and directly compare the frame feature difference (cosine distance) between two adjacent frames. If the difference exceeds a threshold, we treat the later frame as a surprise frame.
We compare these two methods across all \vso{} variants.
For each \vso{} duration, we keep the experimental setup identical except for the surprise threshold, which we tune for both methods.
As shown in \cref{fig:ablation_on_vso}, using prediction error as the surprise measurement consistently outperforms adjacent-frame similarity across different video durations. 

\finding{}{\faBookmark~~Predictive sensing provides a more principled approach to modeling the spatiotemporal dynamics of video data than static similarity measures based on per-frame features.}
While our current system employs a simple predictive head as an initial prototype, 
future integration of a more capable world model could produce richer and more reliable surprise signals, ultimately enabling broader advances in spatial supersensing.

\subsection{Case Study II: Surprise-driven continual video segment for \vsisuper{} Count.}\label{sec:case_study_2_vsc}
While \vso{} focuses on evaluating the long-term observation and recall abilities of MLLMs, a more challenging test of supersensing would involve testing a model's capacity to interpret its sensory input, navigate across varied environments, and perform cumulative, multihop reasoning. For example, the model might need to complete a task in one environment, move to another, and ultimately integrate information from all experiences to reach a final decision.

\begin{figure}[!h]
    \centering
    \includegraphics[width=0.575\linewidth]{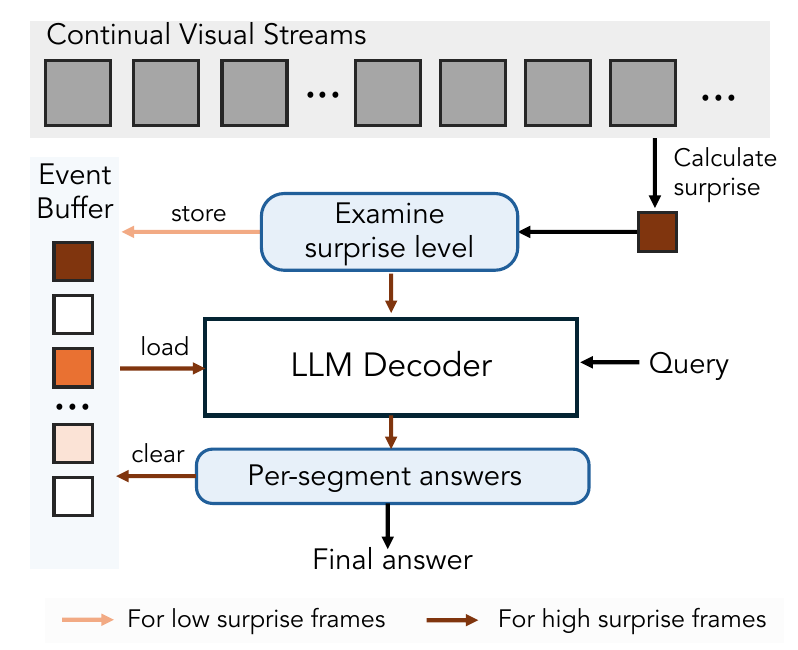}
    \caption{
        \textbf{Illustration of our surprise-driven event segmentation framework for \vsc.}
        The model continuously accumulates frame features in an event buffer. When a high-surprise frame is detected, the buffered features are summarized to produce a segment-level answer, and the buffer is cleared to start a new segment. This process repeats until the end of the video, after which all segment answers are aggregated to form the final output. Color shading (dark$\rightarrow$light) reflects the degree of surprise.
        }
    \label{fig:vsi_long_count_framework}
\end{figure}

\paragraph{Surprise-driven event segmentation.}
An \emph{event} can be understood as a spatiotemporally coherent segment of experience~\cite{kurby2008segmentation}. In the context of spatial supersensing, an event corresponds to a continuous experience of being situated within a specific space and sensing its environment. This definition emphasizes that real sensory experience is typically organized into locally coherent segments---episodes where perceptual, spatial, and temporal features remain relatively stable or consistent. Event segmentation, then, is the process of parsing a continuous stream of sensory input into discrete, meaningful units based on changes in this coherence. Such segmentation is essential for reasoning and behavior~\cite{dominey2021narrative}: it allows an agent (biological or artificial) to form structured representations of experience, detect boundaries where significant change occurs, and update predictions about the environment accordingly. Recent studies highlight that prediction error and changes in working memory/context are two possible mechanisms driving segmentation ~\cite{nolden2024prediction,shim2024generating}.

In the \vsisuper{} Count (VSC) benchmark, we examine a simple setting where surprise is used to segment continuous visual input, identifying scene changes as natural breakpoints that divide the video stream into \emph{spatially coherent} segments. This approach also parallels human problem-solving: when counting objects across a large area, people typically focus on one section at a time before combining the results. This behavior is also related to the ``doorway effect''~\cite{radvansky2011walking}, in which passing through a doorway or entering a new room creates a natural boundary in memory.
As illustrated in \cref{fig:vsi_long_count_framework}, the model continuously accumulates frame features in an event buffer. When a high-surprise frame is detected, the buffered features are summarized to produce a segment-level answer, and the buffer is cleared to start a new segment. This cycle repeats until the end of the video, after which all segment answers are aggregated to form the final output.

\begin{figure}[h]
    \centering
    \begin{subfigure}[t]{0.329\textwidth}
        \centering
        \includegraphics[width=\linewidth]{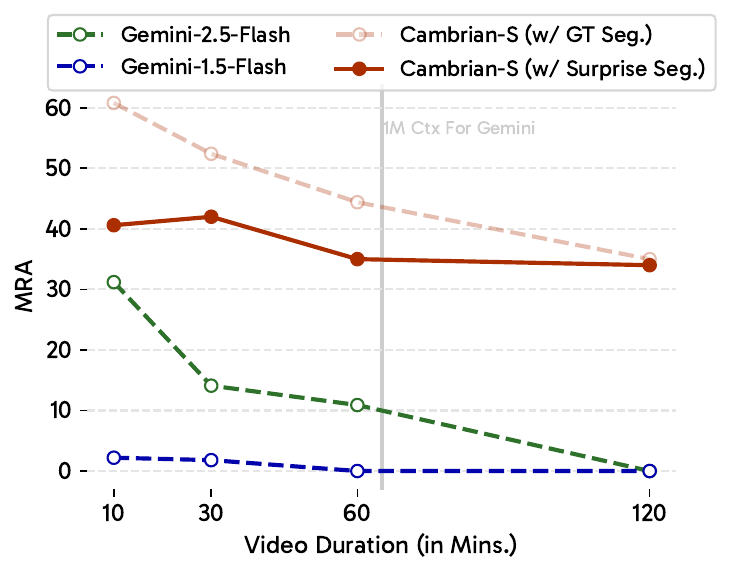}
        \vspace{-1.5em}
        \caption{\centering\small \vsc results}\label{fig:count-results}
    \end{subfigure}\hfill
    \begin{subfigure}[t]{0.329\textwidth}
        \centering
        \includegraphics[width=\linewidth]{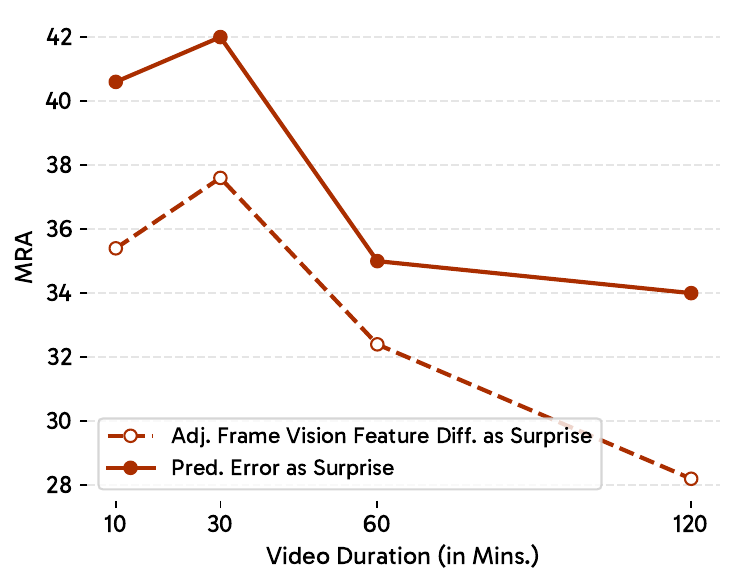}
        \vspace{-1.5em}
        \caption{\centering\small Different surprise measurement}\label{fig:count-comparison}
    \end{subfigure}\hfill
    \begin{subfigure}[t]{0.329\textwidth}
        \centering
        \includegraphics[width=\linewidth]{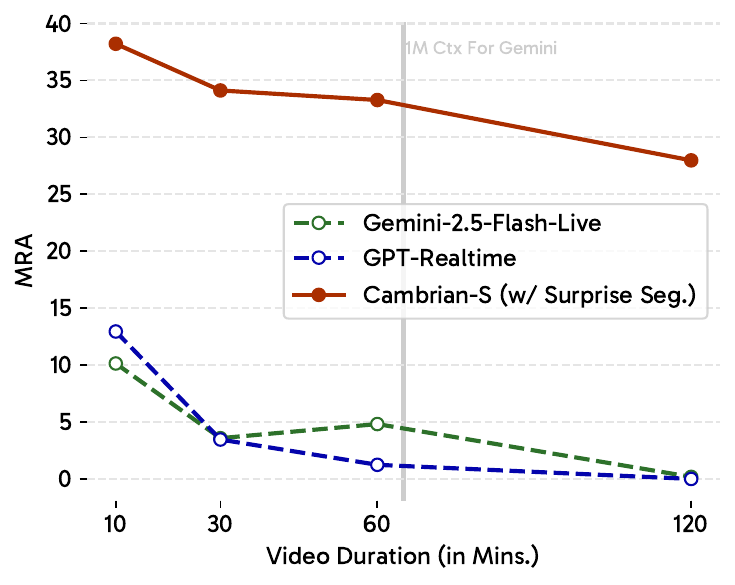}
        \vspace{-1.5em}
        \caption{\centering\small Streaming evaluation}\label{fig:vsc_streaming_results}
    \end{subfigure}
    \vspace{-0.5em}
    \caption{
        \textbf{Performance analysis on \vsc{}.}
        (a) \cambrianS{} with surprise-driven event segmentation achieves consistently higher and more stable performance across all video lengths compared to Gemini-2.5-Flash;
        (b) Ablation: prediction error as surprise outperforms adjacent-frame similarity;
        (c) Streaming evaluation: Although GPT-Realtime and Gemini-Live are marketed as ``live assistants'', they achieve less than 15\% MRA and their performance drops to near zero on long videos, while our method maintains substantially higher performance.
    }\label{fig:vsc_analysis}
\end{figure}

\begin{figure}[h]
    \centering
    \includegraphics[width=\linewidth]{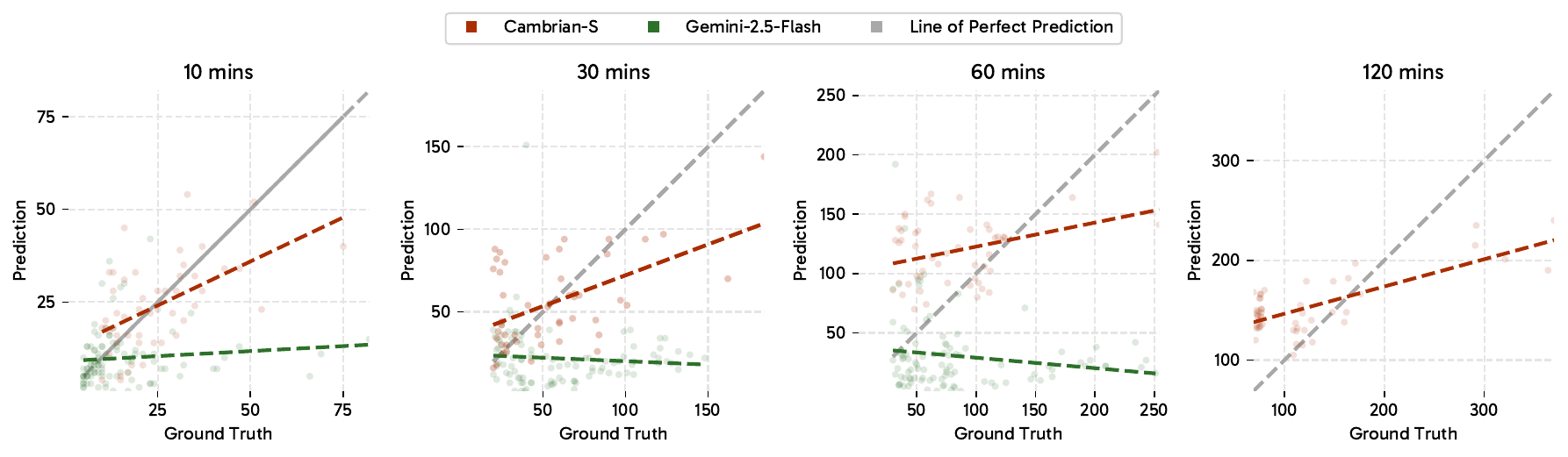}
    \caption{
        \textbf{\cambrianS{} scales to higher ground truth object counts whereas Gemini saturates.}
        Predicted counts are plotted against ground-truth counts for videos of different lengths (10, 30, 60, and 120 minutes).
        Using surprise-driven segmentation, \cambrianS{}'s predicted counts grow approximately linearly with the ground-truth, tracking the $y=x$ perfect-count line (gray dashed),
        whereas Gemini-2.5-Flash's predicted counts remain clustered near small values and fail to increase with ground-truth count, indicating early saturation and poor extrapolation to larger counts.
    }\label{fig:count_pred_distribution}
\end{figure}

\paragraph{Results.}
Gemini-1.5-Flash attains near-zero performance on \vsc{} (\cref{fig:count-results}), showing the task's difficulty. Although Gemini-2.5-Flash yields much better results on 10-minute videos, its performance declines rapidly on longer videos.
In contrast, the surprise-driven event segmentation approach used by \cambrianS{} (\emph{w/} Surprise Seg.) achieves higher and more stable performance across all video lengths. When the video is segmented using ground-truth scene transitions (\ie, \cambrianS{} \emph{w/} GT Seg.), performance improves further, representing an approximate upper bound.
A deeper analysis in \cref{fig:count_pred_distribution} reveals that Gemini-2.5-Flash's predictions are confined to a limited range and do not scale as more objects appear in the video.
In contrast, \cambrianS{} (\emph{w/} Surprise Seg.) produces counts that, while not yet fully accurate, exhibit a stronger correlation with the true object numbers, indicating better generalization.

\paragraph{Ablation on surprise measurement.}

We compare our surprise-driven approach with a baseline using adjacent-frame feature similarity (\cref{fig:count-comparison}).
For both methods, we report the best results after hyperparameter tuning. Consistent with our observations in VSR, using prediction error as a measure of surprise consistently outperforms appearance similarity across all video durations by a notable margin.

\paragraph{Evaluation in streaming setup.}
As the correct answer in \vsc{} evolves throughout the video, we create a streaming QA setup where the same question is asked at 10 different timestamps. The final performance is averaged across all queries.
We benchmark against commercial MLLMs marketed for live visual input. As shown in \cref{fig:vsc_streaming_results}, although Gemini-Live and GPT-Realtime are intended for streaming scenarios, they achieve under 15\% MRA on 10-minute videos and their performance declines to near zero on 120-minute streams. \cambrianS{}, however, shows stronger performance, reaching 38\% MRA on 10-minute streams and maintaining around 28\% at 120 minutes.

\paragraph{Summary.} Across both VSR recall and VSC counting tasks, predictive sensing through surprise-driven memory and event segmentation enables \cambrianS{} to overcome the fixed-context limitations described in \cref{sec:limits}. Although this remains an early prototype, it highlights the potential for building AI systems that not only see but also anticipate, select, and organize experience. Such systems move beyond frame-level Q\&A toward constructing implicit world models that support deeper spatial reasoning, scale across unbounded temporal horizons, and achieve supersensing that rivals and ultimately surpasses human visual intelligence.


\section{Related Work}\label{sec:related_work}

\paragraph{Video Multimodal Large Language Models} The strong linguistic understanding capabilities of pretrained LLMs~\cite{brown2020language,touvron2023llama,bai2023qwen,touvron2023llama2}, combined with the representational power of vision foundation models used as feature extractors~\cite{radford2021learning,zhai2023sigmoid,tschannen2025siglip,he2022masked,fan2025scaling}, have driven significant advances in extending these models beyond text to achieve semantic perception of visual content, primarily in the image domain~\cite{hurst2024gpto,liu2023visual,li2024llava,bai2023qwenvl,tong2024cambrian,team2023gemini,chen2024internvl,wang2024qwen2vl,li2023blip2}. This momentum has spurred growing research into video-based MLLMs~\cite{li2024llama,li2024llava,zhang2024video,song2024moviechat,bai2025qwen2,zhu2025internvl3,zhang2023video,li2023videochat,zohar2024apollo,marafioti2025smolvlm}, which are seen as a key step toward connecting multimodal intelligence with real-world applications such as embodied agents~\cite{kim2024openvla,yang2024virl}. As emphasized throughout this paper, developing a truly capable supersensing system requires rethinking several core aspects, including how progress is benchmarked, what constitutes the right data, which architectural designs are most effective, and what modeling objectives best align with the system's goals.

\paragraph{Streaming Video Understanding}
Video is a continuous and potentially infinite stream of visual signals. While humans process it effortlessly, its unbounded nature challenges video MLLMs because token lengths increase with duration, causing rising computational and storage costs. Recent work has explored several approaches to address this problem: \textit{Efficient architectural design}. The quadratic cost of self-attention makes it hard to handle long videos. Recent methods~\cite{li2024videomamba,ren2025vamba} use simpler, faster architectures~\cite{wang2020linformer,gu2023mamba,katharopoulos2020transformers} that reduce computation and work better with longer inputs. \textit{Context window expansion}. The fixed context length in pre-trained LLMs limits their understanding of long-term content. Recent work~\cite{chen2024longvila,zhang2024long,chen2025scaling} extends this window by careful system design, enabling models to handle and reason over longer video sequences. \textit{Retrieval-augmented video understanding}. To process long videos, some approaches retrieve only the most relevant segments from a larger collection~\cite{korbar2024text,pan2025timesearch,wang2024videoagent} and use them as context for further analysis.\textit{Visual token reduction or compression}. Other methods shorten the input by reducing visual tokens across or within frames~\cite{shen2024longvu,li2024videochat,jiang2025token,li2025lion, chai2024auroracap}, making it easier to handle long video sequences. While these methods improve performance, they largely treat continuous videos as standard sequence modeling problems, similar to text. We believe future MLLMs should build internal predictive models to efficiently process continuous visual streams, as humans do.

\paragraph{Visual Spatial Intelligence}
Understanding spatial relationships from visual inputs is crucial for perceiving and interacting with the physical world. As multimodal models become more physically grounded, interest in spatial intelligence has surged, leading to new benchmarks~\cite{yang2024think,ramakrishnan2024does,yin2025spatial,majumdar2024openeqa,yeh2025seeing,li2025sti,xu2025multi,team2025gemini} and research focused on enhancing models' spatial reasoning capabilities~\cite{yang2025mindjourney,ma2025spatialreasoner,ouyang2025spacer,du2024embspatial,chen2024spatialvlm,cheng2024spatialrgpt,cai2024spatialbot,liu2024coarsecorrespondenceelicit3d,li2024topviewrs,zhu2024llava3d,ray2025sat}. In this paper, we study visual spatial intelligence through the concept of spatial supersensing in videos and explore ways to strengthen MLLMs' spatial reasoning by refining data curation, optimizing training strategies, and introducing new paradigms.

\paragraph{Predictive Modeling} 
A learned internal predictive model~\cite{craik1967nature,ha2018world} allows an intelligent agent to represent and simulate aspects of its environment, enabling more effective planning and decision-making. Model predictive control (MPC)~\cite{garcia1989model} applies similar principles in control theory, leveraging internal forward models to anticipate future trajectories and select optimal actions in real time. This concept draws inspiration from how humans form mental models of the world~\cite{rao1999predictive,hohwy2013predictive,friston2010free} and how these internal representations influence behavior (\eg, \emph{unconscious inference}~\cite{von1867handbuch}), serving as simplified abstractions of reality that enable prediction and efficient action. A growing body of work has explored the idea of predictive modeling through self-supervised representation learning~\cite{assran2023self,assran2025v}, and text- or action-conditioned video generation~\cite{zhou2024dino,yang2023learning,bar2025navigation,chen2024simple,bai2025whole,garrido2025intuitive}.
In this paper, motivated by how humans leverage internal world models to process unbounded sensory input efficiently and effectively, we investigate how to equip MLLMs with a similar predictive sensing capability.


\section{Conclusion}
We highlight the importance of and propose a hierarchy for spatial \emph{supersensing} capabilities in videos, arguing that achieving superintelligence requires AI systems to move beyond text-based knowledge and semantic perception, the current focus of most MLLMs, to also develop spatial cognition and predictive world models. To measure progress, we introduce $\vsisuper$ and find that current MLLMs struggle with it. To test whether current progress is limited by data, we curate \vsidata and train our spatially grounded MLLM, \cambrianS, on it. Although \cambrianS performs well on standard benchmarks, its results on \vsisuper reveal the limitations of the current MLLM paradigm. We prototype predictive sensing, using latent frame prediction and surprise estimation to handle unbounded visual streams. It improves \cambrianS performance on \vsisuper and marks an early step toward spatial supersensing.

\noindent\textbf{Limitations.} 
Our goal is to present a conceptual framework that encourages the community to reconsider the importance of developing spatial supersensing. As a long-term research direction, our current benchmark, dataset, and model design remain limited in quality, scale, and generalizability, and the prototype serves only as a proof of concept. Future work should explore more diverse and embodied scenarios and build stronger connections with recent advances in vision, language, and world modeling.


\section*{Acknowledgments}
We are grateful to Cambrian-1~\cite{tong2024cambrian} for the excellent codebase, which served as the launching point for our research. Thanks to the TorchXLA team for helpful discussions
on TPU, TorchXLA, and JAX distributed training infrastructure. 
We also thank Anjali Gupta, Sihyun Yu, Oscar Michel, Boyang Zheng, Xichen Pan, Weiyang Jin, and Arijit Ray for reviewing this manuscript and providing constructive feedback. This work was primarily supported by the Google TPU Research Cloud (TRC) program and
the Google Cloud Research Credits program (GCP19980904).
E.B. is supported by the DoD NDSEG Fellowship Program.
S.X. acknowledges support from the MSIT IITP grant (RS-2024-00457882) and the NSF award IIS-2443404.


\clearpage

\addcontentsline{toc}{section}{References}
\bibliography{arxiv}
\bibliographystyle{plain}


\clearpage
\appendix

\section*{Appendix}\label{app:appendix}

This appendix provides comprehensive implementation details, experimental results, and supplementary analyses supporting the main paper:
\begin{itemize}[noitemsep,topsep=0pt,parsep=0pt,partopsep=0pt,leftmargin=1.5em]
    \item \S~\ref{appx:benchmark_diagnostic_test_results} presents detailed diagnostic test results for video MLLM benchmarks under different evaluation setups.
    \item \S~\ref{appx:vsisuper} describes the \vsisuper benchmark, including implementation details, visualizations, and streaming setups for both Recall and Count tasks.
    \item \S~\ref{appx:vsidata} provides comprehensive documentation of the \vsidata dataset, including question type taxonomy, QA-pair construction pipeline, ablation studies, and qualitative examples.
    \item \S~\ref{appx:impl_details} details the \cambrianS model architecture, training data mixture, training recipe across all four stages, and infrastructure setup.
    \item \S~\ref{appx:cambrians_additional_results} presents additional experimental results including detailed evaluation setups, performance on image and video benchmarks across all model scales, ablations on image-video data contributions, and analysis of the trade-off between spatial sensing and general video understanding.
    \item \S~\ref{appx:pred_sensing} describes predictive sensing components, including latent frame prediction implementation details, memory framework design for \vsisuper Recall, agentic framework design for \vsisuper Count, and comparisons with existing long-video methods.
\end{itemize}

\section{Benchmark Diagnostic Test Results}\label{appx:benchmark_diagnostic_test_results}

We provide detailed results of \cref{fig:benchmark_analysis} in \cref{tab:detailed_benchmark_analysis}.

\begin{table}[h]
    \centering
    \caption{
        \textbf{Detailed results of our improved Cambrian-1-7B on video MLLM benchmarks under different evaluation setups.}
    }\label{tab:detailed_benchmark_analysis}
    \resizebox{0.85\textwidth}{!}{
    \begin{tabular}{l|ccccccccccc}
        \toprule
            Evaluation Setups &
            \rotatebox{90}{VideoMME} &
            \rotatebox{90}{EgoSchema} &
            \rotatebox{90}{VideoMMMU} &
            \rotatebox{90}{LongVideoBench} &
            \rotatebox{90}{Tomato} &
            \rotatebox{90}{MVBench} &
            \rotatebox{90}{Perception Test} &
            \rotatebox{90}{HourVideo} &
            \rotatebox{90}{VSI-Bench} &
            \rotatebox{90}{\vsisuper Recall} &
            \rotatebox{90}{\vsisuper Count} \\
        \hline
        \textit{Chance-Level} & 25.0 & 20.0 & 14.0 & 25.0 & 22.0 & 27.3 & 33.3 & 20.0 & 34.0 & 25.0 & 0.0 \rule{0pt}{10pt} \\
        \hline
        \rowcolor{navyblue!5}
        \textit{Cambrian-1-7B (Our upgraded)} & & & & & & & & & & & \\
        Blind Test & 31.2 & 31.9 & 25.0 & 42.5 & 7.8 & 19.6 & 40.7 & 24.3 & 17.4 & 20.0 & 0.0 \\
        Single Frame & 41.6 & 44.0 & 29.0 & 46.9 & 15.8 & 46.1 & 52.1 & 27.7 & 20.4 & 19.7 & 0.0 \\
        Multiple (32) Frames & 53.7 & 48.1 & 31.9 & 51.4 & 18.9 & 55.4 & 55.6 & 31.6 & 25.8 & 22.7 & 0.0 \\
        (32) Frame Captions & 55.3 & 52.4 & 40.1 & 52.2 & 16.8 & 47.7 & 55.6 & 29.5 & 21.8 & 9.6 & 0.5 \\
        \bottomrule
    \end{tabular}}
\end{table}

\section{\vsisuper{} Benchmark}\label{appx:vsisuper}

\subsection{\vsisuper{} Recall}

\paragraph{Implementation details.}
To construct this benchmark, we begin with videos from the VSI-Bench collection~\cite{yang2024think}. Annotators select videos and manually insert an unusual object from a curated pool into four distinct frames using Gemini-2.0-Flash, focusing on placing the objects in plausible locations. For each insertion, the annotators record the object's location and its order of appearance. We then combine these edited clips with randomly sampled unedited videos to produce final videos with lengths of 10, 30, 60, 120, and 240 minutes. For each duration, we create 60 videos, each with one corresponding question. We downsample videos to 1 frame per second to ensure the model can always see the edited frames during inference.

\paragraph{Visualization.}
We present qualitative examples of edited frames of our \vso video dataset in \cref{fig:sor_examples}. The inserted objects appear visually plausible at their locations, which is a direct result of our high-quality annotations.

\subsection{\vsisuper{} Count}

\paragraph{Implementation Details.}
To build \vsisuper{} Count, we concatenate videos from VSI-Bench~\cite{yang2024think} and sum their object counts to create a new ground truth. This process requires two additional normalization steps. First, we unify the object category labels from the different source datasets  (\ie, ScanNet~\cite{dai2017scannet}, ScanNet++~\cite{yeshwanth2023scannet++}, and ARKitScenes~\cite{dehghan2021arkitscenes}). Second, we address a data bias towards small object quantities by rebalancing the question-answer pairs to create a more uniform distribution of counts. The final benchmark includes videos with lengths of 10, 30, 60, and 120 minutes, each accompanied by 50 corresponding questions. Different from \vso, all videos in \vsc{} are downsampled to 24 FPS.

\paragraph{Streaming setups.} For the streaming setup, we repeatedly query the total number of objects in a video at 10 distinct timestamps. To construct the ground truth at these query timestamps, we need to determine the first appearance time of each unique object in the video. To find these appearance times, we use the method proposed by the VSI-Bench~\cite{yang2024think}. This allows for the direct calculation of the ground truth object count at any given timestamp.

\section{\vsidata{} Dataset}\label{appx:vsidata}

In this section, we provide more details for our \vsidata{} dataset, including the question type definition, question-answer pair construction pipeline, and some examples for each data source. 

\subsection{Details of Question Type Definition}\label{appx:vsidata:question_type_definition}

\paragraph{Taxonomy.}
When curating visual-spatial intelligence supervised fine-tuning datasets, an important perspective is how to define the question type. Inspired by VSI-Bench~\cite{yang2024think}, we expand its task definition in a more systematic manner. As shown in \cref{tab:vsidata:question_type_definition}, we distinguish these question types in four perspectives:
\begin{itemize}
    \item \texttt{Spatial-temporal attributes:} We categorize questions into five distinct spatial-temporal attribute types: size (comparing or measuring object/space dimensions), direction (orientation in space), count (enumeration of objects), distance (proximity between objects), and appearance order (temporal sequence of objects appearing in videos).
    \item \texttt{Relative versus absolute:} Questions are classified as relative when they involve comparison between multiple objects (\eg, ``which is larger?''), or absolute when they require specific measurements or quantities (\eg, ``what is the height in meters?''). This distinction applies across most attribute types.
    \item \texttt{Perspective taking:} This dimension captures the viewpoint from which spatial relationships are evaluated. Questions may be posed from the camera's perspective (\eg, ``from the camera's perspective, is the object on the left or right?'') or from the perspective of specific objects in the scene (\eg, ``facing the object$_1$ from object$_2$...'')
    \item \texttt{Modality:} Questions are categorized based on whether they can be answered using static images only, or require dynamic video information. Some attribute types, like appearance order, are only applicable to videos, while others like size can be questioned in either modality.
\end{itemize}
Additionally, following VSI-Bench, we also categorize our question types into three different groups (\ie, \textit{Configuration}, \textit{Measurement}, or \textit{Spatiotemporal}) according to their different spatiotemporal characteristics.

\begin{table}[h]
    \centering
    \caption{\textbf{Taxonomy of spatiotemporal question types in \vsidata.} Questions are stratified along five axes: attribute type, relative vs. absolute (Rel./Abs.), perspective, modality (V: video, I: image), and group. An example question template is provided for each type.}
    \label{tab:vsidata:question_type_definition}
    \resizebox{1\textwidth}{!}  
    {
    \renewcommand{\arraystretch}{1}
    \begin{tabular}{lccccp{8cm}}
        \toprule
        \textbf{Types} & \textbf{Rel./Abs.} & \textbf{Perspective} & \textbf{Modality} & \textbf{Group} & \textbf{Example template} \\
        \midrule
        \multirow{3}{*}{Size} 
            & Rel. & — & V \& I & Configuration & ``Between \{\textit{object$_1$}\} and \{\textit{object$_2$}\}, which is larger?'' \\ 
            & Abs. & — & V \& I & Measurement & ``What is the height of the \{\textit{object}\} in \{\textit{unit}\}?'' \\
            & Abs. & — & V \& I & Measurement & ``What is the room's size in \{\textit{unit}\}?'' \\
        \midrule
        \multirow{5}{*}{Direction} 
            & Rel. & Camera & I & Configuration & ``From the camera's perspective, is the \{\textit{object}\} on the left or the right?'' \\
            & Rel. & Object & V \& I & Configuration & ``Facing the \{\textit{object$_1$}\} from the \{\textit{object$_2$}\}, would the \{\textit{object$_3$}\} be placed left, right, or back?'' \\
            & Abs. & Object & V \& I & Measurement & ``Standing at \{\textit{object$_1$}\}, facing toward \{\textit{object$_2$}\}, how far clockwise do I rotate (in degrees) to see the \{\textit{object$_3$}\}?'' \\
        \midrule
        \multirow{2}{*}{Count} 
            & Rel. & — & V \& I & Configuration & ``Are there fewer \{\textit{object$_1$}\}  than \{\textit{object$_2$}\} ?'' \\
            & Abs. & — & V \& I & Measurement & ``How many \{\textit{object}\}  are present?'' \\
        \midrule
        \multirow{6}{*}{Distance} 
            & Rel. & Camera & I & Configuration & ``Which object is closer to the camera, the \{\textit{object\_1}\} or the \{\textit{object\_2}\}?'' \\
            & Rel. & Object & V \& I & Configuration & ``Which is nearer to the \{\textit{object\_3}\}, the \{\textit{object\_1}\} or the \{\textit{object\_2}\}?'' \\
            & Abs. & Object & V \& I & Measurement & ``What is the distance between the \{\textit{object\_1}\} and the \{\textit{object\_2}\} in \{\textit{unit}\}?'' \\
        \midrule
        \multirow{3}{*}{Appr. Order}
            & \multirow{3}{*}{—}  & \multirow{3}{*}{—} & \multirow{3}{*}{V} & \multirow{3}{*}{Spatiotemporal} & ``Determine how \{\textit{object\_1}\}, \{\textit{object\_2}\}, \{\textit{object\_3}\}, and \{\textit{object\_4}\} are ordered by their initial appearances in the video'' \\
        \bottomrule
    \end{tabular}
    }
\end{table}

\begin{figure}[h]
    \centering
    \begin{minipage}[t]{0.49\textwidth}
        \centering
        \includegraphics[width=\linewidth]{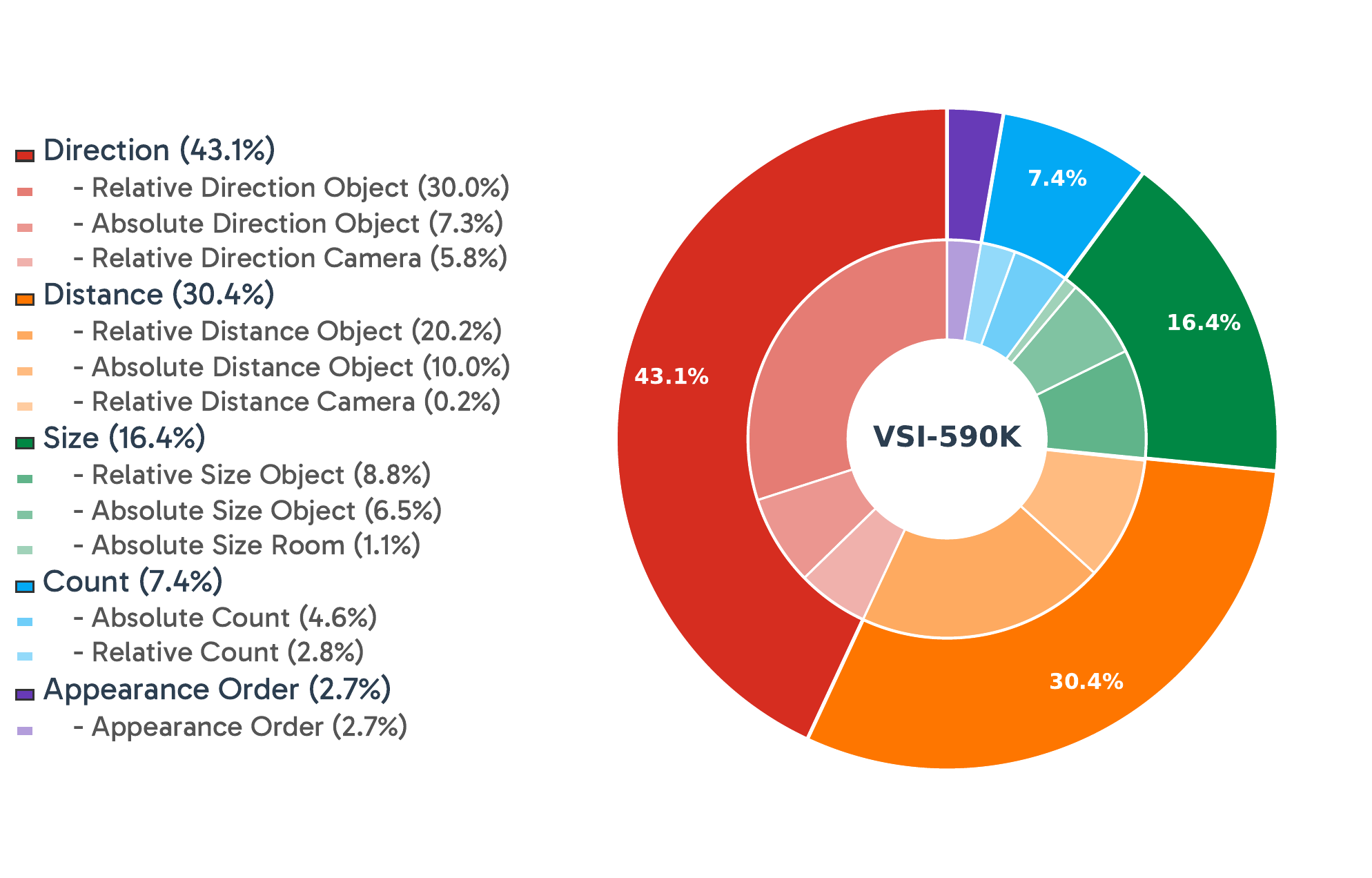}
    \end{minipage}\hfill
    \begin{minipage}[t]{0.49\textwidth}
        \centering
        \includegraphics[width=\linewidth]{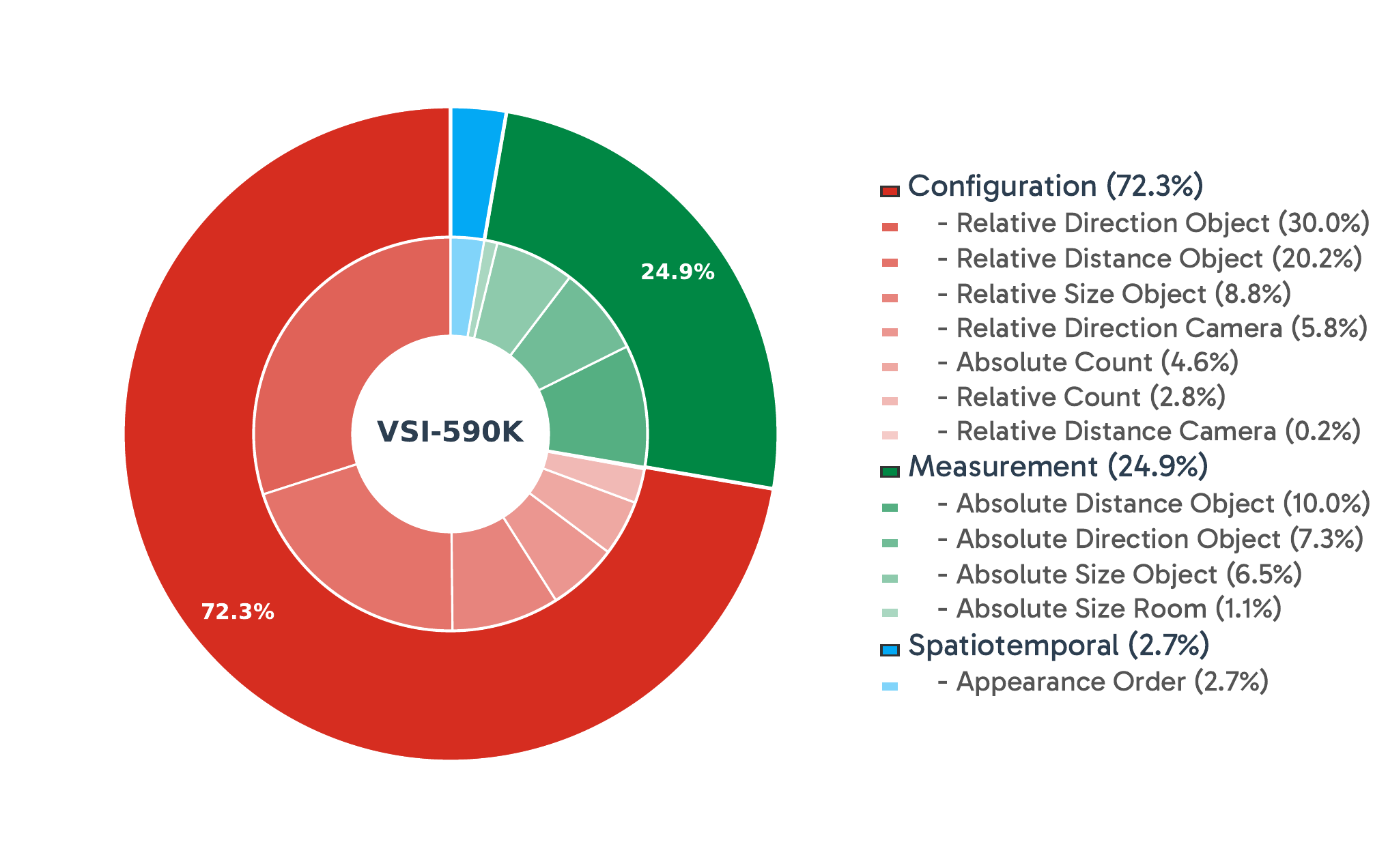}
    \end{minipage}
    \caption{
        \textbf{\vsidata{} dataset statistics.}
        QAs are grouped by: question types (left) and task groups (right).
    }\label{fig:vsi_piechart}
\end{figure}

\subsection{Detailed QA-Pair Construction Pipeline}

We introduce the concrete pipeline used for curating \vsidata{} here.

\paragraph{3D-annotated real videos.}
For the 3D-annotated real videos, we follow the practice established by Thinking in Space~\cite{yang2024think}. We begin by researching all publicly available datasets containing both 3D instance-level annotations and video or panorama images. From these datasets, we extract key information including \textit{object counts}, \textit{object bounding boxes}, and \textit{room size} measurements, which we then standardize into a unified format. Afterward, this structured information is incorporated into augmented question templates to create paired question-answer sets.

\paragraph{3D-annotated simulated videos and images.}
For simulated data, which inherently contains rich annotations, we followed a procedure similar to that used for 3D-annotated real videos. As for ProcTHOR~\cite{deitke2022ProcTHOR}, our primary effort is generating 3D scenes with randomly placed agents to render traverse videos. For Hypersim~\cite{roberts2021hypersim}, which provides image-level rather than scene-level 3D annotations, we utilize individual images with their corresponding 3D annotations. In both cases, we extract the necessary information, convert it to our designed unified format, and incorporate it into augmented question templates, following the same approach used for 3D-annotated real videos.

\paragraph{Unannotated web-crawled real videos.} For unannotated web-crawled real videos, as shown in Algorithm~\ref{algo:vsidata:qa_gen_web_video}, we implement a multi-stage processing pipeline. We begin by sampling frames at regular intervals and filtering out blurry images. For each valid frame, we employ the open-vocabulary object detector Grounding-DINO~\cite{liu2024grounding} with predefined categories of interest. When a frame contains sufficient valid objects, we use SAM2~\cite{ravi2024sam} to extract instance-wise semantic masks.
Besides, to transform 2D image content into 3D representations, we employ VGGT~\cite{wang2025vggt} to extract 3D point sets for each image and integrate them with the previously generated instance masks. Notably, we apply an erosion algorithm to refine the instance masks, which mitigates inaccurate point cloud estimations at object boundaries. This pipeline has enabled us to create pseudo-annotations from approximately 19,000 room tour videos from YouTube and robotic learning datasets, yielding diverse spatial question-answer pairs across various room types and layouts without manual 3D annotations. By processing individual frames rather than complete videos, our pipeline ensures higher quality semantic extraction and more reliable reconstruction results, avoiding the noise and inconsistent issues typically encountered when applying reconstruction and semantic extraction techniques to entire video sequences.

\subsection{Additional Ablation Study}

\begin{table}[!h]
    \caption{
        \textbf{Ablation study on \vsidata task groups.} We study models' performance change when one certain task group are omitted from the training data.
    }\label{tab:vsidata:task_group_ablation}
    \centering
    \setlength\tabcolsep{2pt} 
    \setlength{\tabcolsep}{0.5em}
    \resizebox{0.65\textwidth}{!}{
    \begin{tabular}{l |c cccc cccc}
        \toprule
        &
            \multicolumn{9}{c}{VSI-Bench} \\
        \vsidata Mixture
        &
            \rotatebox{90}{Avg} &
            \rotatebox{90}{Obj Ct} &
            \rotatebox{90}{Abs Dst} &
            \rotatebox{90}{Obj Sz} &
            \rotatebox{90}{Rm Sz} &
            \rotatebox{90}{Rel Dst} &
            \rotatebox{90}{Rel Dir} &
            \rotatebox{90}{Rte Pln} &
            \rotatebox{90}{Ap Ord} \\
        \hline
        \rowcolor{navyblue!5}
        \rule{0pt}{10pt}
        All & 63.2 & 73.5 & 49.4 & 71.4 & 70.1 & 66.9 & 61.5 & 36.6 & 76.4 \\
        \hline
        \rule{0pt}{10pt}
        \emph{w/o.} Configuration & 51.9 & 46.2 & 43.0 & 70.4 & 66.0 & 48.0 & 36.8 & 27.3 & 77.3 \\
        \rule{0pt}{10pt}
        \emph{w/o.} Measurement & 49.7 & 74.5 & 19.1 & 31.1 & 38.5 & 63.9 & 55.6 & 35.1 & 79.5 \\
        \rule{0pt}{10pt}
        \emph{w/o.} Spatiotemporal & 58.1 & 73.7 & 47.7 & 70.9 & 65.2 & 68.3 & 58.9 & 32.5 & 47.6 \\
        \bottomrule
    \end{tabular}}
\end{table}

\cref{tab:vsidata:task_group_ablation} presents an ablation study on how different task groups affect the model's spatial sensing capability. Our results show that all three task groups—configuration, measurement, and spatiotemporal—are integral, as removing any one of them degrades performance. We further assess spatial reasoning using the held-out \textit{Route Plan} subtask and find that the configuration group is the most influential, whereas the measurement group is the least. We attribute this outcome to the fact that route planning requires a holistic understanding of the spatial layout, which is more explicitly provided by configuration QA pairs compared to measurement and spatiotemporal tasks.

\subsection{Examples of \vsidata}

To better illustrate \vsidata, we provide qualitative visualization results in~\cref{fig:annotated_real_video_1,fig:annotated_real_video_2,fig:annotated_real_video_3,fig:annotated_simulated_video,fig:annotate_simulated_video_frame,fig:unannotated_image_1,fig:unannotated_image_2}. These visualizations demonstrate that \vsidata delivers great diversity and quality for spatial question-answering supervised fine-tuning.

\begin{algorithm}[!h]
    \small
    \caption{
        \textbf{QA generation pipeline for unannotated web-scrawled videos}
    }\label{algo:vsidata:qa_gen_web_video}
    \KwIn{Video sequence $V$, valid category list $\mathcal{C}_{\text{valid}}$, invalid category list $\mathcal{C}_{\text{invalid}}$, sampling interval $\Delta t$, blur threshold $\tau_{\text{blur}}$, minimum object count $\theta_{\text{min}}$, minimum 3D point count $\theta_{\text{3D}}$, erosion kernel $K_{\text{erosion}}$}
    \KwOut{Selected frame set $\mathcal{F}$, Question-answer pairs $\mathcal{Q}$}
    
    Initialize $\mathcal{F} \leftarrow \emptyset$, $\mathcal{Q} \leftarrow \emptyset$\;
    $\mathcal{S} \leftarrow \text{SampleFrames}(V, \Delta t)$ \tcp*{Sample frames at interval $\Delta t$}
    
    \ForEach{frame $f \in \mathcal{S}$}{
        \If{$\text{BlurDetection}(f) > \tau_{\text{blur}}$}{
            \textbf{continue}\;
        }
        $\mathcal{O} \leftarrow \text{GroundingDINO}(f, \mathcal{C}_{\text{valid}} \cup \mathcal{C}_{\text{invalid}})$ \tcp*{Detect objects from both category lists}
        
        \If{$\exists o \in \mathcal{O} : \text{category}(o) \in \mathcal{C}_{\text{invalid}}$}{
            \textbf{continue}\;
        }
        
        $\mathcal{O}_{\text{valid}} \leftarrow \{o \in \mathcal{O} : \text{category}(o) \in \mathcal{C}_{\text{valid}}\}$\;
        \If{$|\mathcal{O}_{\text{valid}}| < \theta_{\text{min}}$}{
            \textbf{continue}\;
        }
        $\mathcal{M} \leftarrow \emptyset$ \tcp*{Initialize mask set}
        
        \ForEach{object $o \in \mathcal{O}_{\text{valid}}$}{
            $b \leftarrow \text{GetBoundingBox}(o)$\;
            $m \leftarrow \text{SAM2}(f, b)$ \tcp*{Generate mask using SAM2}
            $m' \leftarrow \text{Erode}(m, K_{\text{erosion}})$ \tcp*{Apply erosion on the masks}
            $\mathcal{M} \leftarrow \mathcal{M} \cup \{m'\}$\;
        }
        
        $\mathcal{P}_{\text{map}} \leftarrow \text{VGGT}(f)$ \tcp*{Generate 3D point map using VGGT}
        $\mathcal{P} \leftarrow \emptyset$ \tcp*{Initialize 3D point set}
        
        \ForEach{mask $m \in \mathcal{M}$}{
            $P \leftarrow \text{ExtractMaskedPoints}(m, \mathcal{P}_{\text{map}})$ \tcp*{Extract 3D points covered by mask}
            \If{$|P_{\text{valid}}| \geq \theta_{\text{3D}}$}{
                $\mathcal{P} \leftarrow \mathcal{P} \cup \{P\}$\;
            }
        }
                
        \If{$|\mathcal{P}| > 0$}{
            $q \leftarrow \text{QAGenerator}(\mathcal{P})$ \tcp*{Generate QA pairs from 3D geometry}
            $\mathcal{Q} \leftarrow \mathcal{Q} \cup \{q\}$\;
            $\mathcal{F} \leftarrow \mathcal{F} \cup \{f\}$\;
        }
    }
    \textbf{Return} $\mathcal{F}$, $\mathcal{Q}$\;
\end{algorithm}

\section{\cambrianS Implementation Details}\label{appx:impl_details}

In this section, we provide holistic training details of our \cambrianS{} models.

\subsection{Model Architecture}

Following the original Cambrian-1~\cite{tong2024cambrian} and common practices in most MLLMs~\cite{liu2023visual,li2024llava}, our model (both our upgraded Cambrian-1 and \cambrianS) integrates a pre-trained vision encoder, a pre-trained language model as the decoder, and a vision-language connector to bridge these two modalities. Specifically, we employ SigLIP2-So400M~\cite{tschannen2025siglip} as the vision encoder. This encoder was trained using a combination of losses: text next-token-prediction (LocCa~\cite{wan2024locca}), image-text contrastive (or sigmoid~\cite{radford2021learning, zhai2023sigmoid}), and masked self-prediction (SILC~\cite{naeem2024silc}/TIPS~\cite{maninis2024tips}). For the language model, we utilize the instruction-tuned Qwen2.5 LLMs~\cite{yang2024qwen2.5}. Unlike Cambrian-1, which used SVA for a deeper vision-language fusion, we employ a simpler GELU-activated~\cite{dauphin2017language} two-layer MLP as the vision-language connector to maintain a balance between performance and efficiency.

\subsection{Training Data Mixture}

\begin{figure}[!h]
    \centering
    \includegraphics[width=1.0\linewidth]{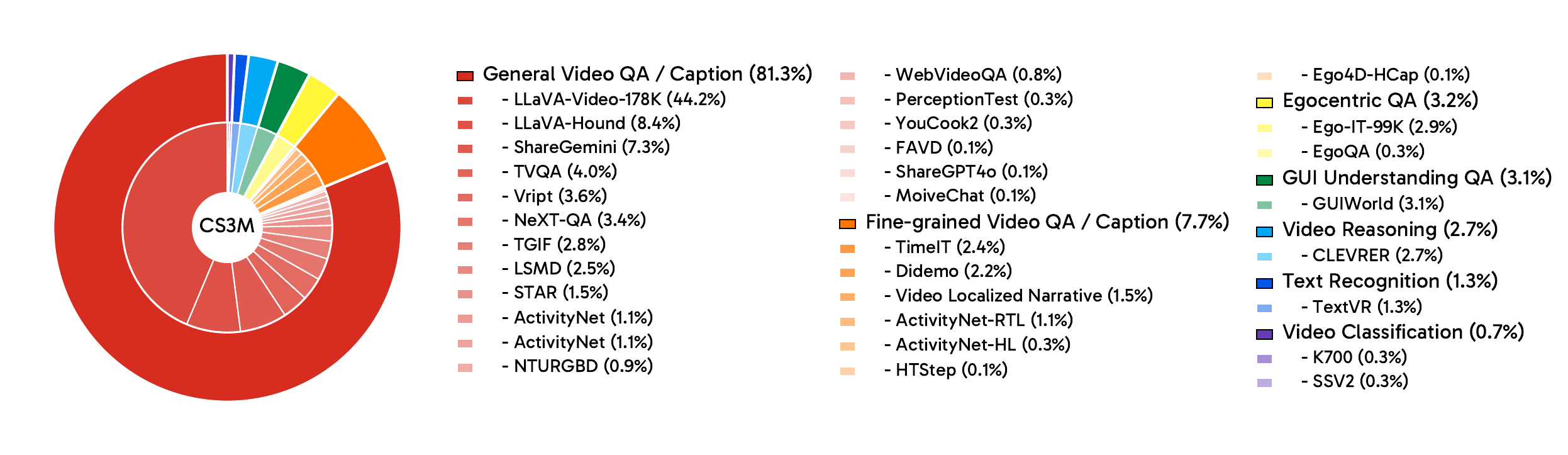}
    \caption{\textbf{General video instruction tuning datasets of \osmix, used in \cambrianS stage 3 \& 4 training.}}
    \label{fig:osmix_sources}
\end{figure}

As mentioned in \cref{sec:limits:cambrian-s}, our \cambrianS models are trained with four training stages (See~\cref{fig:training_stage_simple_illustration}). For the first two stages (\ie, vision-language alignment stage and image instruction tuning stage), we refer readers to Cambrian-1~\cite{tong2024cambrian} for the detailed training data mixture.
In the third stage, we finetune the image instruction-tuned models \osmix, and during the last stage, we conduct spatial video instruction tuning by finetuning the model on \vsidata.
\osmix is our curated video instruction tuning dataset with around 3M video QA samples, built upon a set of open-sourced video datasets (\eg,  LLaVA-Video~\cite{zhang2024video}, ShareGPT4o~\cite{cui2025comprehensive},VideoChat2~\cite{li2024mvbench}, MovieChat~\cite{song2024moviechat}, EgoIT~\cite{yang2025egolife}, Perception Test~\cite{patraucean2023perception}, Vript~\cite{yang2024vript},VideoChatGPT-Plus~\cite{maaz2024videogpt+}, Ego4D~\cite{grauman2022ego4d}, HowTo100M~\cite{miech2019howto100m},HD-VILA~\cite{xue2022advancing}, HTStep~\cite{afouras2023ht}, TimeIT~\cite{ren2024timechat}, HowToInterlink7M \cite{wang2024cosmo}, GUI-World~\cite{chen2024gui}, Video-Localized-Narratives \cite{voigtlaender2023connecting}, and \etc).
We detail its composition in \cref{fig:osmix_sources}.

\subsection{Training Recipe}\label{appx:cambrians_training_recipe}

\begin{table}[!h]
    \centering
    \caption{
        \textbf{Training configuration for stage 1 and stage 2.}
    }\label{tab:improved_cambrian_training_setups}
    \resizebox{0.75\textwidth}{!}{
        \begin{tabular}{l|cc}
        \toprule
        & \text{Stage 1} (Vision-Language Alignment) & \text{Stage 2} (Image Instruction Tuning) \\
        \hline
        \rowcolor{gray!10}
        \textit{Model} & & \\
        Vision Encoder & \multicolumn{2}{c}{SigLIP2-So400M} \\
        Language Decoder & \multicolumn{2}{c}{Qwen2.5-{0.5B, 1.5B, 3B, 7B}-Instruct} \\
        VL-Connector & \multicolumn{2}{c}{2$\times$MLP-GELU} \\
        \hline
        \rowcolor{gray!10}
        \textit{Data Recipe} & & \\
        Data & Cambrian-Alignment-2.5M & Cambrian-7M \\
        Image Resolution & Pad (384$\times$384) & AnyRes~(Up to 9 sub-images) \\
        \# of Tokens per Image & 729 & Up to 7,290 \\
        \hline
        \rowcolor{gray!10}
        \textit{Training Recipe} & & \\
        Max Sequence Length & 2,048 & 8,192 \\
        Trainable Module & VL-Connector & VL-Connector \& LLM \\
        Learning Rate & $1\times10^{-3}$ & $1\times10^{-5}$ \\
        Batch Size & 512 & 256 \\
        Warmup Ratio & 0.06 & 0.03 \\
        \bottomrule
        \end{tabular}
    }
\end{table}

\paragraph{Stage 1: Vision-language alignment.}

We freeze most of the model's parameters and train only the vision-language connector on the Cambrian-Alignment-2.5M dataset~\cite{tong2024cambrian}. Input images are padded to a fixed resolution of $384\times384$, and the maximum sequence length is set to 2048.

\paragraph{Stage 2: Image instruction tuning.}

We unfreeze both the vision-language connector and the LLM decoder, while keeping the vision encoder frozen. The model is then fine-tuned on the Cambrian-7M image instruction tuning dataset. Compared to Cambrian-1~\cite{tong2024cambrian}, we adopt the AnyRes strategy~\cite{liu2024improved} to enhance the model's image understanding capabilities. Specifically, input images are resized while preserving aspect ratio, then divided into multiple $384\times384$ sub-images. This enables the model to handle images with higher and more flexible resolutions.
To accommodate the increased number of visual tokens introduced by the AnyRes strategy, we extend the sequence length to 8192.
Detailed training configurations for stage 1 and 2 are provided in \cref{tab:improved_cambrian_training_setups}.

\begin{table}[!h]
    \centering
    \caption{
        \textbf{Training recipe for \cambrianS stage 3 and stage 4.}
    }\label{tab:cambrian_s_training_setups}
    \resizebox{0.95\textwidth}{!}{
        \begin{tabular}{l|cc}
        \toprule
        & \textbf{Stage 3} (General Video Instruction Tuning) & \textbf{Stage 4} (Spatial Video Instruction Tuning) \\
        \hline
        \rowcolor{gray!10}
        \textit{Model} & & \\
        Vision Encoder & \multicolumn{2}{c}{SigLIP2-So400M} \\
        Language Decoder & \multicolumn{2}{c}{Qwen2.5-{0.5B, 1.5B, 3B, 7B}-Instruct} \\
        VL-Connector & \multicolumn{2}{c}{2$\times$MLP-GELU} \\
        \hline
        \rowcolor{gray!10}
        \textit{Data Recipe} & & \\
        Data Source & \osmix{} & \vsidata{} + 590K general Video IT data (sampled from \osmix) \\
        Video Frame Resolution & Pad (384$\times$384) & Pad (384$\times$384) \\
        Frame Sampling Strategy & Uniform & Uniform \\
        \# Frames per Video & 64 & 128 \\
        \# Tokens per Video Frame & 64 & 64 \\
        \hline
        \rowcolor{gray!10}
        \textit{Training Recipe} & & \\
        Max Sequence Length & 8,192 & 16,384 \\
        Trainable Modules &\multicolumn{2}{c}{VL-Connector and LLM} \\
        Learning Rate &\multicolumn{2}{c}{$1\times10^{-5}$} \\
        Global Batch Size &\multicolumn{2}{c}{256} \\
        Warmup Ratio &\multicolumn{2}{c}{0.03} \\
        \bottomrule
        \end{tabular}
    }
\end{table}

\paragraph{Stage 3: General video instruction tuning.}

To equip the model with general video understanding capabilities, we perform video instruction tuning on a mixture of curated \osmix video data and sampled image instruction data from Cambrian-7M. As in previous stages, the vision encoder remains frozen, and the remaining modules are fine-tuned.
For image data, we reuse the sampling strategy from stage 2. For video data, we uniformly sample 64 frames per video, resize them to $384\times384$, and further downsample their feature maps to $8\times8$, \ie, 64 tokens per frame.

\paragraph{Stage 4: Spatial video instruction tuning.}

The final stage focuses on enhancing the model's spatial reasoning capabilities by fine-tuning on our proposed \vsidata. To preserve general video and image understanding, we mixed 590K video samples from \osmix and 120K image samples from Cambrian-7M.
Training settings are mostly consistent with stage 3, except for two key changes: (1) we increase the number of frames per video to 128, and (2) we extend the sequence length to 16,384, both to support richer temporal modeling.
Detailed configurations for stage 3 and 4 are listed in \cref{tab:cambrian_s_training_setups}.

\subsection{Infrastructure}

All models in this paper are trained using TPU v4 Pods with the TorchXLA framework.
To support large-scale video instruction tuning—where long sequence lengths introduce prohibitive computational and memory costs—we leverage GSPMD~\cite{xu2021gspmd} and FlashAttention~\cite{dao2022flashattention} implemented by Pallas.

GSPMD is an automatic parallelization system designed for flexible and user-friendly large-scale distributed training. It allows users to write training code as if for a single device, and then scale effortlessly across hundreds of devices with minimal changes.
Our training framework is based on TorchXLA and GSPMD to shard data, model parameters, activations, and optimizer states across multiple devices. This reduces the peak memory usage and improves training throughput.

To accommodate long sequences, we integrate FlashAttention backed by Pallas, which significantly reduces TPU HBM (V-Mem) usage under long-context inputs. This enables us to scale the input sequence length up to 16,384 tokens for the 7B model on a TPU v4-512 Pod.

\section{\cambrianS{} Additional Results}\label{appx:cambrians_additional_results}

\subsection{Detailed Evaluation Setups}

We describe the evaluation settings used for most image and video benchmarks, excluding \vsisuper. For image inputs, following the \textit{any-resolution} design adopted in our training pipeline, each image is resized while preserving its aspect ratio, and its resolution is maximized so that it can be partitioned into at most nine 384$\times$384 sub-images. For video inputs, we apply uniform frame sampling with a fixed number of frames. Specifically, checkpoints from stage 1 and stage 2 are evaluated with 32 uniformly sampled frames, while those from stage 3 and stage 4 use 64 and 128 frames, respectively.

\subsection{Detailed Performance on Image and Video Benchmarks}

\cref{tab:image_benchmarks_results} and \cref{tab:video_benchmarks_results} detail the performance of all our checkpoints (from stage 1 to stage 4 and from 0.5B to 7B) on image-based and video-based MLLM benchmarks, respectively. For image benchmarks, we report the results on MME~\cite{yin2024survey}, MMBench~\cite{liu2024mmbench}, SeedBench~\cite{li2024seed}, GQA~\cite{hudson2019gqa}, ScienceQA~\cite{saikh2022scienceqa}, MMMU~\cite{yue2024mmmu}, MathVista~\cite{lu2023mathvista}, AI2D~\cite{kembhavi2016diagram}, ChartQA~\cite{masry2022chartqa}, OCRBench~\cite{liu2024ocrbench}, TextVQA~\cite{zhou2018towards}, DocVQA~\cite{mathew2021docvqa}, MMVP~\cite{tong2024eyes}, RealworldQA~\cite{xai_grok1_5v_2024}, and CVBench~\cite{tong2024cambrian}, following Cambrian-1's grouping strategy.

\begin{table}[!h]
    \centering
    \caption{
        \textbf{Detailed results of \cambrianS checkpoints on image MLLM benchmarks.}
    }\label{tab:image_benchmarks_results}
    \resizebox{\textwidth}{!}{
    \begin{tabular}{l|ccccc |ccccc |ccccc |ccccc}
    \toprule
    & 
    \multicolumn{5}{c}{General} &
    \multicolumn{5}{c}{Knowledge} &
    \multicolumn{5}{c}{OCR \& Chart} &
    \multicolumn{5}{c}{Vision-Centric}  \\
    Method &
    \rotatebox{90}{Avg} &
    \rotatebox{90}{MME$^\text{P}$} &
    \rotatebox{90}{MMB} &
    \rotatebox{90}{SEED$^\text{I}$} &
    \rotatebox{90}{GQA} &
    \rotatebox{90}{Avg} &
    \rotatebox{90}{SQA$^\text{I}$} &
    \rotatebox{90}{MMMU$^\text{V}$} &
    \rotatebox{90}{MathVista$^\text{M}$} & \rotatebox{90}{AI2D} &
    \rotatebox{90}{Avg} &
    \rotatebox{90}{ChartQA} &
    \rotatebox{90}{OCRBench} &
    \rotatebox{90}{TextVQA}  &
    \rotatebox{90}{DocVQA}  &
    \rotatebox{90}{Avg} &
    \rotatebox{90}{MMVP} &
    \rotatebox{90}{RealworldQA} &
    \rotatebox{90}{CV-Bench$^\text{2D}$} &
    \rotatebox{90}{CV-Bench$^\text{3D}$} \\
    \hline
    \rowcolor{gray!10}
    \textit{Open-source Models} & &  &  &  &  &  &  &  &  &  &  &  &  &  &  &  &  &  &  & \rule{0pt}{10pt}
    \\
    Mini-Gemini-HD-8B  & 72.7 & 1606.0 & 72.7 & 73.2 & 64.5 & 55.7 & 75.1 & 37.3 & 37.0 & 73.5 & 62.9 & 59.1 & 47.7 & 70.2  & 74.6 & 51.5 & 18.7 & 62.1 & 62.2 & 63.0 \rule{0pt}{10pt} \\

    LLaVA-NeXT-8B & 72.5 & 1603.7 & 72.1 & 72.7 & 65.2 & 55.6 & 72.8 & 41.7 & 36.3 & 71.6 & 63.9 & 69.5 & 49.0 & 64.6  & 72.6 & 56.6 & 38.7 &  60.1 & 62.2 & 65.3 \rule{0pt}{10pt} \\

   Cambrian-1-8B & 73.1 & 1,547.1 & 75.9 & 74.7 & 64.6 & 61.3 & 80.4 & 42.7 & 49.0 & 73.0 & 71.3 & 73.3 & 62.4 & 71.7 & 77.8 & 65.0 & 51.3 & 64.2 & 72.3 & 72.0 \rule{0pt}{10pt} \\


   \hline
   \rowcolor{navyblue!5} \cambrianS-7B \rule{0pt}{10pt} & & & & & & & & & & & & & & & & & & & & \\
   Stage 1 & 11.5 & 209.9 & 29.6 & 5.6 & 0.1 & 2.5 & 3.1 & 2.2 & 2.9 & 1.7 & 7.4 & 0.9 & 27.6 & 0.9 & 0.1 & 0.9 & 0.0 & 2.7 & 0.8 & 0.0 \\
   Stage 2 & 74.9 & 1604.6 & 79.0 & 76.3 & 64.0 & 63.9 & 83.7 & 48.7 & 45.3 & 78.1 & 79.1 & 78.9 & 67.6 & 79.2 & 90.6 & 66.3 & 53.3 & 67.7 & 70.0 & 74.0 \\
   Stage 3 & 74.4 & 1583.9 & 79.7 & 76.4 & 62.4 & 60.4 & 82.2 & 46.2 & 36.1 & 77.0 & 75.5 & 75.3 & 64.0 & 77.1 & 85.6 & 67.0 & 58.0 & 66.1 & 71.8 & 72.3 \\
   Stage 4 & 74.8 & 1598.4 & 80.4 & 77.0 & 61.8 & 64.6 & 82.7 & 48.0 & 50.6 & 76.9 & 75.2 & 74.7 & 64.8 & 76.6 & 84.8 & 70.5 & 60.0 & 64.8 & 74.3 & 83.0 \\
   \hline
   \rowcolor{navyblue!5} \cambrianS-3B \rule{0pt}{10pt} & & & & & & & & & & & & & & & & & & & & \\
   Stage 1 & 8.3 & 9.3 & 31.7 & 1.0 & 0.0 & 0.9 & 0.9 & 0.9 & 0.1 & 1.7 & 7.0 & 0.0 & 28.1 & 0.0 & 0.1 & 0.7 & 0.0 & 2.1 & 0.7 & 0.0 \\
   Stage 2 & 71.9 & 1524.6 & 74.8 & 74.2 & 62.1 & 55.5 & 78.7 & 42.8 & 27.8 & 72.7 & 72.0 & 69.8 & 63.9 & 71.5 & 82.7 & 59.0 & 37.3 & 62.4 & 65.6 & 70.7 \\
   Stage 3 & 72.1 & 1495.7 & 76.5 & 75.1 & 61.8 & 58.5 & 79.4 & 42.2 & 41.3 & 71.2 & 69.6 & 68.0 & 61.3 & 69.6 & 79.4 & 62.5 & 46.0 & 61.2 & 70.6 & 72.4 \\
   Stage 4 & 71.5 & 1485.6 & 76.0 & 75.1 & 60.8 & 58.7 & 78.7 & 42.1 & 43.0 & 70.9 & 69.6 & 70.0 & 60.5 & 68.7 & 79.1 & 65.6 & 50.0 & 60.1 & 76.1 & 76.3 \\
   \hline
   \rowcolor{navyblue!5} \cambrianS-1.5B \rule{0pt}{10pt} & & & & & & & & & & & & & & & & & & & & \\
   Stage 1 & 11.7 & 282.1 & 28.6 & 0.8 & 3.2 & 3.8 & 6.9 & 4.2 & 1.4 & 2.6 & 7.9 & 1.0 & 27.8 & 1.4 & 1.5 & 0.7 & 0.0 & 0.0 & 2.9 & 0.0 \\
   Stage 2 & 68.5 & 1417.3 & 71.3 & 71.2 & 60.6 & 50.9 & 75.5 & 41.1 & 20.8 & 66.1 & 68.0 & 64.8 & 59.9 & 68.8 & 78.6 & 54.4 & 39.3 & 59.7 & 60.3 & 58.3 \\
   Stage 3 & 68.1 & 1423.2 & 70.5 & 72.1 & 58.7 & 52.6 & 72.4 & 40.8 & 32.3 & 64.8 & 64.2 & 59.5 & 57.6 & 66.7 & 72.9 & 54.6 & 40.0 & 59.9 & 60.7 & 57.8 \\
   Stage 4 & 68.0 & 1394.4 & 70.1 & 73.5 & 58.7 & 54.7 & 72.3 & 42.0 & 39.7 & 64.7 & 65.6 & 63.1 & 58.0 & 66.6 & 74.8 & 59.2 & 43.3 & 54.5 & 62.6 & 76.3 \\
   \hline
   \rowcolor{navyblue!5} \cambrianS-0.5B \rule{0pt}{10pt} & & & & & & & & & & & & & & & & & & & & \\
   Stage 1 & 10.1 & 379.6 & 10.7 & 9.0 & 1.8 & 6.2 & 8.4 & 8.9 & 1.9 & 5.5 & 3.0 & 0.2 & 7.9 & 2.0 & 1.9 & 10.9 & 0.7 & 10.6 & 20.1 & 12.3 \\
   Stage 2 & 57.7 & 1124.3 & 56.6 & 61.7 & 56.1 & 38.6 & 61.5 & 31.0 & 10.5 & 51.5 & 56.0 & 51.1 & 51.0 & 58.7 & 63.1 & 41.2 & 23.3 & 51.8 & 45.6 & 44.1 \\
   Stage 3 & 58.6 & 1200.0 & 55.8 & 63.5 & 55.3 & 41.2 & 62.7 & 32.6 & 18.0 & 51.4 & 52.1 & 46.6 & 46.8 & 56.0 & 59.1 & 45.5 & 22.0 & 52.8 & 52.2 & 54.9 \\
   Stage 4 & 60.0 & 1190.8 & 60.7 & 66.4 & 53.5 & 44.0 & 63.4 & 34.0 & 28.6 & 50.1 & 52.6 & 48.0 & 47.1 & 56.6 & 58.6 & 48.7 & 26.0 & 51.1 & 51.6 & 66.2 \\
    \bottomrule
    \end{tabular}
}
\end{table}

\begin{table}[!h]
    \centering
    \caption{
        \textbf{Detailed results of \cambrianS checkpoints on video MLLM benchmarks.}
    }\label{tab:video_benchmarks_results}
    \resizebox{0.75\textwidth}{!}{
        \begin{tabular}{l|l|ccccccccc}
        \toprule
            Model & Base LLM &
            \rotatebox{90}{VSI-Bench} &
            \rotatebox{90}{Tomato} &
            \rotatebox{90}{HourVideo} &
            \rotatebox{90}{{Video\textsuperscript{MME}}} &
            \rotatebox{90}{{EgoSchema}} &
            \rotatebox{90}{{Video\textsuperscript{MMMU}}} &
            \rotatebox{90}{{LongVBench}} &
            \rotatebox{90}{{MVBench}} &
            \rotatebox{90}{{Percept.\ Test}} \\
        \hline
        \rowcolor{navyblue!5}
        \cambrianS-7B & \multirow{5}{*}{Qwen2.5-7B} & & & & & & & & & \\
        Stage 1 & & 21.4 & 21.0 & 27.5 & 44.3 & 42.9 & 11.3 & 32.3 & 43.9 & 44.4 \\
        Stage 2 & & 24.6 & 20.1 & 31.3 & 52.3 & 47.5 & 28.1 & 51.1 & 49.2 & 53.5 \\
        Stage 3 & & 35.7 & 30.3 & 38.9 & 62.8 & 76.9 & 38.3 & 56.7 & 66.3 & 70.8 \\
        Stage 4 & & 67.5 & 27.9 & 36.5 & 63.3 & 76.3 & 38.3 & 59.4 & 64.8 & 69.8 \\
        \hline
        \rowcolor{navyblue!5}
        \cambrianS-3B & \multirow{5}{*}{Qwen2.5-3B} & & & & & & & & & \\
        Stage 1 & & 0.7 & 16.5 & 0.7 & 15.9 & 19.5 & 8.4 & 23.8 & 30.6 & 18.6 \\
        Stage 2 & & 22.3 & 21.3 & 31.7 & 49.4 & 42.2 & 26.0 & 48.7 & 44.5 & 47.0 \\
        Stage 3 & & 23.3 & 26.3 & 35.9 & 58.9 & 73.4 & 27.1 & 52.0 & 61.0 & 65.7 \\
        Stage 4 & & 57.3 & 26.0 & 36.8 & 60.1 & 73.6 & 26.3 & 52.3 & 60.2 & 65.9 \\
        \hline
        \rowcolor{navyblue!5}
        \cambrianS-1.5B & \multirow{5}{*}{Qwen2.5-1.5B} & & & & & & & & & \\
        Stage 1 & & 21.1 & 23.5 & 26.2 & 40.1 & 33.0 & 18.7 & 38.5 & 40.8 & 45.2 \\
        Stage 2 & & 22.6 & 24.6 & 34.4 & 47.8 & 38.2 & 20.7 & 46.9 & 45.3 & 49.8 \\
        Stage 3 & & 23.4 & 23.1 & 33.2 & 56.1 & 67.8 & 28.6 & 49.4 & 58.2 & 63.6 \\
        Stage 4 & & 54.8 & 22.2 & 31.2 & 56.4 & 69.0 & 25.0 & 50.2 & 57.1 & 63.2 \\
        \hline
        \rowcolor{navyblue!5}
        \cambrianS-0.5B & \multirow{5}{*}{Qwen2.5-0.5B} & & & & & & & & & \\
        Stage 1 & & 16.7 & 23.6 & 23.4 & 26.4 & 21.5 & 13.1 & 25.0 & 34.3 & 37.0 \\
        Stage 2 & & 19.6 & 20.0 & 27.9 & 37.4 & 29.7 & 17.3 & 39.0 & 40.2 & 46.3 \\
        Stage 3 & & 18.8 & 23.9 & 29.5 & 41.8 & 63.8 & 16.7 & 44.9 & 50.7 & 56.1 \\
        Stage 4 & & 50.4 & 23.8 & 28.1 & 44.0 & 62.4 & 15.9 & 43.8 & 51.8 & 56.0 \\
        \bottomrule
        \end{tabular}
    }
\end{table}

\subsection{Contributions from Image-based and Video-based Instruction Tuning}

\begin{table}[h]
    \centering
    \caption{
        \textbf{Video MLLM performance trained with different proportions of image and video data.}
    }\label{tab:image_video_scale_results}
    \resizebox{0.75\textwidth}{!}{
        \begin{tabular}{c|c|ccccccccc}
        \toprule
        Image data & Video data & \rotatebox{90}{VSI-Bench} &
        \rotatebox{90}{Tomato} &
        \rotatebox{90}{HourVideo} &
        \rotatebox{90}{{Video\textsuperscript{MME}}} &
        \rotatebox{90}{{EgoSchema}} &
        \rotatebox{90}{{Video\textsuperscript{MMMU}}} &
        \rotatebox{90}{{LongVBench}} &
        \rotatebox{90}{{MVBench}} &
        \rotatebox{90}{{Percept.\ Test}} \\
        \hline
        \rowcolor{navyblue!5}
        \textit{Chance-Level} & - & 34.0  & 22.0 & 20.0 & 25.0 & 20.0 & 14.0 & 25.0 & 27.3 & 33.3 \rule{0pt}{10pt} \\
        \hline
        \multirow{5}{*}{1M}
        & 0\%   & 26.0 & 20.2 & 32.5 & 52.1 & 46.9 & 32.0 & 51.4 & 50.5 & 54.2 \rule{0pt}{10pt} \\
        & 25\%  & 32.4 & 25.4 & 36.2 & 60.4 & 47.0 & 40.1 & 53.5 & 57.0 & 61.9 \\
        & 50\%  & 33.3 & 27.2 & 36.2 & 61.7 & 47.1 & 40.1 & 53.2 & 59.2 & 64.3 \\
        & 75\%  & 32.7 & 28.8 & 34.4 & 60.7 & 48.7 & 37.7 & 53.3 & 59.5 & 66.3 \\
        & 100\% & 34.4 & 28.4 & 35.1 & 61.3 & 48.9 & 39.6 & 53.0 & 60.1 & 67.5 \\
        \hline
        \multirow{5}{*}{4M}
        & 0\%   & 26.7 & 20.5 & 31.8 & 53.1 & 44.8 & 32.0 & 52.1 & 51.5 & 54.9 \rule{0pt}{10pt} \\
        & 25\%  & 32.3 & 26.7 & 37.0 & 61.3 & 45.0 & 38.6 & 53.1 & 57.6 & 61.9 \\
        & 50\%  & 31.9 & 27.4 & 37.2 & 61.9 & 45.7 & 38.1 & 54.2 & 59.5 & 65.2 \\
        & 75\%  & 33.8 & 27.9 & 36.2 & 61.1 & 47.3 & 40.9 & 53.1 & 60.1 & 67.0 \\
        & 100\% & 33.8 & 28.0 & 35.5 & 60.5 & 50.2 & 40.2 & 52.2 & 60.5 & 67.7 \\
        \hline
        \multirow{5}{*}{7M}
        & 0\%   & 25.8 & 18.9 & 31.6 & 53.7 & 48.1 & 31.9 & 52.5 & 51.4 & 55.4 \rule{0pt}{10pt}\\
        & 25\%  & 31.5 & 24.6 & 36.7 & 61.3 & 48.8 & 37.7 & 54.7 & 58.3 & 62.3 \\
        & 50\%  & 31.4 & 27.6 & 36.6 & 61.0 & 49.0 & 37.9 & 53.6 & 59.7 & 65.6 \\
        & 75\%  & 31.8 & 27.0 & 35.7 & 61.8 & 50.7 & 38.0 & 53.0 & 60.2 & 67.9 \\
        & 100\% & 32.6 & 27.7 & 37.3 & 62.1 & 52.4 & 39.4 & 54.3 & 60.6 & 68.8 \\
        \bottomrule
        \end{tabular}
    }
\end{table}

To elaborate on the respective contributions of image-based and video-based instruction tuning to a model's final video understanding capabilities, we conducted a series of experiments. These experiments employed varying proportions of image and video data during the finetuning stages, and we observed the resulting performance trends across diverse video benchmarks.

\noindent More specifically, for the initial image MLLM training, we randomly sampled 1M, 4M, and 7M image question-answering (QA) pairs from Cambrian-7M to train distinct models. Subsequently, for video-specific finetuning, we randomly sampled 25\%, 50\%, 75\%, and 100\% of video QA pairs from LLaVA-Video-178K ($\sim$1.6M data samples in total) to perform video-only finetuning on each of these pretrained image MLLMs. The hyperparameters for image instruction tuning and video finetuning were maintained as detailed in Table~\ref{tab:improved_cambrian_training_setups} and Table~\ref{tab:cambrian_s_training_setups}, respectively. The experimental results, presented in Table~\ref{tab:image_video_scale_results}, yield the following observations:

\begin{itemize}
    \item \textit{Models trained with more image data do not inherently outperform those trained with less when evaluated on video benchmarks without finetuning.} As indicated in the table, direct evaluation on video benchmarks reveals comparable performance across all three models, which were initially trained on 1M, 4M, and 7M image datasets, respectively.
    \item \textit{Finetuning on video data can be generally beneficial for models pretrained with larger image datasets, though not universally.} When all models were finetuned on 100\% video data, the model initially trained on 7M images outperformed the other two on 5 out of 9 video benchmarks (specifically, HourVideo, VideoMME, EgoSchema, LongVideoBench, and Perception Test).
    \item \textit{Incorporating video data into the training process consistently benefits performance across all video benchmarks.} We observed that finetuning an image-based MLLM with video data, even a small portion such as 25\%, improved its performance on all evaluated video benchmarks.
    \item \textit{Increasing the amount of video data used for finetuning does not guarantee consistent performance improvements across all benchmarks.} While video finetuning is generally advantageous, some benchmarks (\eg, VideoMME, VSI-Bench, Tomato) do not show further gains with more video data. For instance, models finetuned with 100\% video data exhibited performance on par with those finetuned with only 25\% video data on the VideoMME benchmark. Only EgoSchema, MVBench, and Perception Test demonstrated consistent benefits from increased video data, a phenomenon we hypothesize is related to the underlying video distribution of the training videos.
\end{itemize}

\subsection{On the Trade-off between Spatial Sensing and General Video Understanding}

\begin{figure}[!h]
    \centering
    \includegraphics[width=\linewidth]{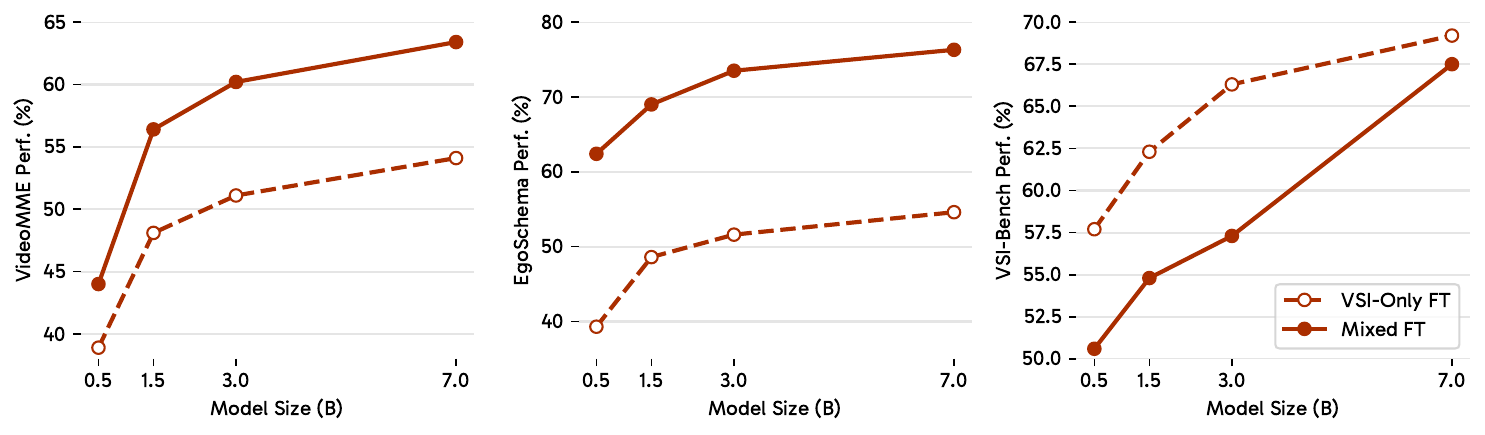}
    \caption{
        \textbf{On the trade-off between spatial-sensing and general video understanding.}
    }\label{fig:vsionlyft_vs_mixedft}
\end{figure}

In \cref{sec:limits:training_recipe}, we compare model performance when fine-tuned either on \vsidata{} alone or on a mixture of \vsidata{} and general video data. We observe that fine-tuning on \vsidata{} alone consistently yields higher performance on spatial sensing tasks, whereas mixed-data fine-tuning offers a better balance between spatial sensing and general video understanding.
To further explore this trade-off across model scales, we conduct fine-tuning after stage 3 using either \vsidata{} alone or the mixed dataset, under four different model sizes: 0.5B, 1B, 3B, and 7B parameters. We then evaluate these models on both general video understanding and spatial sensing benchmarks, as shown in \cref{fig:vsionlyft_vs_mixedft}.

The results confirm that the previous conclusion holds across all scales: \vsidata-only fine-tuning excels at spatial sensing, while mixed-data fine-tuning provides a better overall balance. Notably, however, the performance gap on VSI-Bench narrows as model size increases. We attribute this to the greater capacity of larger models to learn and retain diverse capabilities. This trend suggests that scaling to even larger models may further mitigate the spatial sensing performance drop typically observed when fine-tuning with mixed data.

\section{Predictive Sensing}\label{appx:pred_sensing}

\subsection{Latent Frame Prediction Implementation Details}\label{appx:pred_sensing:lfp_details}

\paragraph{Latent frame prediction head.}\label{appx:pred_sensing:nfp_head}

As shown in Algorithm~\ref{algo:pred_sensing:nfp_head}, our next-frame prediction head is a simple two-layer MLP with GELU activation~\cite{hendrycks2016gaussian}, running in parallel with the MLLM's original language model head. The output dimension is set to 1152, matching the output dimension of our vision encoder (\ie, \texttt{siglip2-so400m-patch14-384}).

\begin{algorithm}[h]
    \caption{\textbf{Latent frame prediction (LFP) head architecture (in PyTorch style).}}
    \label{algo:pred_sensing:nfp_head}
    \definecolor{mygray}{gray}{0.35}
    \definecolor{codeblue}{rgb}{0.25,0.5,0.5}
    \definecolor{codekw}{rgb}{0.85, 0.18, 0.50}
    \lstset{
      backgroundcolor=\color{white},
      basicstyle=\fontsize{8.0pt}{8.0pt}\color{mygray}\ttfamily\selectfont,
      columns=fullflexible,
      breaklines=true,
      captionpos=b,
      commentstyle=\fontsize{8.0pt}{8.0pt}\color{codekw},
      keywordstyle=\fontsize{8.0pt}{8.0pt}\color{codekw},
    }
    \begin{lstlisting}[language=python]
    LFPHead(           
      Sequential(
        (0): Linear(in_features=3584, out_features=3584, bias=True)
        (1): GELU(approximate=none)
        (2): Linear(in_features=3584, out_features=1152, bias=True)
        )
    )
    \end{lstlisting}
\end{algorithm}

\paragraph{On the balance between LFP and instruction tuning losses.}
\label{appx:pred_sensing:lfp_loss_weight_ablation}

As mentioned in \cref{sec:predictive-sensing:lfp}, to build the model's internal world model, we slightly modify our stage 4, introducing two auxiliary losses (\ie, cosine distance and mean-squared error) to optimize the next frame prediction objectiveness.
A coefficient is applied to balance the LFP loss against the instruction tuning loss, which we ablate in \cref{tab:lfp_loss_weight_ablation}.

\begin{table}[h]
    \centering
    \caption{\textbf{Evaluation results across different benchmarks with varying LFP loss weights.} Our default setup (0.1 loss coefficient) is highlighted in \colorbox{gray!10}{gray}.}
    \label{tab:lfp_loss_weight_ablation}
    \begin{tabular}{l|cccc}
    \toprule
    LFP loss coeffcient & VSI-Bench & VideoMME & EgoSchema & Perception Test \\
    \hline
    \rowcolor{navyblue!5}
    0.0 (\textit{\ie, No LFP Loss}) & 67.5 & 63.4 & 76.8 & 69.9 \rule{0pt}{10pt} \\
    \hline
    \rowcolor{gray!10}
    0.1 & 66.1 & 63.9 & 76.9 & 69.7 \\
    0.5 & 60.8 & 63.6 & 77.2 & 66.4 \\
    1.0 & 56.6 & 61.0 & 72.9 & 65.1 \\
    \bottomrule
    \end{tabular}
\end{table}

\subsection{Memory Framework Design for \vsisuper Recall}
\label{appx:pred_sensing:vso_mem_design}

As introduced in main paper (and shown in Algorithm~\ref{algo:pred_sensing:vso_mem_design}), our predictive memory mechanism comprises three distinct memory levels ($M_s$, $M_l$, $M_w$) and four key transition functions governing their interaction: \textit{Sensory Streaming}, \textit{Memory Compression}, \textit{Memory Consolidation}, and \textit{Retrieval}. This section details the implementation of these functions.

\paragraph{Basic memory units.}
For our implementation, we utilize the \textit{encoded key-value pairs} from each Large Language Model (LLM) layer as the basic memory units. This choice, rather than using output latent features from a vision encoder or vision-language connector, allows us to fully leverage the LLM's internal capabilities for memory construction without requiring external modules. This design decision will be elaborated upon in subsequent sections.

\paragraph{Streaming sensing.}
Each incoming frame is initially processed independently by the vision encoder and the vision-language connector with a window size of $W_s$. Subsequently, it is further encoded by the LLM, referencing selected previous frames.
The key-value pairs from these preceding frames, cached in the \textit{Sensory memory buffer} ($M_s$), provide the necessary context for this encoding step.

\paragraph{Surprise-based memory compression.}
In the meantime of encoding a single frame, we assess its ``surprise'' level.
This is achieved by calculating the difference between the model's prediction for the current frame and the actual ground truth observation (both in the latent feature space).
 When a frame of timestamp $t$ is moved from the sensory memory buffer $M_s$ to the long-term memory $M_l$, if it is deemed non-surprising (\ie, its surprise score is below a predefined threshold $T_s$), we will downsample its' key-value pairs by a factor of 2 along the spatial ($H\times W$) dimension.
 This surprise-based compression mitigates redundancy in the information stored within $M_l$.

\paragraph{Surprise-based memory consolidation.}
Long-term memory $M_l$ is initialized with a predefined budget size $B_{long}$ (\eg, 32,768 tokens).
When the volume of memory tokens surpasses this budget, we apply a \textit{surprise-based} consolidation function to $M_l$ to ensure it remains within the allocated limit.
Our consolidation function is straightforward yet effective: we identify the surprise score associated with each frame in $M_l$.
Then, the frame with the lowest surprise score is removed (or ``forgotten'').
Then, we merge or drop some of these frames according to their surprise scores (we tried three different strategies here: 1. forget the oldest memory, 2. forget the least surprise memory, and 3. forget the least surprise memory while merging adjacent surprise memories if any adjacent surprise memories exist).
This process is iterated until the total size of $M_l$ falls below the budget.

\paragraph{Retrieval.}
Upon receiving a user query $q$, we first retrieve the most relevant frames from the long-term memory ($M_l$) to construct the working memory ($M_w$).
This $M_w$ then serves as the context for answering the user's query.
To perform this retrieval efficiently without resorting to external modules, we utilize the inherent similarity measurement capabilities of the LLM's attention mechanism.
Specifically, for each transformer layer, the user query $q$ is transformed into the attention mechanism's query feature space.
We then compute the similarity between this query feature and the key features of each frame stored in $M_l$.
Similarity is measured using cosine distance, and for simplicity, multi-head features are treated as a single feature.
The $k$ frames with the highest similarity scores have their key-value pairs selected and utilized by the attention mechanism to further encode the user query.

\begin{algorithm}[H]
    \small
    \caption{\textbf{Memory framework design for \vsisuper Recall.}}
    \label{algo:pred_sensing:vso_mem_design}
    \KwIn{Frames $\{f_1, \dots, f_T\}$, User query $q$}
    \KwIn{Encoder $\mathcal{E}$, Decoder $\mathcal{D}$, Surprise Estimator $\mathcal{S}$, Surprise threshold $\tau$}
    \KwIn{Compression function $\mathcal{C}$, Consolidation function $\mathcal{G}$, Retrieval function $\mathcal{R}$}
    \KwIn{Sensory memory $\mathcal{M}_s \gets \emptyset$ with budget $B_s$, Long-term memory $\mathcal{M}_l \gets \emptyset$ with budget $B_l$, Working memory $\mathcal{M}_w \gets \emptyset$}
    
    \For{$t \gets 1$ \KwTo $T$}{
        $z_t \gets \mathcal{E}(f_t, \mathcal{M}_s)$\;
        $\mathcal{M}_s \gets \mathcal{M}_s \cup \{z_t\}$ \tcp*[r]{Streaming sensing}
        
        $s_t \gets \mathcal{S}(f_t, \mathcal{M}_s)$ \tcp*[r]{Surprise estimation}
        
        \While{$|\mathcal{M}_s| > B_s$}{
            Dequeue $z_{\text{old}}$ from $\mathcal{M}_s$\;
            
            $m \gets \mathbf{1}[s_t \geq \tau] \cdot z_{\text{old}} + \mathbf{1}[s_t < \tau] \cdot \mathcal{C}(z_{\text{old}})$ \tcp*[r]{Selective compression}
            
            $\mathcal{M}_l \gets \mathcal{M}_l \cup \{m\}$\;
            
            \If{$|\mathcal{M}_l| > B_l$}{
                $\mathcal{M}_l \gets \mathcal{G}(\mathcal{M}_l)$ \tcp*[r]{Memory consolidation}
            }
        }
    }
    
    $\mathcal{M}_w \gets \mathcal{R}(q, \mathcal{M}_l)$ \tcp*[r]{Retrieve working memory}
    $\hat{a} \gets \mathcal{D}(q, \mathcal{M}_w)$ \tcp*[r]{Answering query with $\mathcal{M}_w$}
    \Return $\hat{a}$
\end{algorithm}

\subsection{Agentic Framework Design for \vsisuper Count}
\label{appx:pred_sensing:vsc_agentic_design}

Algorithm~\ref{algo:pred_sensing:vsc_agentic_design} presents our agentic framework for the \vsisuper Count task. Similar to the memory design in Algorithm~\ref{algo:pred_sensing:vso_mem_design}, we encode sensory frames using a sliding window approach with a window size of $W_s$.
The latent frame prediction module continuously estimates the expected next frame and computes the prediction error to quantify how "surprise" the actual next frame is. As new frame arrivs, the oldest frames that exceed the sensory memory window are dequeued and stored in the long-term memory.
If a dequeued frame is deemed “surprising” (\ie, its prediction error exceeds a predefined threshold $\tau$), which may indicate a scene or spatial boundary, we trigger a query response using the accumulated long-term memory and reset it afterward. The generated response is then stored in the answer memory bank. The final answer is computed as the aggregation of all intermediate answers stored in this bank.

\begin{algorithm}[H]
    \small
    \caption{\textbf{Agentic framework design for \vsisuper Count task.}}
    \label{algo:pred_sensing:vsc_agentic_design}
    \KwIn{Frames $\{f_1, \dots, f_T\}$, user query $q$}
    \KwIn{Encoder $\mathcal{E}$, Decoder $\mathcal{D}$, Surprise Estimator $\mathcal{S}$, threshold $\tau$}
    \KwIn{Sensory memory $\mathcal{M}_s \gets \emptyset$ with budget $B_s$}
    \KwIn{Long-term memory $\mathcal{M}_l \gets \emptyset$, Answer memory bank $\mathcal{M}_{\text{Ans}} \gets \emptyset$}

    \For{$t \gets 1$ \KwTo $T$}{
        $z_t \gets \mathcal{E}(f_t, \mathcal{M}_s)$\;
        $\mathcal{M}_s \gets \mathcal{M}_s \cup \{z_t\}$ \tcp*[r]{Streaming sensing}

        $s_t \gets \mathcal{S}(f_t, \mathcal{M}_s)$ \tcp*[r]{Surprise estimation}

        \If{$|\mathcal{M}_s| > B_s$}{
            Remove oldest $z_{\text{old}}$ from $\mathcal{M}_s$\;
            $\mathcal{M}_l \gets \mathcal{M}_l \cup \{z_{\text{old}}\}$ \tcp*[r]{Store to long-term memory}
        }

        \If{$s_t \geq \tau$}{
            $\hat{a} \gets \mathcal{D}(q, \mathcal{M}_l)$ \tcp*[r]{Answer query using long-term memory}
            $\mathcal{M}_{\text{Ans}} \gets \mathcal{M}_{\text{Ans}} \cup \{\hat{a}\}$\;
            $\mathcal{M}_l \gets \emptyset$ \tcp*[r]{Reset long-term memory}
        }
    }

    \Return \texttt{Sum}($\mathcal{M}_{\text{Ans}}$)
\end{algorithm}

\subsection{Comparisons with Existing Long-video Methods}
\label{appx:pred_sensing:comparisons_with_existing_long_video_methods}

We compare our method (both surprise-driven memory and agentic framework) with existing methods designed for long-video understanding, in \cref{tab:comparisons_with_existing_long_video_methods}.
Specifically, all experiments here are conducted with our LFP-finetuned \cambrianS-7B, with a different strategy to handle the ever-expanding visual sensory input.
For MovieChat, we follow the official implementation in \cite{song2024moviechat}, maintain a fixed-size long-term memory bank, and set the long-term and short-term memory budgets to 64 and 16, respectively.
For Flash-VStream~\cite{zhang2024flash}, as its abstract memory module introduces additional parameters and requires a dedicated training process, we only implement the three remaining memory components (\ie, spatial memory, temporal memory, and retrieved memory), and keeping all other hyperparameters aligned with the default setup.

\begin{table}[!h]
    \centering
    \caption{\textbf{Compare our framework with existing long-video methods on \vsisuper.}}
    \label{tab:comparisons_with_existing_long_video_methods}
    \resizebox{0.75\textwidth}{!}{
        \begin{tabular}{r|ccccc|cccc}
            & \multicolumn{5}{c|}{\vso (Duration in Mins.)} & \multicolumn{4}{c}{\vsc (Duration in Mins.)} \\
            Eval Setups & 10 & 30 & 60 & 120 & 240 & 10 & 30 & 60 & 120 \\
            \hline
            MovieChat & 18.3 & 21.7 & 16.7 & 26.7 & 25.6 & 0.0 & 0.0 & 0.0 & 0.0 \rule{0pt}{10pt} \\
            Flash-VStream & 28.3 & 33.3 & 23.3 & 28.3 & 31.7 & 0.0 & 0.0 & 0.0 & 0.0 \rule{0pt}{10pt} \\
            \hline
            Ours & 45.0 & 41.7 & 40.0 & 40.0 & 40.0 & 40.6 & 42.0 & 35.0 & 34.0 \rule{0pt}{10pt} \\
        \end{tabular}
    }
\end{table}

\begin{center}
    \begin{minipage}{1.0\linewidth}
        \centering
        \includegraphics[width=\linewidth]{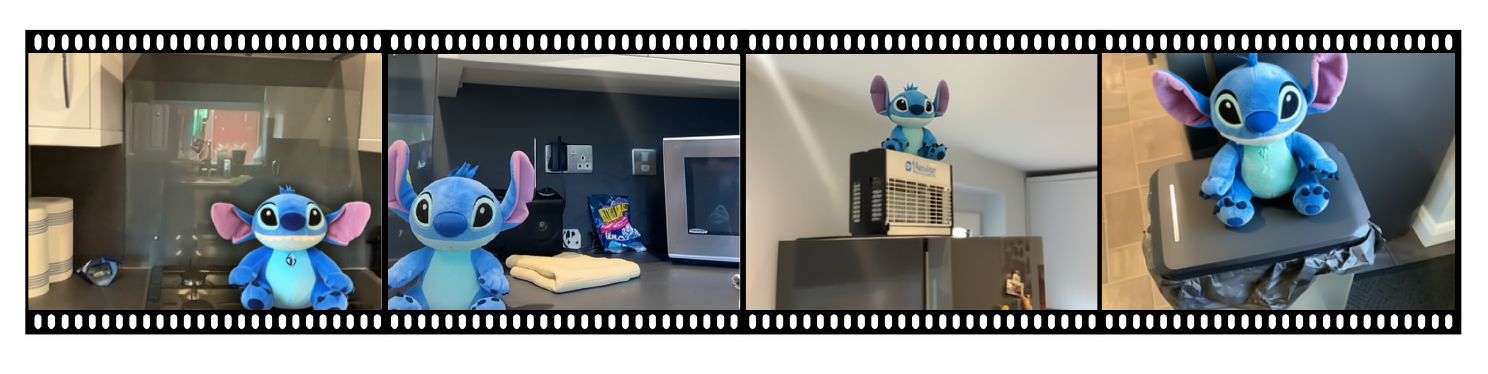}
        \vspace{-1cm}
        \begin{tcolorbox}[colback=gray!5!white, colframe=gray!75!gray, width=0.97\linewidth]
            Which of the following correctly represents the order in which the Stitch appeared in the video? \\
            A. Stove, Trash bin, Refrigerator, Counter\hfill B. Trash bin, Refrigerator, Counter, Stove \\
            C. Stove, Counter, Refrigerator, Trash bin\hfill D. Trash bin, Stove, Counter, Refrigerator
        \end{tcolorbox}
        \vspace{-0.5em}

        \includegraphics[width=\linewidth]{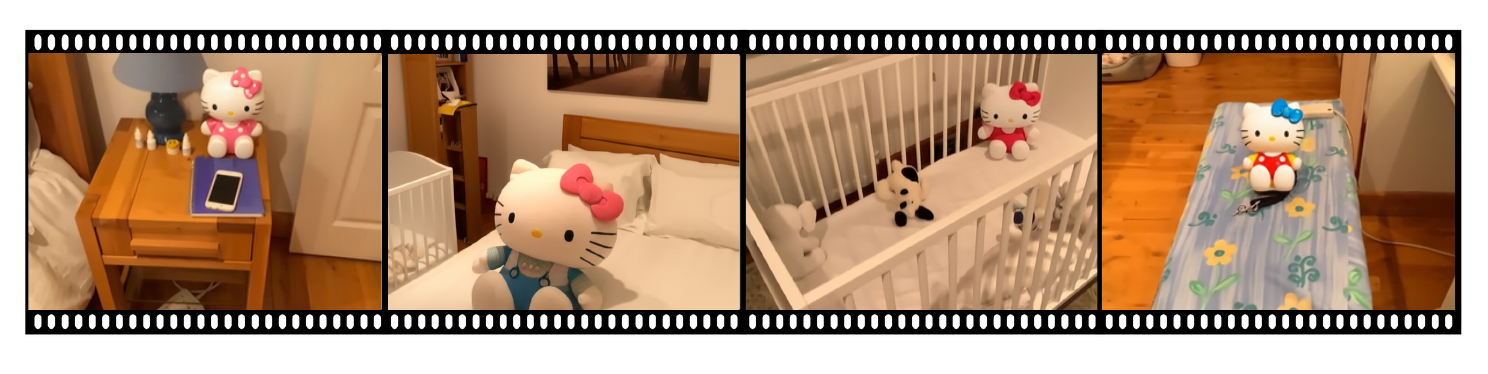}
        \vspace{-1cm}
        \begin{tcolorbox}[colback=gray!5!white, colframe=gray!75!gray, width=0.97\linewidth]
            Which of the following correctly represents the order in which the Hello Kitty appeared in the video? \\
            A. Nightstand, Bed, Crib, Blue bench\hfill B. Blue bench, Crib, Nightstand, Bed \\
            C. Bed, Nightstand, Blue bench, Crib\hfill D. Blue bench, Bed, Crib, Nightstand
        \end{tcolorbox}
        \vspace{-0.5em}

        \includegraphics[width=\linewidth]{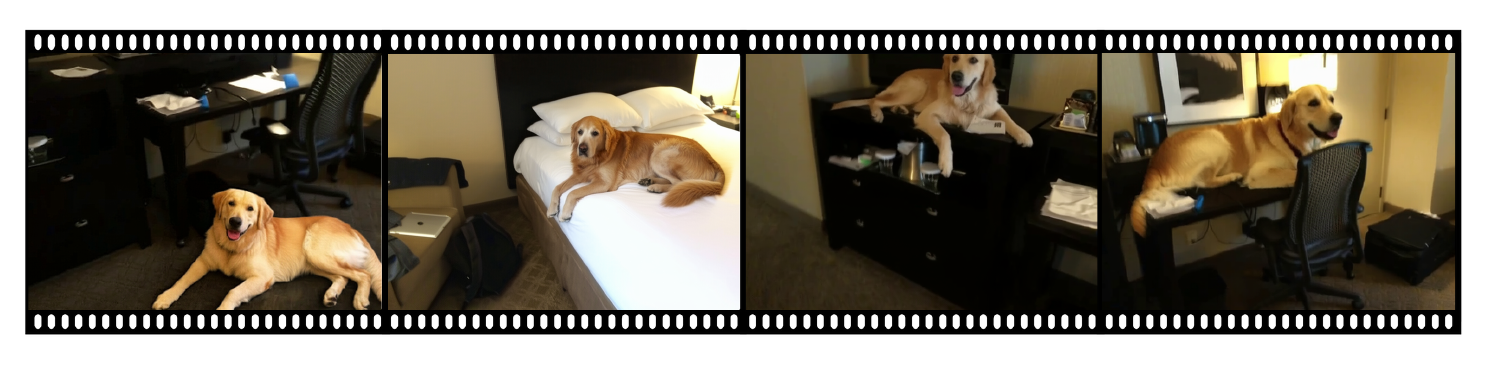}
        \vspace{-1cm}
        \begin{tcolorbox}[colback=gray!5!white, colframe=gray!75!gray, width=0.97\linewidth]
            Which of the following correctly represents the order in which the Golden Retriever appeared in the video? \\
            A. Bed, Table, Chest of drawers, Floor\hfill B. Table, Chest of drawers, Bed, Floor \\
            C. Chest of drawers, Floor, Table, Bed\hfill D. Floor, Bed, Chest of drawers, Table
        \end{tcolorbox}
        \vspace{-0.5em}

        \includegraphics[width=0.85\linewidth]{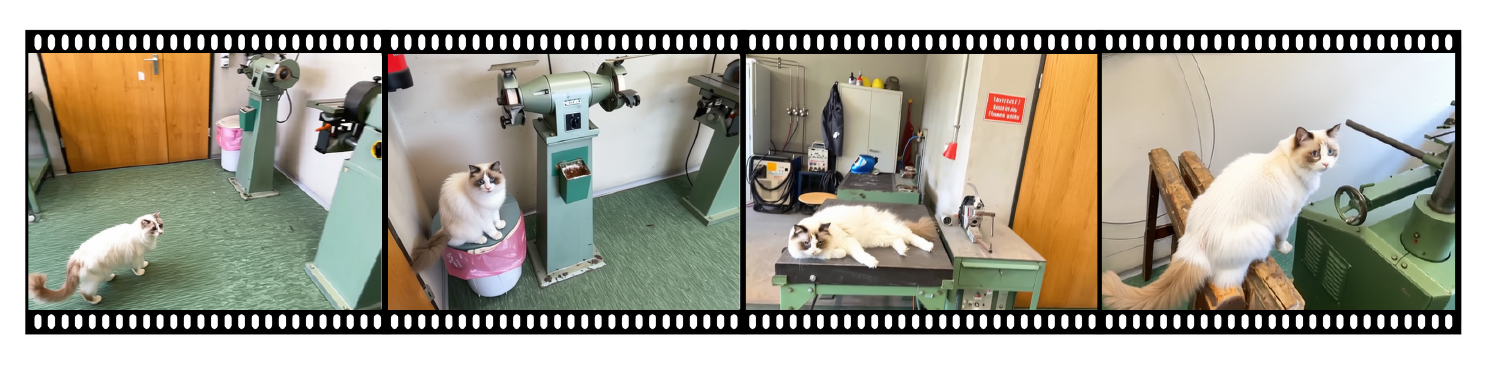}
        \vspace{-1cm}
        \begin{tcolorbox}[colback=gray!5!white, colframe=gray!75!gray, width=0.97\linewidth]
            Which of the following correctly represents the order in which the white Ragdoll cat appeared in the video? \\
            A. Ground, Trash bin, Bench, Table\hfill B. Table, Bench, Ground, Trash bin \\
            C. Ground, Trash bin, Table, Bench\hfill D. Trash bin, Bench, Table, Ground
        \end{tcolorbox}
        \captionof{figure}{More examples of our \vsisuper Recall benchmark. Note that only edited frames are visualized.}
        \label{fig:sor_examples}
    \end{minipage}
\end{center}

\begin{center}
    \begin{minipage}{1.0\linewidth}
        \centering
        \includegraphics[width=\linewidth]{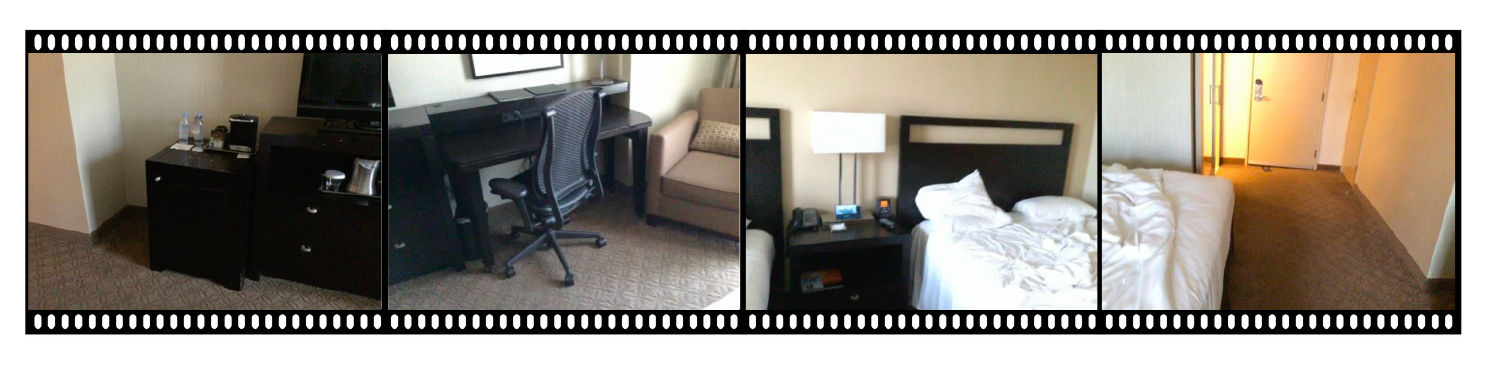}
        \vspace{-1cm}
        \begin{tcolorbox}[colback=gray!5!white, colframe=gray!75!gray, width=0.97\linewidth, title=Absolute Direction (Object)]
            Standing by the backpack, looking toward the table, how far counterclockwise in degrees must I turn to see the trash bin? \\
            Answer: 334.09
        \end{tcolorbox}
        \vspace{-0.9em}

        \begin{tcolorbox}[colback=gray!5!white, colframe=gray!75!gray, width=0.97\linewidth, title=Absolute Distance]
            Measuring from the closest points of each, how far apart are the chair and the door in meters? \\
            Answer: 2.32
        \end{tcolorbox}
        \vspace{-0.9em}

        \begin{tcolorbox}[colback=gray!5!white, colframe=gray!75!gray, width=0.97\linewidth, title=Absolute Distance]
            Considering the chair and the door, which object's longest edge is the shorter? \\
            A. Door \\
            B. Chair
        \end{tcolorbox}
        \vspace{-0.9em}
        \captionof{figure}{Examples of \vsidata (Annotated Real Video).}
        \label{fig:annotated_real_video_1}
    \end{minipage}
\end{center}

\begin{center}
    \begin{minipage}{1.0\linewidth}
        \centering
        \includegraphics[width=\linewidth]{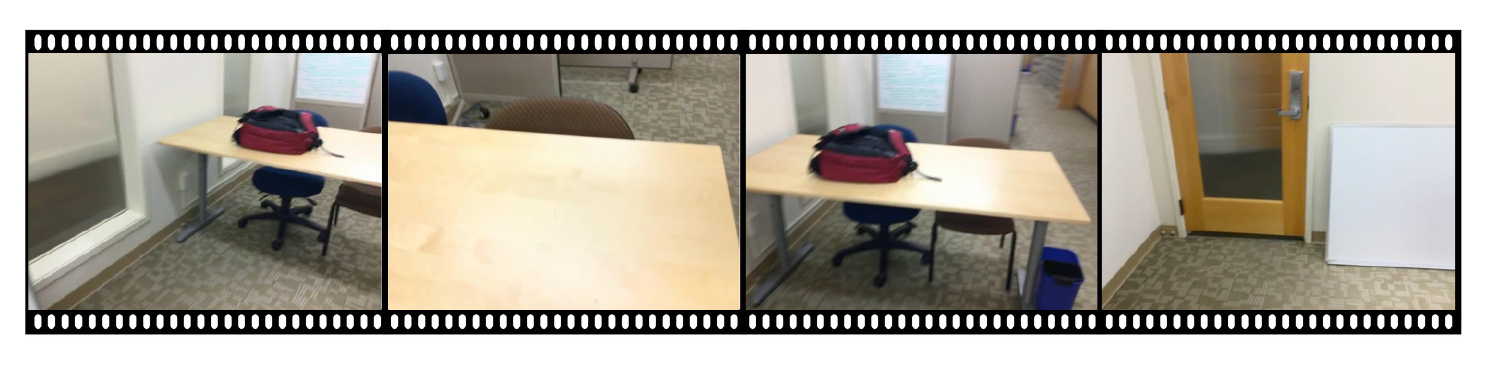}
        \vspace{-1cm}
        \begin{tcolorbox}[colback=gray!5!white, colframe=gray!75!gray, width=0.97\linewidth, title=Object Appearance Order]
            Determine the initial appearance order of these categories in the video: door, chair, lamp, refrigerator. \\
            A. refrigerator, door, lamp, chair\hfill B. refrigerator, chair, door, lamp \\
            C. refrigerator, chair, lamp, door\hfill D. door, chair, lamp, refrigerator
        \end{tcolorbox}
        \vspace{-0.9em}

        \begin{tcolorbox}[colback=gray!5!white, colframe=gray!75!gray, width=0.97\linewidth, title=Absolute Size]
            Provide the longest side's length for the door in inches. \\
            Answer: 72.00
        \end{tcolorbox}
        \vspace{-0.9em}

        \begin{tcolorbox}[colback=gray!5!white, colframe=gray!75!gray, width=0.97\linewidth, title=Room Size]
            Indicate the room's dimensions in square feet. If there's more than one room, estimate their total size. \\
            Answer: 232.76
        \end{tcolorbox}
        \vspace{-0.9em}
        \captionof{figure}{Examples of VSI-590K  (Annotated Real Video).}
        \label{fig:annotated_real_video_2}
    \end{minipage}
\end{center}

\begin{center}
    \begin{minipage}{1.0\linewidth}
        \centering
        \includegraphics[width=\linewidth]{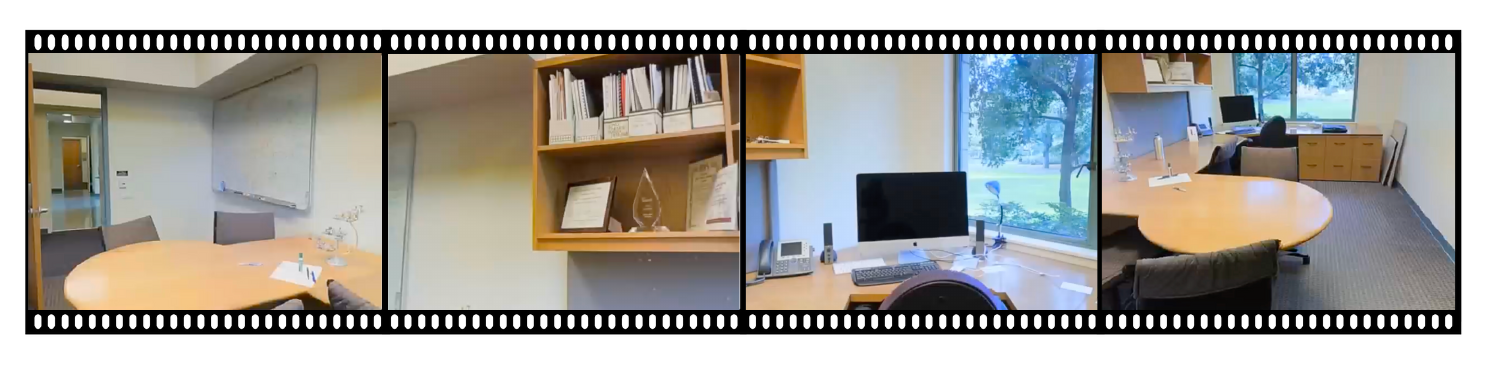}
        \vspace{-1cm}
        \begin{tcolorbox}[colback=gray!5!white, colframe=gray!75!gray, width=0.97\linewidth, title=Relative Direction (Object Perspective)]
            Facing the door while standing near the window, in which of the following positions is the board relative to me: front-left, front-right, back-left, or back-right? Use Cartesian quadrants, with me at the origin looking toward positive y-axis \\
            A. Back-right \\
            B. Front-right \\
            C. Front-left \\
            D. Back-left
        \end{tcolorbox}
        \vspace{-0.9em}

        \begin{tcolorbox}[colback=gray!5!white, colframe=gray!75!gray, width=0.97\linewidth, title=Relative Distance (Object Perspective)]
            Identify the object among (bookcase, chair, board, door) that is closest to the window based on the shortest distance between their closest points. Choose the nearest instance if several exist. \\
            A. Bookcase \\
            B. Chair \\
            C. Board \\
            D. Door
        \end{tcolorbox}
        \vspace{-0.9em}

        \captionof{figure}{Examples of VSI-590K  (Annotated Real Video).}
        \label{fig:annotated_real_video_3}
    \end{minipage}
\end{center}

\begin{center}
    \begin{minipage}{1.0\linewidth}
        \centering
        \includegraphics[width=\linewidth]{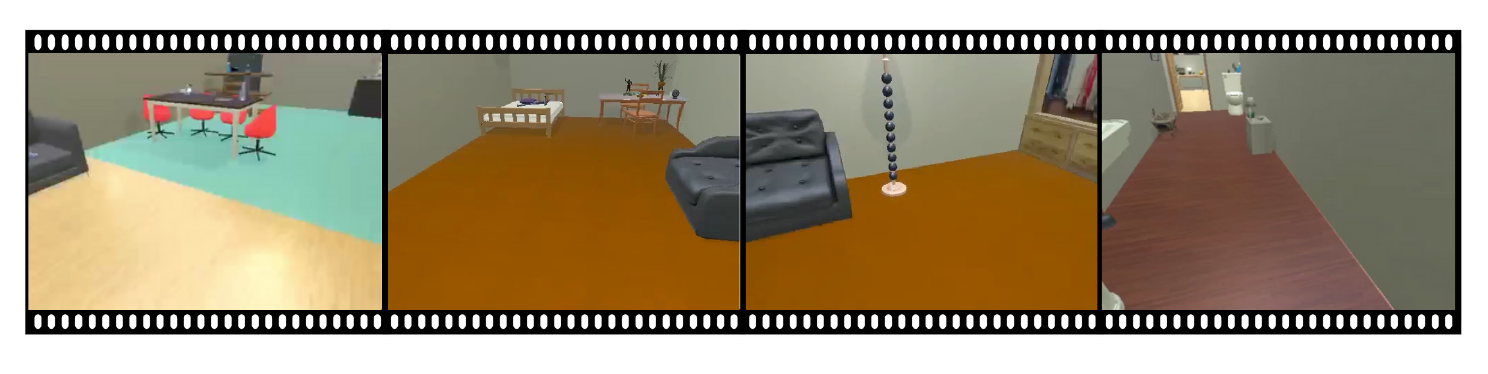}
        \vspace{-1cm}
        \begin{tcolorbox}[colback=gray!5!white, colframe=gray!75!gray, width=0.97\linewidth, title=Relative Distance (Object Perspective)]
            If I am standing by the dresser and facing the chair, is the closet to my left, right, or back? An object is to my back if I would have to turn at least 135 degrees in order to face it. \\
            A. Left \\
            B. Right \\
            C. Back
        \end{tcolorbox}

        \captionof{figure}{Examples of VSI-590K  (Annotated Simulated Video).}
        \label{fig:annotated_simulated_video}
    \end{minipage}
\end{center}

\begin{center}
    \begin{minipage}{1.0\linewidth}
        \centering
        \includegraphics[width=0.5\linewidth]{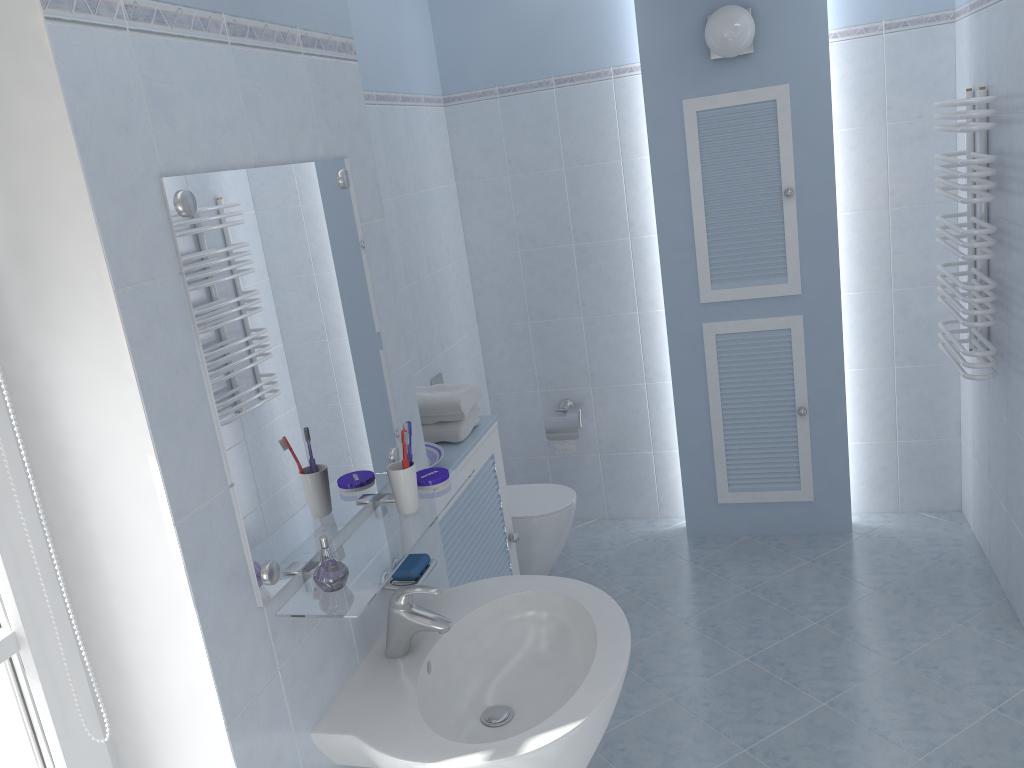}
        \vspace{-0.7em}
        \begin{tcolorbox}[colback=gray!5!white, colframe=gray!75!gray, width=0.97\linewidth, title=Relative Direction (Object Perspective)]
            With the toilet beside me and facing the cabinet, is the lamp positioned front-left, front-right, back-left, or back-right relative to me, based on Cartesian plane quadrants? \\
            A. Back-right \\
            B. Front-right \\
            C. Front-left \\
            D. Back-left
        \end{tcolorbox}
        \vspace{-0.9em}

        \begin{tcolorbox}[colback=gray!5!white, colframe=gray!75!gray, width=0.97\linewidth, title=Relative Distance (Object Perspective)]
            Identify the object among (bookcase, chair, board, door) that is closest to the window based on the shortest distance between their closest points. Choose the nearest instance if several exist. \\
            A. Bookcase \\
            B. Chair \\
            C. Board \\
            D. Door
        \end{tcolorbox}
        \vspace{-0.9em}

        \captionof{figure}{Examples of VSI-590K  (Annotated Simulated Video (Frame)).}
        \label{fig:annotate_simulated_video_frame}
    \end{minipage}
\end{center}

\begin{center}
    \begin{minipage}{1.0\linewidth}
        \centering
        \includegraphics[width=0.5\linewidth]{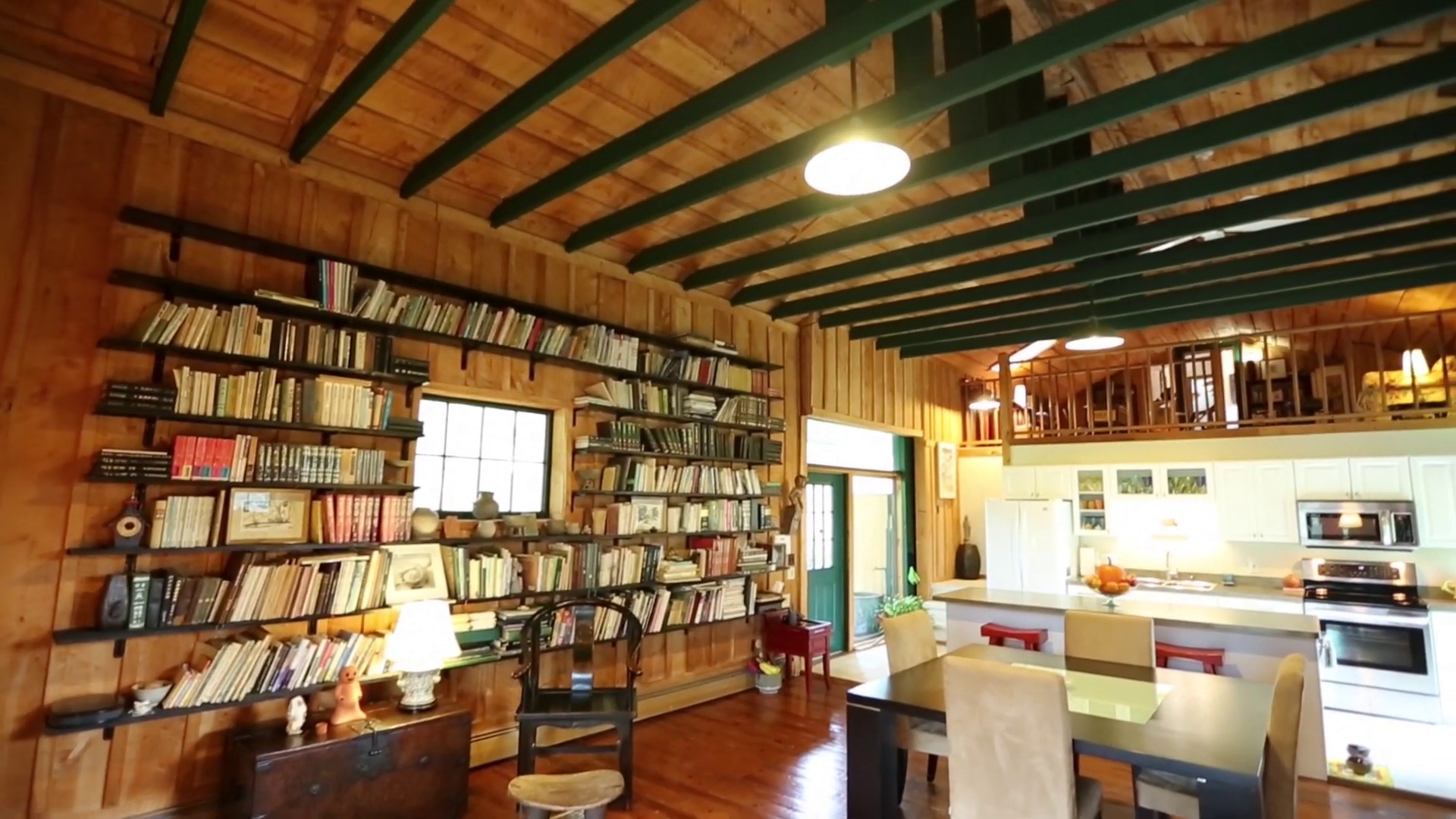}
        \vspace{-0.7em}
        \begin{tcolorbox}[colback=gray!5!white, colframe=gray!75!gray, width=0.97\linewidth, title=Object Counting (Relative)]
            If counted, would chairs be fewer than, more than, or equal in number to tables? \\
            A. Fewer \\
            B. More \\
            C. Equal
        \end{tcolorbox}
        \vspace{-0.9em}

        \begin{tcolorbox}[colback=gray!5!white, colframe=gray!75!gray, width=0.97\linewidth, title=Relative Direction (Camera Perspective)]
            Through the camera's lens, is the sink captured on the left or right part of the scene? \\
            A. Right \\
            B. Left
        \end{tcolorbox}
        \vspace{-0.9em}

        \captionof{figure}{Examples of VSI-590K  (Unannotated Real Video (Frame)).}
        \label{fig:unannotated_image_1}
    \end{minipage}
\end{center}

\begin{center}
    \begin{minipage}{1.0\linewidth}
        \centering
        \includegraphics[width=0.5\linewidth]{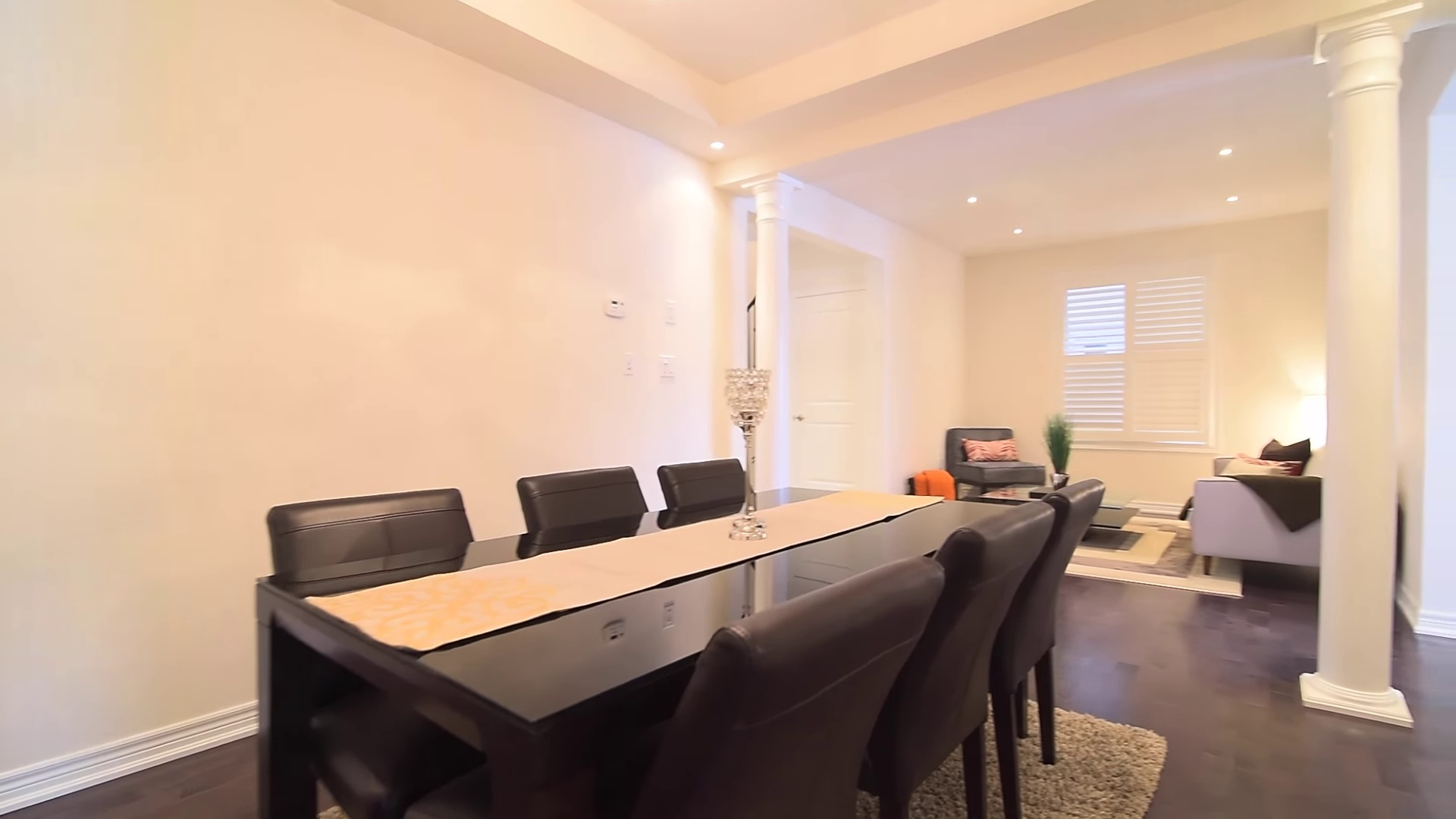}
        \vspace{-0.7em}
        \begin{tcolorbox}[colback=gray!5!white, colframe=gray!75!gray, width=0.97\linewidth, title=Object Counting (Absolute)]
            What would be the count if you tallied all the chairs? \\
            Answer: 6
        \end{tcolorbox}
        \vspace{-0.9em}

        \begin{tcolorbox}[colback=gray!5!white, colframe=gray!75!gray, width=0.97\linewidth, title=Relative Distance (Camera Perspective)]
            In terms of proximity to the camera, which is closer: a table or a sofa? \\
            A. Table \\
            B. Sofa
        \end{tcolorbox}
        \vspace{-0.9em}

        \captionof{figure}{Examples of \vsidata  (Unannotated Real Video (Frame)).}
        \label{fig:unannotated_image_2}
    \end{minipage}
\end{center}

\end{document}